\documentclass[11pt]{article}
\usepackage[numbers]{natbib}
\usepackage{./macros/packages}
\usepackage{./macros/editing-macros}
\usepackage{./macros/formatting}
\usepackage{./macros/statistics-macros}
\usepackage{multirow}
\usepackage{booktabs}
\usepackage{wrapfig}

\usepackage[textsize=tiny]{todonotes}
\begin{document}

% Control whitespace around equations
\abovedisplayskip=8pt plus0pt minus3pt
\belowdisplayskip=8pt plus0pt minus3pt

% ------------------------------------------------------------------------
% Main Paper Body
% ------------------------------------------------------------------------

% ------------------------------------------------------------------------
% Default title and authorship
% ------------------------------------------------------------------------
\begin{center}
  {\huge Differentiable Discrete Event Simulation \vspace{.2cm}
  \\ for Queuing Network Control } \\
  \vspace{.5cm} {\Large Ethan Che ~~~ Jing Dong ~~~ Hongseok Namkoong} \\
  \vspace{.2cm}
  {\large Columbia Business School} \\
  \vspace{.2cm}
  \texttt{\{eche25, jing.dong, namkoong\}@gsb.columbia.edu}
\end{center}

% ------------------------------------------------------------------------
% Abstract
% ------------------------------------------------------------------------

\begin{abstract}%
  Queuing network control is essential for managing congestion in job-processing systems 
such as service systems, communication networks, and manufacturing processes. Despite growing interest in applying reinforcement learning (RL) techniques,  queueing network control
%model-free reinforcement learning (RL) have been proposed to train control policies for arbitrary network topologies. 
poses distinct challenges, including high stochasticity, large state and action spaces, and lack of stability. 
To tackle these challenges, we propose a scalable framework for policy optimization based on differentiable discrete event simulation. 
Our main insight is that by implementing a well-designed smoothing technique for discrete event dynamics, we can compute $\pathwise$ policy gradients for large-scale queueing networks using auto-differentiation software (e.g., Tensorflow, PyTorch) and GPU parallelization. Through extensive empirical experiments, we observe that our policy gradient estimators are several orders of magnitude more accurate than typical $\reinforce$-based  estimators. 
In addition, we propose a new policy architecture, which drastically improves stability while maintaining the flexibility of neural-network policies.
In a wide variety of scheduling and admission control tasks, we demonstrate that training control policies with pathwise gradients leads to a 50-1000x improvement in sample efficiency over state-of-the-art RL methods.
Unlike prior tailored approaches to queueing, our methods can flexibly handle realistic scenarios, including systems operating in non-stationary environments and those with non-exponential interarrival/service times.

% To tackle these challenges, 
% we propose a scalable framework for policy optimization based on differentiable simulation
% and demonstrate substantial improvements in  sample efficiency in multi-class queuing networks.
% %and reduces sample inefficiency of existing learning approaches. 
% Our main algorithmic insight is to compute pathwise gradients for 
% discrete event dynamical systems with auto-differentiation (e.g. Tensorflow or PyTorch).
% We address the non-differentiability of the discrete event dynamics through carefully designed smoothing. 
% We find that for scheduling and admission control tasks, training control policies with pathwise gradients leads to a 50-1000x improvement in sample efficiency over state-of-the-art RL methods.
% Our methods can handle realistic instances, including systems operating in a non-stationary environment and with non-exponential system inputs (i.e., interarrival and service times), which are relevant for many practical settings.\hntodo{This last bit is confusing because everyone who don't know D and G (meaning almost every reader) will be left perplexed why black-box RL methods don't apply. Need to qualify by saying queueing-specific innovations that stabilize SoTA RL methods etc...}

%%% TeX-master: "abstract"
%%% Local Variables: %%% mode: latex %%% TeX-master: "main" %%% End:

\end{abstract}

\section{Introduction}
\label{section:introduction}

% {\color{blue} talk about the figures, try to highlight broader implications/operational problems. Main algorithmic insight is .... (1) our smoothing can be applied more broadly to other discrete event environments (2) our overall method of combining structure + data is useful. CLARIFY SAMPLE EFFICIENCY (learn more from a single trajectory)}

Queuing models are a powerful modeling tool to conduct performance analysis and optimize operational policies in diverse applications such as service systems (e.g., call centers~\cite{aksin2007modern}, healthcare delivery systems~\cite{armony2015patient}, ride-sharing platforms~\cite{banerjee2022pricing}, etc), computer and communication systems~\cite{harchol2013performance, neely2022stochastic}, manufacturing systems~\cite{shanthikumar2007queueing}, and financial systems (e.g., limit order books~\cite{cont2010stochastic}). Standard tools for queuing control analysis involve establishing structural properties of the underlying Markov decision process (MDP) or leveraging analytically more tractable approximations such as fluid~\cite{dai1995stability, chen1991discrete} or diffusion approximations~\cite{harrison1998heavy, stolyar2004maxweight, harrison1989scheduling, mandelbaum2004scheduling}. These analytical results often give rise to simple control policies that are easy to implement and interpret. However, these policies only work under restrictive modeling assumptions and can be highly sub-optimal outside of these settings. Moreover, deriving a good policy for a given queuing network model requires substantial queuing expertise and can be theoretically challenging. 

\begin{figure}[ht]
\hspace{-1em}
    \includegraphics[height = 2.1in]{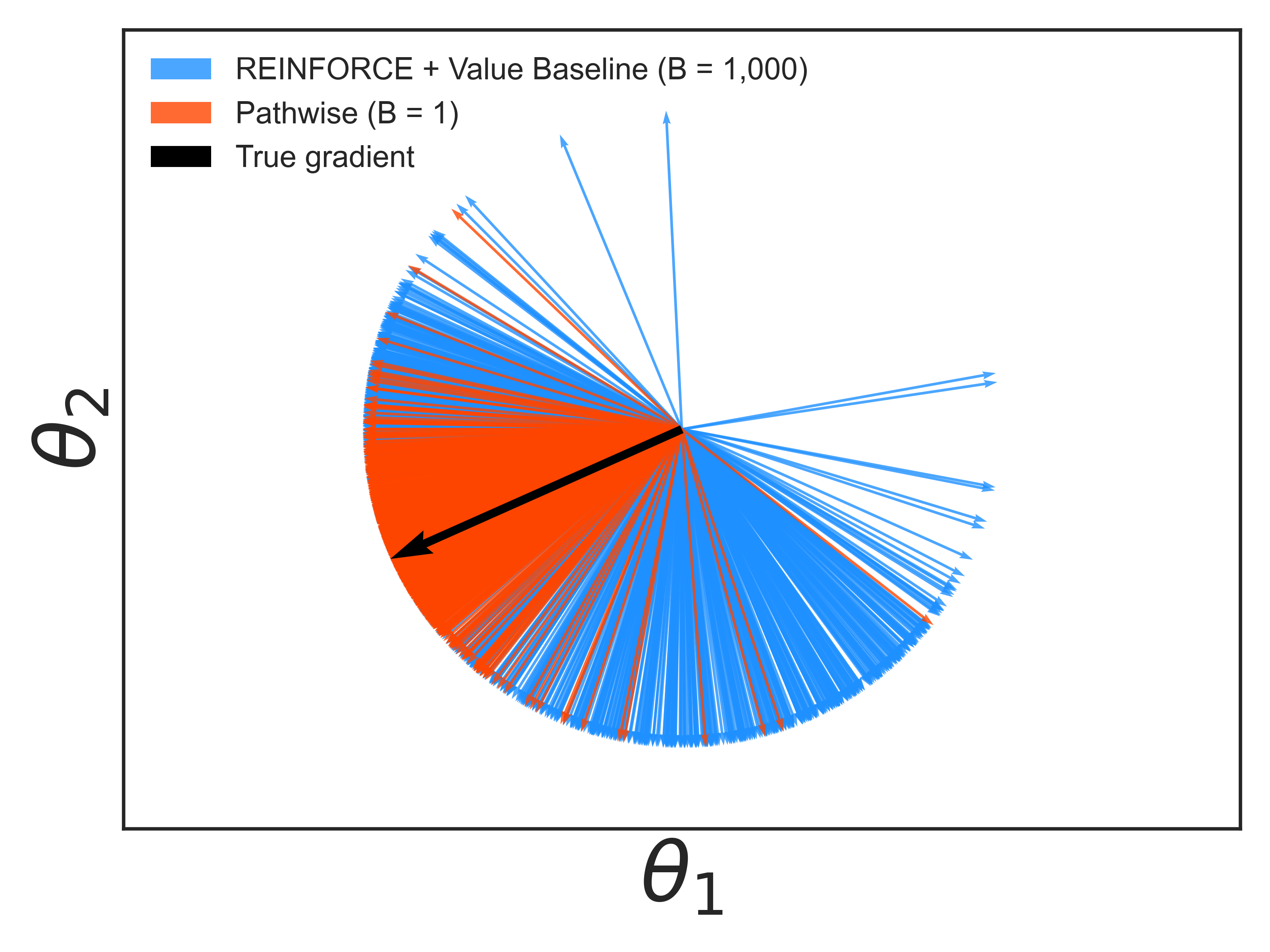}
    \includegraphics[height = 2.1in]{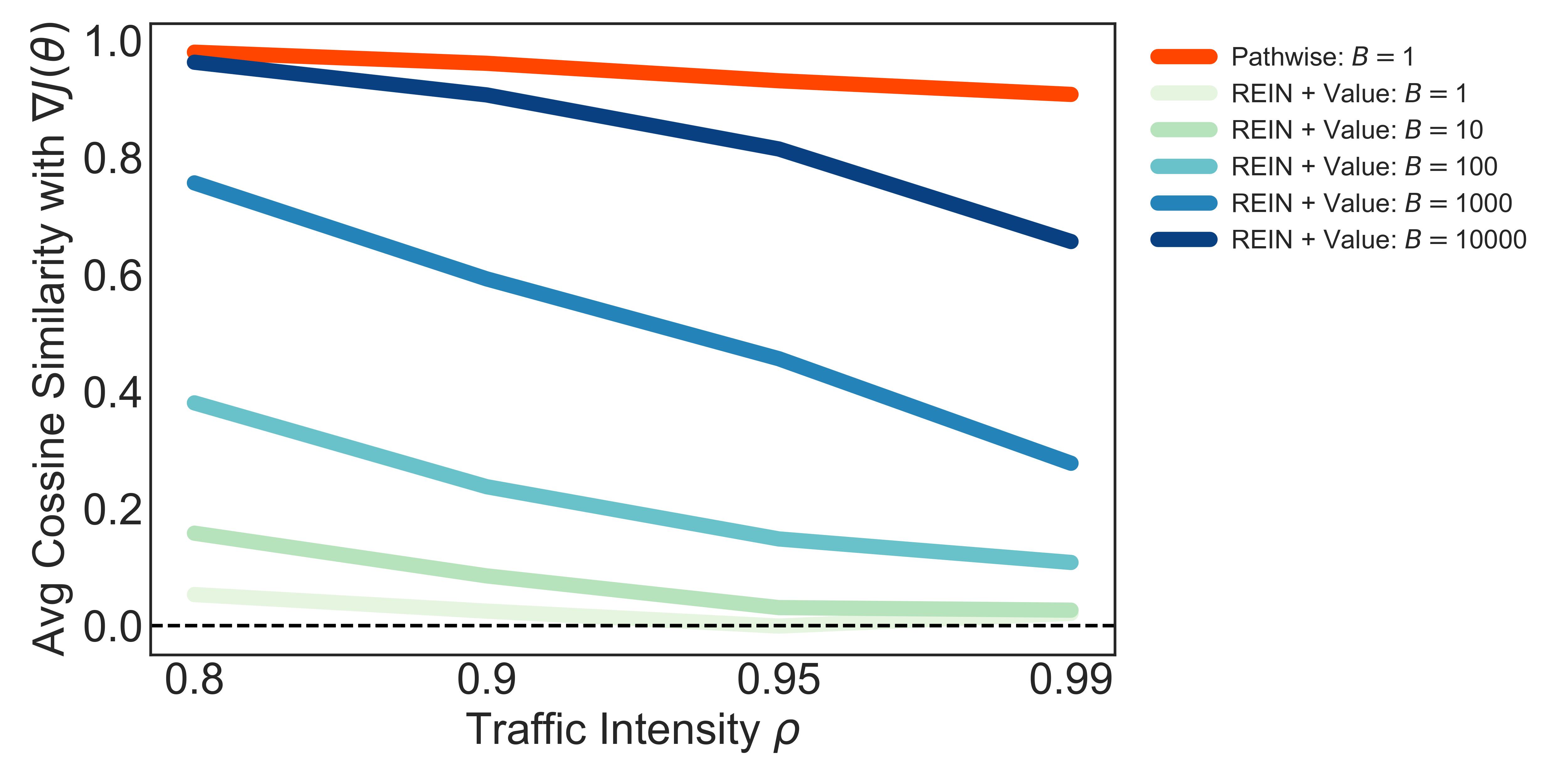}
    
    \caption{Improvements in sample efficiency of our proposed $\pathwise$ policy gradient estimator over a standard model-free RL estimator, $\reinforce$. (Left) Samples of policy gradient estimators for a parameterized MaxPressure policy in a criss-cross network with traffic intensity $\rho=0.9$ (see Example~\ref{example:criss-cross}). Each draw of the $\reinforce$ estimator is averaged over $B = 10^3$ trajectories and is equipped with a value function baseline, which is fitted using $10^6$ state transitions. The $\pathwise$ estimator uses only a single trajectory, and no value function. Despite using less data, it is more closely aligned with the true gradient. (Right) Average cosine similarity (higher is better) of policy gradient estimators with the true policy gradient (see~\eqref{eqn:similarity} for more details) across different levels of traffic intensity for the criss-cross network. For $\reinforce$, we plot the cosine similarity of the estimator under different batch sizes $B=1,..,10^{4}$. We see that the efficiency advantages of $\pathwise$, with only 1 trajectory, are greater under higher traffic intensities, even outperforming $\reinforce$ with a value function baseline and $B=10^{4}$ trajectories.}
    \label{fig:gradient_cossim_eval} 
\end{figure}

Recent advances in reinforcement learning (RL) have spurred growing interest in applying learning methodologies to solve queuing control problems, which benefit from increased data and computational resources ~\cite{dai2022queueing,walton2021learning,liu2022rl}. These algorithms hold significant potential for generating effective controls for complex, industrial-scale networks encountered in real-world applications, which typically fall outside the scope of theoretical analysis. 
However, standard model-free RL algorithms~\cite{schulman2017proximal, schulman2015trust, mnih2016asynchronous} often under-perform in queuing control, even when compared to simple queuing policies~\cite{pavse2024learning, liu2022rl}, unless proper modifications are made. This under-performance is primarily due to the unique challenges posed by queuing networks, including (1) high stochasticity of the trajectories, (2) large state and action spaces, and (3) lack of stability guarantees under sub-optimal policies~\cite{dai2022queueing}.
For example, when applying policy gradient methods, typical policy gradient estimators based on coarse feedback from the environment (observed costs) suffer prohibitive error due to high variability (see, e.g., the $\reinforce$ estimators in Figure~\ref{fig:gradient_cossim_eval}). 

To tackle the challenges in applying off-the-shelf RL solutions for queuing control, we propose a new scalable framework for policy optimization that incorporates domain-specific queuing knowledge. Our main algorithmic insight is that queueing networks possess key structural properties that allow for several orders of magnitude more accurate gradient estimation. 
By leveraging the fact that the dynamics of discrete event simulations of queuing networks are governed by observed exogenous randomness (interarrival and service times), we propose a differentiable discrete event simulation framework. This framework enables the computation of a $\pathwise$ gradient of a performance objective (e.g., cumulative holding cost) with respect to actions. 

\begin{figure}[t]
\centering
\hspace{-2em}
    \includegraphics[height = 2.4in]{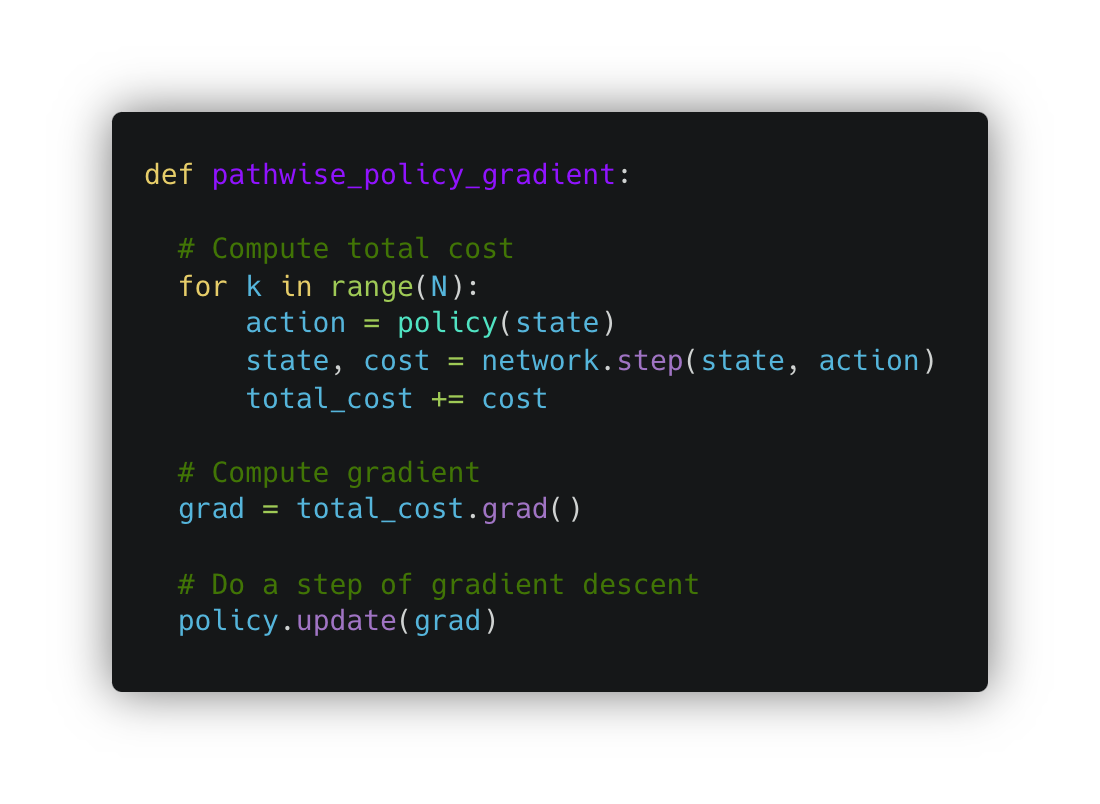}
    \includegraphics[height = 2.2in]{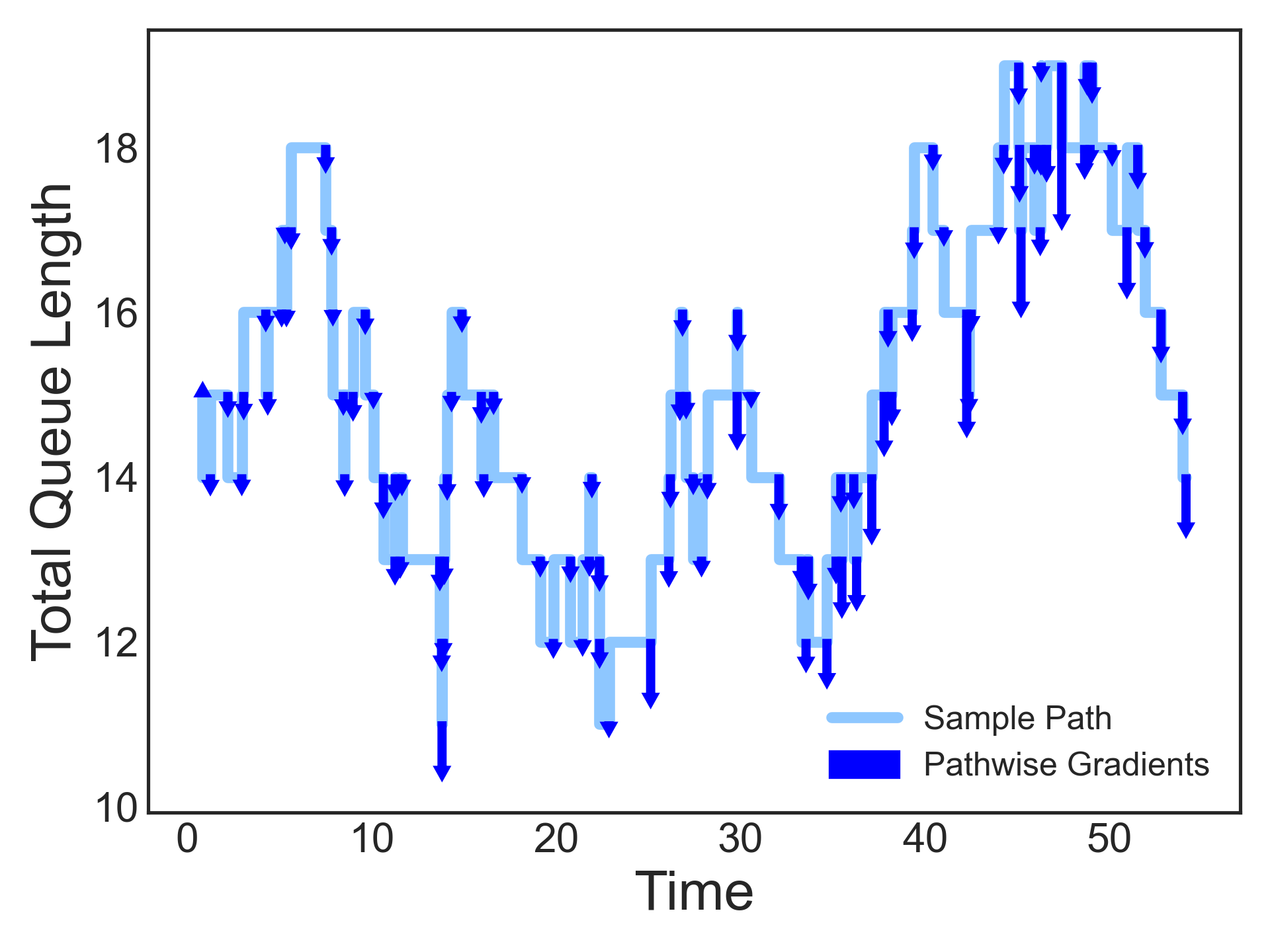}
    
\caption{(Left) Pseudo-code of a single gradient step of our proposed $\pathwise$ estimator. Computing the estimator requires only a few lines of code to compute the cost incurred by the policy. Once this cost is calculated, the sample path gradient is computed automatically via reverse-mode auto-differentiation. Unlike standard methods such as infinitesimal perturbation analysis or likelihood-ratio estimation, we can apply the same code for any network without any bespoke modifications. Unlike model-free gradient estimators like $\reinforce$, our method does not need a separate value function fitting step, managing a replay buffer, feature/return normalization, generalized advantage estimation, etc., as it has a low variance without any modification. (Right) A sample path of the total queue length (light blue) for a multi-class queuing network (see Example~\ref{example:multiclass}) under a randomized priority scheduling policy. Along the path, we display the gradients (dark blue) computed using our framework of the average cost with respect to each action produced by the policy: $\nabla_{u_{k}} \frac{1}{N}\sum_{k=0}^{N-1} c(x_{k},u_{k})\tau^{*}_{k+1}$.
    % (Right) Pseudo-code for a typical $\reinforce$ estimator is more involved, not only requiring the cost to be computed but also many additional steps, such as a value function fitting step to produce a baseline, management of a replay buffer, and normalization of states, costs, and returns which must be carefully tuned~\cite{engstrom2020implementation}. For modern policy gradient algorithms such as PPO~\cite{schulman2017proximal}, additional features are required, such as clipping of the advantages, monitoring the KL-divergence between iterates, etc. As a result, standard implementations often involve more than 10,000 lines of code, spread across multiple modules~\cite{huang2022cleanrl}.
    }
    \label{fig:code}
\end{figure}

Our proposed gradient estimator, denoted as the $\pathwise$ estimator, can then be used to efficiently optimize the parameters of a control policy through stochastic gradient descent (SGD). By utilizing the known structure of queuing network dynamics, our approach provides finer-grained feedback on the sensitivity of the performance objective to any action taken along the sample path. This offers an infinitesimal counterfactual analysis: how the performance metric would change if the scheduling action were slightly perturbed.  
Rather than relying on analytic prowess to compute these gradients, we utilize the rapid advancements in scalable auto-differentiation libraries such as PyTorch~\citep{PaszkeGrChChYaDeLiDeAnLe17} to efficiently compute gradients over a single sample path or a batch of sample paths. 
Our proposed approach supports very general control policies, including neural network policies, which have the potential to improve with more data and computational resources. 
Notably, our method seamlessly handles large-scale queuing networks and large batches of data via GPU parallelization. Unlike off-the-shelf RL solutions whose performance is exceedingly sensitive to implementation details~\cite{huang2022implementation, ilyas2018closer}, our method is easy to implement (see e.g., Figure~\ref{fig:code}) and requires minimal effort for parameter tuning.

Across a range of queueing networks, we empirically observe that our $\pathwise$ estimator substantially improves the sample efficiency and stability of learning algorithms for queuing network control while preserving the flexibility of learning approaches.  In Figure~\ref{fig:gradient_cossim_eval}, we preview our main empirical findings which show that $\pathwise$ gradients lead to a 50-1000x improvement in sample efficiency over model-free policy gradient estimators (e.g., $\reinforce$~\cite{williams1992simple}). 
%in a variety of tasks and settings. 
Buoyed by the promising empirical results, we
provide several theoretical insights explaining the observed efficiency gains.

Our proposed approach draws inspiration from gradient estimation strategies developed in the stochastic modeling and simulation literature, particularly infinitesimal perturbation analysis (IPA)~\cite{glasserman1990gradient, ho1983infinitesimal, johnson1989infinitesimal}. While IPA has been shown to provide efficient gradient estimators for specific small-scale queuing models (e.g., the $G/G/1$ queue), it is well-known that unbiased IPA estimates cannot be obtained for general multi-class queuing networks due to non-differentiability of the sample path ~\citep{cao1987first,fu1994smoothed,fu2012conditional}. Our framework overcomes this limitation by proposing a novel smoothing technique based on insights from fluid models/approximations for queues and tools from the machine learning (ML) literature. To the best of our knowledge, our method is the first to provide a gradient estimation framework capable of handling very general and large-scale queuing networks and various control policies. Our modeling approach is based on discrete-event simulation models, and as a result, it can accommodate non-stationary and non-Markovian inter-arrival and service times, requiring only samples instead of knowledge of the underlying distributions. 

Our second contribution is a simple yet powerful modification to the control policy architecture. It has been widely observed that training a standard RL algorithm, such as proximal policy optimization~\cite{schulman2017proximal} (PPO), may fail to converge due to instabilities arising from training with random initialization. To address this issue, researchers have proposed either switching to a stabilizing policy when instability occurs \citep{liu2022rl} or imitating (behavior cloning) a stabilizing policy at the beginning \citep{dai2022queueing}. However, both methods limit policy flexibility and introduce additional complexity in the training process. We identify a key source of the problem: generic policy parameterizations (e.g., neural network policies) do not enforce work conservation, leading to scenarios where even optimized policies often assign servers to empty queues. 
To address this, we propose a modification to standard  policy parameterizations in deep reinforcement learning,
% \hntodo{Most readers will not know what "standard softmax" means} 
which we refer to as the `work-conserving softmax'. This modification is compatible with standard reinforcement learning algorithms and automatically guarantees work conservation. Although work conservation does not always guarantee stability, we empirically observe across many scenarios that it effectively eliminates instability in the training process, even when starting from a randomly initialized neural network policy.
% This approach precludes the need to introduce or clone a stabilizing policy. 
% It does not impose any restrictions on the expressivity of the policy and can even reduce the number of parameters required to approximate an optimal policy. 
This modification not only complements our gradient estimator but is also compatible with other model-free RL approaches. We find that while PPO without any modifications fails to stabilize large queuing networks and leads to runaway queue lengths, PPO with the work-conserving softmax remains stable from random initialization and can learn better scheduling policies than traditional queuing policies.

Since rigorous empirical validation forms the basis of algorithmic progress, we provide a thorough empirical validation of the effectiveness of the differentiable discrete event simulator for queuing network control. We construct a wide variety of benchmark control problems, ranging from learning the $c\mu$-rule in a simple multi-class queue to scheduling and admission control in large-scale networks. Across the board, we find that our proposed $\pathwise$ gradient estimator achieves significant improvements in sample efficiency over model-free alternatives, which translate to downstream improvements in optimization performance.
\begin{itemize}[itemsep=0pt]
\item In a careful empirical study across 10,800 parameter settings, we find that for {\bf 94.5\%} of these settings our proposed $\pathwise$ gradient estimator computed along a single sample path achieves greater estimation quality than $\reinforce$ with {\bf 1000x} more data (see section~\ref{sec:gradient_efficiency}).
\item In a scheduling task in multi-class queues, gradient descent with $\pathwise$ gradient estimator better approximates the optimal policy (the $c\mu$-rule) and achieves a smaller average cost than $\reinforce$ with a value function baseline and {\bf 1000x} more data (see section~\ref{section:learn_cmu}).
\item In an admission control task, optimizing the buffer sizes with $\pathwise$ gradient estimator achieves smaller costs than randomized finite differences (SPSA~\cite{spall1992multivariate}) with {\bf 1000x} more data, particularly for higher-dimensional problem instances (see section~\ref{section:admission}).
\item For large-scale scheduling problems, policy gradient with $\pathwise$ gradient estimator and work-conserving softmax policy architecture achieves a smaller long-run average holding cost than traditional queuing policies
and state-of-the-art RL methods such as $\ppo$, which use {\bf 50x} more data (see section~\ref{section:results}). Performance gains are greater for larger networks with non-exponential noise.
\end{itemize}
These order-of-magnitude improvements in sample efficiency translate to improved computational efficiency when drawing trajectories from a simulator and improved data efficiency if samples of event times are collected from a real-world system.

Overall, these results indicate that one can achieve significant improvements in sample efficiency by incorporating the specific structure of queuing networks, which is under-utilized by model-free reinforcement learning methods. In section~\ref{section:case_study}, we investigate the $M/M/1$ queue as a theoretical case study and show that even with an optimal baseline, $\reinforce$ has a sub-optimally large variance under heavy traffic compared to a pathwise policy gradient estimator. This analysis identifies some of the statistical limitations of $\reinforce$, and illustrates that a better understanding of the transition dynamics, rather than narrowly estimating the value-function or $Q$-function, can deliver large improvements in statistical efficiency. Given the scarcity of theoretical results comparing the statistical efficiency of different policy gradient estimators, this result may be of broader interest.

Our broad aim with this work is to illustrate a new paradigm for combining the deep, structural knowledge of queuing networks developed in the stochastic modeling literature with learning and data-driven approaches.
Rather than either choosing traditional queuing policies, which can be effective for certain queueing control problems but do not improve with data, or choosing model-free reinforcement learning methods, which learn from data but do not leverage known structure, our framework offers a favorable midpoint: we leverage structural insights to extract much more informative feedback from the environment, which can nonetheless be used to optimize black-box policies and improve reliability. Beyond queuing networks, our algorithmic insight provides a general-purpose tool for computing gradients in general discrete-event dynamical systems. Considering the widespread use of discrete-event simulators with popular modeling tools such as AnyLogic~\cite{AnyLogic2024} or Simio~\cite{Simio2024} and open-source alternatives such as SimPy~\cite{matloff2008introduction}, the tools developed in this work can potentially be applied to policy optimization problems in broader industrial contexts.

The organization of this paper is as follows. In section~\ref{section:related_work}, we discuss connections with related work. In section~\ref{section:model}, we introduce the discrete-event dynamical system model for queuing networks. In section~\ref{section:gradient}, we introduce our framework for gradient estimation. In section~\ref{section:gradient_eval}, we perform a careful empirical study of our proposed gradient estimator, across estimation and optimization tasks.
In section~\ref{section:optimization}, we discuss the instability issue in queuing control problems and our proposed modification to the policy architecture to address this. In section~\ref{section:results}, we empirically investigate the performance of our proposed pathwise gradient estimation and work-conserving policy architecture in optimizing scheduling policies for large-scale networks. In section~\ref{section:case_study}, we discuss the $M/M/1$ queue as a theoretical case study concerning the statistical efficiency of $\reinforce$ compared to $\pathwise$ estimators. Finally, section~\ref{section:conclusion} concludes the paper and discusses extensions.

%{\color{blue} Emphasize that the advantages are that it is easy to implement (requires very little effort for parameter tuning leverages existing auto-differentiation tools); highly computationally efficient, and thus can be applied to solve large-scale problems; very flexible, and thus can be applied to study non-stationary and non-Markovian systems. }
%\end{itemize}
%%% Local Variables:
%%% mode: latex
%%% TeX-master: "main"
%%% End:

\section{Related Work}
\label{section:related_work}

We discuss connections to related work in queuing theory, reinforcement learning, and gradient estimation in machine learning and operations research.

\paragraph{Scheduling in Queuing Networks}

Scheduling is a long-studied control task in the queuing literature for managing queues with multiple classes of jobs ~\cite{harrison1989scheduling, meyn2008control}. Standard policies developed in the literature include static priority policies such as the $c\mu$-rule~\cite{cox1961queues}, threshold policies~\cite{rosberg1982optimal}, policies derived from fluid approximations~\cite{avram1994optimal, chen1993dynamic, meyn2001sequencing}, including discrete review policies~\cite{harrison1996bigstep, maglaras2000discrete}, policies that have good stability properties such as MaxWeight ~\cite{stolyar2004maxweight} and MaxPressure ~\cite{dai2005maximum}. Many of these policies satisfy desirable properties such as throughput optimality~\cite{tassiulas1990stability, armony2003queueing}, or cost minimization~\cite{cox1961queues, mandelbaum2004scheduling} for certain networks and/or in certain asymptotic regimes. In our work, we aim to leverage some of the theoretical insights developed in this literature to design reinforcement learning algorithms that can learn faster and with less data than model-free RL alternatives. We also use some of the standard policies as benchmark policies when validating the performance of our $\pathwise$ policy gradient algorithm.

\paragraph{Reinforcement Learning in Queueing Network Control}
Our research connects with the literature on developing reinforcement learning algorithms for queuing network control problems~\citep{moallemi2008approximate, shah2020stable, qu2020scalable, dai2022queueing, liu2022rl, wei2024sample, pavse2024learning}. These works apply standard model-free RL techniques (e.g. $Q$-learning, $\ppo$, value iteration, etc.) but introduce novel modifications to address the unique challenges in queuing network control problems. Our work differs in that we propose an entirely new methodology for learning from the environment based on differentiable discrete event simulation, which is distinct from all model-free RL methods. The works~\cite{shah2020stable,liu2022rl, dai2022queueing, pavse2024learning} observe that RL algorithms tend to be unstable and propose fixes to address this, such as introducing a Lyapunov function into the rewards, or behavior cloning of a stable policy for initialization. In our work, we propose a simple modification to the policy network architecture, denoted as the \emph{work-conserving softmax} as it is designed to ensure work-conservation. We find empirically that \emph{work-conserving softmax} ensures stability with even randomly initialized neural network policies. In our empirical experiments, we primarily compare our methodology with the $\ppo$ algorithm developed in~\cite{dai2022queueing}. In particular, we construct a $\ppo$ baseline with the same hyper-parameters, neural network architecture, and variance reduction techniques as in~\cite{dai2022queueing}, although with our policy architecture modification that improves stability. 
% It is also worth noting that these works typically benchmark their policies in environments with exponentially-distributed inter-arrival and service times, in order to use the discrete MDP representation of the queuing network. Our framework, based in discrete-event simulation, is capable of modeling general inputs and we benchmark our policies in environments with higher-variance noise.

\paragraph{Differentiable Simulation in RL and Operations Research}

While differentiable simulation is a well-studied paradigm for control problems in physics and robotics~\cite{heiden2021neuralsim, hu2019difftaichi, suh2022differentiable, howell2022dojo, schoenholz2020jax}, it has only recently been explored for large-scale operations research problems. For instance, \cite{madeka2022deep,alvo2023neural} study inventory control problems and train a neural network using direct back-propagation of the cost, as sample paths of the inventory levels are continuous and differentiable in the actions. In our work, we study control problems for queuing networks, which are discrete and non-differentiable, preventing the direct application of such methods. To address this, we develop a novel framework for computing pathwise derivatives for these non-differentiable systems, which proves highly effective for training control policies. Another line of work, including \cite{andelfinger2021differentiable, andelfinger2023towards}, proposes differentiable agent-based simulators based on differentiable relaxations. While these relaxations have shown strong performance in optimization tasks, they also introduce unpredictable discrepancies with the original dynamics. We introduce tailored differentiable relaxations in the back-propagation process only, ensuring that the forward simulation remains true to the original dynamics. 

\paragraph{Gradient Estimation in Machine Learning}

Gradient estimation~\cite{mohamed2020monte} is an important sub-field of the machine learning literature, with applications in probabilistic modeling~\cite{kingma2013auto, jang2016categorical} and reinforcement learning~\cite{williams1992simple, sutton1999policy}. There are two standard strategies for computing stochastic gradients~\cite{mohamed2020monte}. The first is the score-function estimator or $\reinforce$~\cite{williams1992simple,sutton1999policy}, which only requires the ability to compute the gradient of log-likelihood but can have high variance~\cite{greensmith2004variance}. Another strategy is the reparameterization trick~\cite{kingma2013auto}, which involves decomposing the random variable into the stochasticity and the parameter of interest, and then taking a pathwise derivative under the realization of the stochasticity. Gradient estimators based on the reparameterization trick can have much smaller variance~\cite{mohamed2020monte}, but can only be applied in special cases (e.g. Gaussian random variables) that enable this decomposition. Our methodology makes a novel observation that for queuing networks, the structure of discrete-event dynamical systems gives rise to the reparameterization trick. Nevertheless, the function of interest is non-differentiable, so standard methods cannot be applied. As a result, our framework also connects with the literature on gradient estimation for discrete random variables~\cite{jang2016categorical, maddison2016concrete,  bengio2013estimating, tucker2017rebar}. In particular, to properly smooth the non-differentiability of the event selection mechanism, we employ the straight-through trick~\cite{bengio2013estimating}, which has been previously used in applications such as discrete representation learning~\cite{van2017neural}. Our work involves a novel application of this technique for discrete-event systems, and we find that this is crucial for reducing bias when smoothing over long time horizons.

\paragraph{Gradient Estimation in Operations Research}
There is extensive literature on gradient estimation for stochastic systems~\cite{glasserman1990gradient, glasserman1992derivative, glynn1987likelilood, cao1985convergence, fu2012conditional}, some with direct application to queuing optimization ~\cite{l1994stochastic, glynn1990likelihood, fu2012conditional}.
Infinitesimal Perturbation Analysis (IPA)~\cite{glasserman1990gradient, ho1983infinitesimal, johnson1989infinitesimal} is a standard framework for constructing pathwise gradient estimators, which takes derivatives through stochastic recursions that represent the dynamics of the system. While IPA has been applied successfully to some specific queuing networks and discrete-event environments more broadly~\cite{suri1987infinitesimal}, standard IPA techniques cannot be applied to general queuing networks control problems, as has been observed in~\cite{cao1987first}. There has been much research on outlining sufficient conditions under which IPA is valid, such as the commuting condition in~\cite{glasserman1990gradient, glasserman1992derivative} or the perturbation conditions in~\cite{cao1985convergence}, but these conditions do not hold in general. Several extensions to IPA have been proposed, but these alternatives require knowing the exact characteristics of the sampling distributions and bespoke analysis of event paths~\cite{fu2012conditional, fu1994smoothed}. Generalized likelihood-ratio estimation~\cite{glynn1987likelilood} is another popular gradient estimation framework, which leverages an explicit Markovian formulation of state transitions to estimate parameter sensitivities. However, this requires knowledge of the distributions of stochastic inputs, and even with this knowledge, it may be difficult to characterize the exact Markov transition kernel of the system. Finally, finite differences~\cite{fu1997optimization} and finite perturbation analysis~\cite{ho1983infinitesimal, cao1987first} are powerful methods, particularly when aided with common random numbers~\cite{glynn1989optimization, glasserman1992some}, as it requires minimal knowledge about the system. However, it has been observed that performance can scale poorly with problem dimension~\cite{glasserman2004monte, glynn1989optimization}, and we also observe this in an admission control task (see Section~\ref{section:admission}).

Our contribution is proposing a novel, general-purpose framework for computing pathwise gradients through careful smoothing, which only requires samples of random input (e.g., interarrival times and service times) rather than knowledge of their distributions.
%and can be applied to any multi-class queuing problem under any (differentiable) scheduling policy. 
Given the negative results about the applicability of IPA for general queuing network control problems (e.g., general queuing network model and scheduling policies), we introduce bias through smoothing to achieve generality. It has been observed in~\cite{eckman2020biased} that biased IPA surrogates can be surprisingly effective in simulation optimization tasks such as ambulance base location selection. Our extensive empirical results confirm this observation and illustrate that while there is some bias, it is very small in practice, even over long time horizons ($>10^5$ steps).

% provides a novel and efficient way to compute/approximate the pathwise gradients through careful smoothing. It can be thought of as a biased IPA estimator with carefully controlled bias. It has been shown in~\cite{eckman2020biased} that despite non-differentiability issues that prevent the use of standard IPA estimators, biased IPA surrogates are surprisingly effective in simulation optimization tasks such as ambulance base location selection. In the following sections, we demonstrate through extensive numerical experiments that even though our proposed estimator is biased, the bias is very small in practice even over long time horizons ($>100,000$ steps). In addition, unlike many existing gradient estimators, our approach is general and scalable, which allows us to compare our approach to standard model-free gradient estimators like $\reinforce$ across a wide range of large-scale settings.

%\section{Multi-class Queuing Networks}
\section{Discrete-Event Dynamical System Model for Queuing Networks}
\label{section:model}

We describe multi-class queuing networks as discrete-event dynamical systems. This is different from the standard Markov chain representation, which is only applicable when inter-arrival and service times are exponentially distributed.
To accommodate more general event-time distributions, the system description not only involves the queue lengths, but also auxiliary information such as residual inter-arrival times and workloads. Surprisingly, this more detailed system description leads to a novel gradient estimation strategy (discussed in Section~\ref{section:gradient}) for policy optimization.

We first provide a brief overview of the basic scheduling problem. We then describe the discrete-event dynamics of multi-class queuing networks in detail and illustrate with a couple of well-known examples. While queuing networks have been treated as members of a more general class of Generalized Semi-Markov Processes (GSMPs) that reflect the discrete-event structure of these systems \cite{glynn1989gsmp}, we introduce a new set of notations tailored for queuing networks to elaborate on some of their special structures.
In particular, we represent the discrete event dynamics 
via matrix-vector notation that maps directly to its implementation in auto-differentiation frameworks, allowing for the differentiable simulation of large-scale queueing networks through GPU parallelization. 

\subsection{\label{subsec:control} The Scheduling Problem}

% \hntodo{This entire subsection should be given at the beginning of the section as overall context for what you're trying to do.}
A multi-class queuing network consists of $n$ queues and $m$ servers. The core state variable is the queue lengths associated with each queue, denoted as  $x(t) \in \N_{+}^{n}$, which evolves over continuous time. As a discrete-event dynamical system, the state also includes auxiliary data denoted as $\aux(t)$--- consisting of residual inter-arrival times and workloads at time $t$---which determines state transitions but are typically not visible to the controller.
% \hntodo{Most papers in queueing and applied probability reserve $o(t)$ for small-o notation. Perhaps change this?}

The goal of the controller is to route jobs to servers, represented by an assignment matrix $u \in \{0, 1 \}^{m \times n}$, to manage congestion. More concretely, the problem is to derive a policy $\pi(x)$, which only depends on the observed queue lengths and selects scheduling actions, to minimize the integral of some instantaneous costs $c(x,u)$. A typical instantaneous cost is a linear holding/waiting cost:
\[
c(x,u)=h^{\top}x
\]
% More concretely, the goal of the controller is to  $c(x,u,\xi)$
% be the instantaneous cost under state $x\in\mathcal{X}$ and action
% $u$ and sampled inter-event times.
% 
% lengths: e.g.
for some vector $h\in\R_{+}^{n}$. The objective is to find a
policy $\pi$ that minimizes the cumulative cost over a time horizon:
\begin{equation}
\min_{\pi}\mathbb{E}\left[\int_{0}^{T}c(x(t),\pi(x(t)))dt\right].\label{eq:time-average-cost}
\end{equation}
%a time-averaged cost over a fixed horizon $N$, measured in terms of the total number of events:
Optimizing a continuous time objective can be difficult and may require an expensive discretization procedure. However, discrete-event dynamical systems are more structured in that $x(t)$ is piecewise constant and is only updated when an {\bf event} occurs. For the multi-class queuing networks, events are either arrivals to the network or job completions, i.e., a server finishes processing a job.

It is then sufficient to sample the system only when an event occurs, and we can approximate the continuous-time objective with a performance objective in the discrete-event system over $N$ events,
\begin{equation}
\min_{\pi}
\left\{J_{N}(\pi) := \mathbb{E}\left[\sum_{k=0}^{N-1}c(x_{k},\pi(x_{k}))\tau_{k+1}^*\right]
\right\}
\label{eq:des-cost}
\end{equation}
where $x_k$ is the queue lengths after the $k$th event update. $\tau^{*}_{k+1}$ is an inter-event time that measures the time between the $k$th and $(k+1)$th event, and $N$ is chosen such that the time of the $N$th event, a random variable denoted as $t_{N}$, is ``close" to $T$.
%For this objective to well-approximate the continuous time counterpart, one would need to select $N$ large enough to so that the time of the $N$th event, a random variable denoted as $t_{N}$, is close to $T$.
% We multiply the cost after event $k$ by the inter-event epoch $\tau_{k+1}$
% %in order to properly average the cost by the duration. As a result, this will be 
% so that \eqref{eq:des-cost} is equivalent to the following continuous time objective for $T=t_{N}^*$,

The dynamics of queuing networks are highly stochastic, with large variations across trajectories. Randomness in the system is driven by the random arrival times of jobs and the random workloads (service requirements) of these jobs.  We let $\xi_{1:N}=\{\xi_{i}\}_{i=1}^{N}$ denote a single realization, or `trace', of these random variables over the horizon of $N$ events.
% Due to the exogenous structure of the discrete event system, we can % compute the cost of a policy under a single trace $\xi_{1:N}=\{\xi_{i}\}_{i=1}^{N}$;
We can then view the expected cost~\eqref{eq:des-cost} more explicitly as a policy cost averaged over traces. 
In addition, we focus on a parameterized family of policies $\{\pi_{\theta}:\theta\in\Theta\}$, for some $\Theta\subseteq\R^{d}$, in order to optimize~\eqref{eq:des-cost} efficiently. In this case, we utilize the following shorthand $J_N(\theta;\xi_{1:N})$ for the
policy cost over a single trace and $J_N(\theta)$ for the average policy
cost under $\pi_{\theta}$,
which leads to the parameterized control problem:
\begin{equation}
\min_{\theta}
\left\{J_{N}(\theta) := \mathbb{E}\left[J(\theta;\xi_{1:N})\right]
:= 
\mathbb{E}\left[\sum_{k=0}^{N-1}c(x_{k},\pi_{\theta}(x_{k}))\tau^{*}_{k+1}\right]
\right\}.
\label{eq:param-cost}
\end{equation}
We now turn to describe the structure of the transition dynamics of multi-class queuing networks, to elaborate how scheduling actions affect the queue lengths.

% Under exponentially distributed inter-arrival and service times, we
% can oftentimes represent the dynamics of the queuing network as a relatively simple discrete-time Markov chain (DTMC) via uniformization and proper truncation.\hntodo{CITE} To handle non-exponentially distributed system inputs, we also need to keep track of the residual inter-arrival and service times to have a Markovian system descriptor. On the other hand, discrete-event simulation is a standard modeling approach that can accommodate general system inputs. We find that its structure enables novel gradient estimation strategies (discussed in Section~\ref{section:gradient}) for policy optimization.

\subsection{System Description}
Recall that the multi-class queuing network consists of $n$ queues and $m$ servers, where each queue is associated with a job class, and different servers can be of different compatibilities with various job classes. Recall that $x(t)\in\mathbb{N}_{+}^{n}$ denotes the lengths of the queues at time $t \in \R_{+}$.
The queue lengths $x(t)$ are updated by one of two types of events: job arrivals or job completions. 
Although the process evolves in continuous time, it is sufficient to track the system only when an event occurs. We let $k\in \N_{+}$ count the $k$th event in the system, and let $t_{k}$ denote the time immediately after the $k$th event occurs. By doing so, we arrive at a discrete-time representation of the system. Given that we do not assume event times are exponential, the queue lengths $x_{k}$ alone are not a Markovian descriptor of the system. Instead, we must consider an  {\bf augmented state} $s_{k} = (x_{k}, \aux_{k})$, where $x_k \in \N^{n}_{+}$ is the vector of {\bf queue lengths} and $\aux_{k} = (\tau_{k}^{A}, w_{k}),\in \R^{2n}_{+}$ is an auxiliary state vector that includes {\bf residual inter-arrival times} $\tau_{k}^{A} = \{ \arrival \}_{j = 1}^{n} \in \R^{n}_{+}$ and {\bf residual workloads} $w_{k} = \{ \service \}_{j = 1}^{n} \in \R^{n}_{+}$ of the `top-of-queue' jobs in each queue. The auxiliary state variables determine the sequence of events.

More explicitly, for each queue $j\in[n]$, the residual inter-arrival time $\arrival$ keeps track of the time remaining until the next arrival to queue $j$ occurs. Immediately after an arrival to queue $j$ occurs, the next inter-arrival time is drawn from a probability distribution $F_{j}^{A}$. When a job arrives to queue $j$, it comes with a workload (service requirement) drawn from a distribution $F_{j}^{S}$. We allow the distributions $F_{j}^{A}$'s and $F_{j}^{S}$'s to vary with time, i.e., the interarrival times and service requirements can be time-varying. For notational simplicity, we will not explicitly denote the time dependence here. We refer to the residual workload at time $t_{k}$ of the top-of-queue job in queue $j$ as $\service$, which specifies how much work must be done before the job completion. A job is only processed if it is routed to a server $i\in[m]$, in which case the server processes the job at a constant service rate $\mu_{ij} \in \R_{+}$. We refer to $\mu \in \R_{+}^{m \times n}$  as the matrix of service rates.
Under this scheduling decision, the \textbf{residual processing time}, i.e., the amount of time required
to process the job, is $\tau^{S}_{k,j}= \service/\mu_{ij}$. 
% We emphasize that it is the processing time \emph{not} the raw workload $\service$ that determines when the job is processed, and this is directly influenced by the actions taken by the controller.
% We
% denote $\mu\in\mathbb{R}_{+}^{m\times n}$ as the matrix of service rates where $\mu_{ij}$ is the service rate for server $i$ serving job class $j$.

% Although the process evolves in continuous time, it is sufficient to track the system only when an event occurs. Doing so, we arrive at a discrete-time representation of the system: $s_{k} = (x_{k}, \tau_{k})$, where $x_k \in \N^{n}_{+}$ is the vector queue lengths and $\tau_{k} \in \R^{2n}_{+}$ is the vector of {\bf residual event times } (i.e., interarrival and service times) immediately after the $k$th event occurs. Unlike the Markovian representation of queuing networks under exponential event times, our description is more general and includes the residual event times as state variables, even though typical queuing policies do not explicitly use this information.

The augmented state $s_{k}$ is a valid Markovian descriptor of the system and we now describe the corresponding  transition function $f$ such that
\[
s_{k+1} = f(s_{k}, u_{k}, \xi_{k+1}),
\]
where $u_{k}$ is an action taken by the controller and $\xi_{k+1}$ contains external randomness arising from new inter-arrival times or workloads drawn from $F^{A}_{j}$'s or $F^{S}_{j}$'s depending on the event type.

The transition is based on the next event, which is the event with the minimum residual time. The controller influences the transitions through the processing times, by deciding which jobs get routed to which servers. We focus on scheduling problems where the space of controls $\mathcal{U}$ are feasible assignments of servers to queues. Let $\ones_{n}\in \R^{n}$ denote an $n$-dimensional vector consisting of all ones. The action space is, 
\begin{equation}
\label{eq:action_space}
\mathcal{U}:=\left\{ u\in\{0,1\}^{m\times n}:u\ones_{n} = \ones_{m}, \ones_{m}^{\top} u=\ones_{n},u\leq M\right\},
\end{equation}
where $M\in\{0,1\}^{m\times n}$ is the {\bf topology} of the network, which indicates which job class can be served by which server. Following existing works on scheduling in queuing networks~\cite{meyn2008control}, we consider networks for which each job class has exactly 1 compatible server.
\begin{assumption} \label{ass:queue}
%(Scheduling) 
For every queue $j$, there is 1 compatible server,
i.e., $\sum_{i=1}^{m}M_{ij}=1$.
\end{assumption}
% We restrict our focus to this class of queues so that it is sufficient to track the service requirements of the top-of-the-queue jobs. Consequently, we do not need to make any assumptions about the order in which jobs within each queue are served (e.g., first-in-first-out (FIFO) or last-in-first-out (LIFO)), as long as the order is not state-dependent.

\begin{figure}[t]
\hspace{-1em}
    \includegraphics[height = 1.7in]{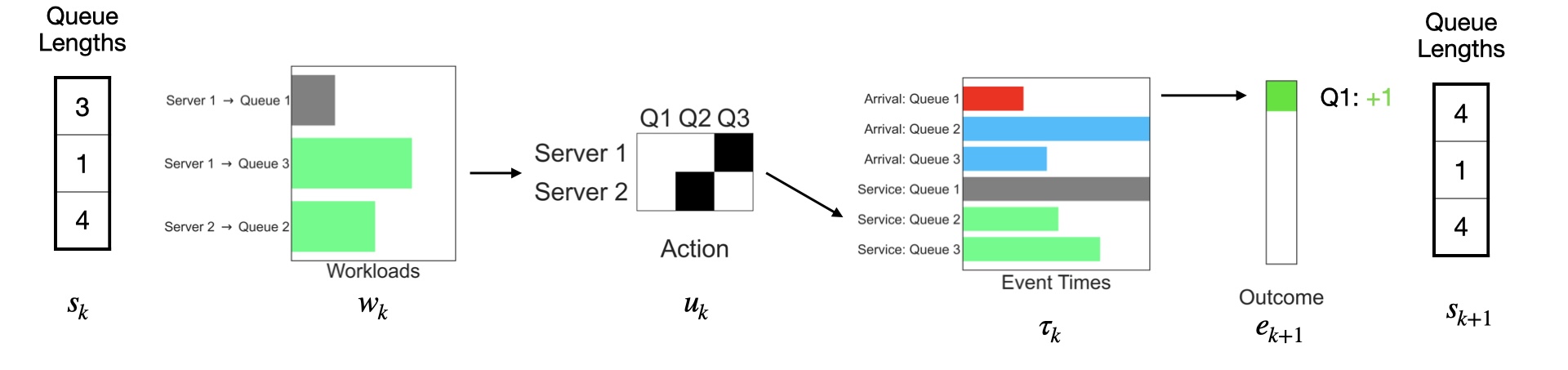}
    
    \caption{One step of the dynamics for the criss-cross network (see Example~\ref{example:criss-cross} and Figure~\ref{fig:networks}). There are 3 queues and 2 servers. Beginning with queue-lengths $x_{k} = (3, 1, 4)$ and workloads $w_{k}$, the action $u_{k}$ assigns server 1 to queue 3 and server 2 to queue 2. The workloads of the selected queues are highlighted in light green. As a result, the valid events are arrivals to queue 1 and queue 3 (queue 2 has no external arrivals) and job completions for queue 2 and queue 3 (queue 1 cannot experience any job completions because no server is assigned). The arrival event to queue 1 has the minimum residual time (highlighted in red) so it is the next event, and $e_{k+1}$ is a one-hot vector indicating this. Since an arrival occurred, the  queue-lengths are updated as $x_{k+1} = (4, 1, 4)$.}
    \label{fig:one_step}
\end{figure}

Given an action $u$, the residual processing time is $\service/\mu_{ij}$ when $u_{ij}=1$ and $\infty$ 
when $u_{ij}=0$. This can be written compactly as 
% \begin{align*}
% \phi_{A_{j}}(\tau^{A_{j}},u) & =\tau_{k,i}^{A}\\
% \phi_{S_{j}}(\tau^{S_{j}},u) & =\frac{\service}{u_{j}^{\top}\mu_{j}}:=\frac{\service}{\sum_{j=1}^{m}u_{ij}\mu_{ij}}
% \end{align*}
\begin{equation}
\label{eqn:processing_time}
\tau^{S}_{k,j} \equiv \frac{\service}{\sum_{i=1}^{m}u_{ij}\mu_{ij}} = \frac{\service}{\servicemap_{j,j}},
\end{equation}
where $\mathbf{diag}(u^{\top} \mu) \in \R^{n \times n}$ extracts the diagonal entries of the matrix $u^{\top} \mu \in \R^{n \times n}$.

As a result, at time $t_{k}$ the {\bf residual event times} $\tau_{k} \in \R_{+}^{2n}$ consists of the residual inter-arrival and processing times,
% We let $\servicemap_{u} \in \R^{2n \times 2n}_{+}$ be a (diagonal) matrix that converts the residual service requirements $\tau_{k}$ to the residual processing times (while keeping the inter-arrival times the same),
% \[
% V_{u_{k}}\tau_{k} = \left(
% \tau_{k}^{A_{1}},...,\tau_{k}^{A_{n}},
% (u_{1}^{\top}\mu_{1})^{-1} \tau_{k}^{S_{1}},...,(u_{n}^{\top}\mu_{n})^{-1} \tau_{k}^{S_{n}}
% \right) \in \R^{2n}_{+}
% \]
\[
\tau_{k} \equiv (\tau_{k}^{A}, \tau_{k}^{S}) = (\tau_{k}^{A}, \servicemapk^{-1} \servicek )
\]
We emphasize that $\tau_{k}$ depends on the action $u$. The core operation in the transition dynamics is the {\bf event selection} mechanism. The next event is the one with the minimum residual time in $\tau_{k}$.
%(arrival time or processing time). 
We define $\event_{k+1} \in \{0,1\}^{2n}$ to be a one-hot vector representing the $\argmin$ of $\tau_{k}$ -- the position of the minimum in $\tau_{k}$:
\begin{align*}
\event_{k+1}(\aux_{k}, u_{k}) &\equiv \argmin (\tau_{k})  \in \{0,1\}^{2n}\tag{Event Select}
%& = \argmin \{ \tau_{k}^{A} , \servicemapk^{-1} \servicek \} 
%\in \{0,1\}^{2n}
\end{align*}
$\event_{k+1}(\aux_{k}, u_{k})$ indicates the type of the $(k+1)$th event.
In particular, if the minimum residual event time is a residual inter-arrival time, then the next event is an arrival to the system. If it is a residual job processing time, then the next event is a job completion.
We denote $\tau^{*}_{k+1}$ to be the {\bf inter-event time}, which is equal to the minimum residual time:
\begin{align}
\tau^{*}_{k+1}(\aux_{k}, u_{k}) 
= \min \{ \tau_{k} \}
\tag{Event Time}
\end{align}
$\tau^{*}_{k+1}(\aux_{k}, u_{k})$ is the time between the $k$th and $(k+1)$th event, i.e. $t_{k+1} - t_{k}$.

After the job is processed by a server, it either leaves the system or proceeds
to another queue. Let $R\in\mathbb{R}^{n\times n}$ denote the {\bf routing matrix}, where the jth column, $R_{j}$ details the change in the queue lengths when a job in class $j$ finishes service. For example, for a tandem queue with two queues, the routing
matrix is
\[
R=\left[\begin{array}{cc}
-1 & 0\\
1 & -1
\end{array}\right]
\]
indicating that when a job in the first queue completes service, it leaves its
own queue and joins the second queue. When a job in the second queue
completes service, it leaves the system. 
%We can incorporate probabilistic routing through random routing matrices.
%We can then write the update to the queue-lengths as involving 

We define the {\bf event matrix} $D$ as a  block matrix of the form
\[
D=[\begin{array}{cc}
I_{n} & R\end{array}],
\]
where $I_n$ is the $n\times n$ identity matrix. The event matrix determines the update to the queue lengths, depending on which event took place. In particular, when the $(k+1)$th event occurs, the update to the queue lengths is
\begin{align}
x_{k+1} &=x_{k}+D \event_{k+1}(\aux_{k}, u_{k})
\tag{Queue Update} 
\end{align}
Intuitively, the queue length of queue $j$ increases by $1$ when the next event is a class $j$ job arrival; the queue lengths update according to $R_{j}$ when the next event is a queue $j$ job completion.

The updates to the auxiliary state $\aux_{k} = (\tau_{k}^{A}, w_{k}) \in \R_{+}^{2n}$ is typically given by
% \begin{align}
% \tau_{k+1}^{A} &= \tau_{k}^{A} - \tau_{k+1}^{*}\ones_{n} + T e_{k+1},
% \quad T_{j} \stackrel{iid}{\sim} F_{j}^{A} \\
% w_{k+1} &= \servicek - \tau_{k+1}^{*} \mathbf{diag}(u^\top \mu) + We_{k+1},
% \quad W_{j} \stackrel{iid}{\sim} F_{j}^{A}
% \end{align}
\begin{align}\label{eq:aux_update}
\left[
\begin{array}{c}
\tau_{k+1}^{A} \\
w_{k+1}
\end{array}
\right]
&= 
\left[
\begin{array}{c}
\tau_{k}^{A} \\
w_{k}
\end{array}
\right]
- 
\underbrace{\tau_{k+1}^{*}\left[
\begin{array}{c}
\ones_{n} \\
\servicemapk
\end{array}
\right]
}_{\text{reduce residual times}} + 
\underbrace{
\left[
\begin{array}{c}
T_{k+1} \\
W_{k+1}
\end{array}
\right] \odot e_{k+1}
}_{\text{draw new times / workloads}}
\hspace{-5em}
\tag{Aux Update} 
\end{align}
where $\odot$ is the element-wise product and
$T_{k+1} =\{T_{(k+1),j}\}_{j=1}^{n} \in \R^{n}$ are new inter-arrival times $T_{(k+1),j}  \sim F_{j}^{A}$ %if an arrival occurs. 
and $W_{k+1} = \{W_{k+1,j}\}_{j=1}^{n}\in \R^{n}$ are workloads $W_{k+1,j}  \sim F_{j}^{S}$.
%If the next event is a class $j$ job arrival occurs, we take a new inter-arrival time $T_{(k+1),j}$ from $T_{k+1}$ to update $\tau_{k+1}^A$. If the next event is a new queue $j$ job completion and queue $j$ is not empty, we take a new workload $W_{(k+1),j}$ from $W_{k+1}$ to update $w_{k+1,j}$. 
Intuitively, after an event occurs, we reduce the residual inter-arrival times by the inter-event time. We reduce workloads by the amount of work applied to the job, i.e., the inter-event time multiplied by the service rate of the allocated server. Finally, if an arrival occurred we draw a new inter-arrival time; if a job was completed, we draw a new workload for the top-of-queue job (if the queue is non-empty).
\begin{wrapfigure}{r}{0.4\textwidth}
  \begin{center}
  \includegraphics[width=0.3\textwidth]{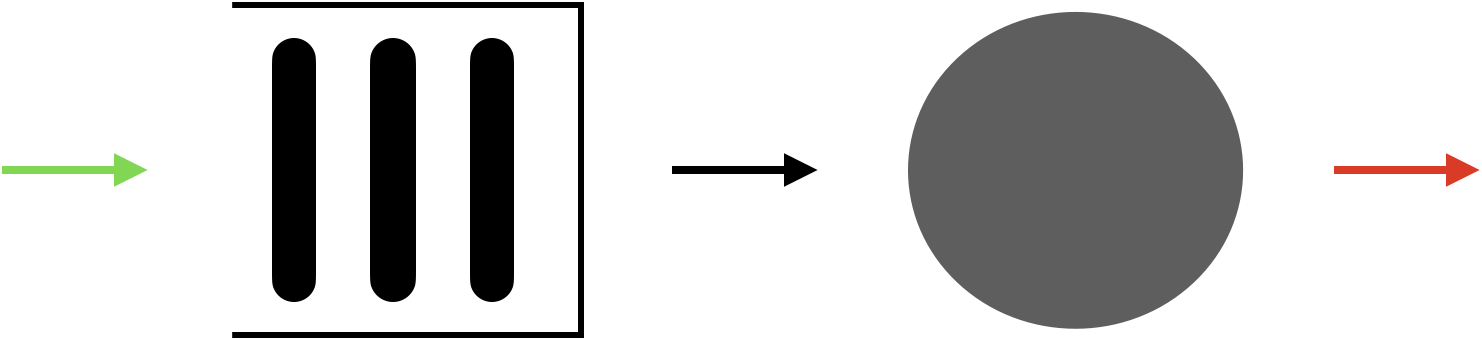}
  \end{center}
  \caption{$M/M/1$ queue.} \label{fig:MM1}
\end{wrapfigure}

There are two boundary cases that make the update slightly different from \eqref{eq:aux_update}. First, if a new job arrives at an empty queue $j$ (either an external arrival or a transition from a job completion), we also need to update $w_{k+1,j}$ to $W_{k+1,j}$. Second, if a queue $j$ job completion leaves an empty queue behind, we set $w_{k+1, j} = \infty$, indicating that no completions can occur for an empty queue.
%If the queue is emptied, then the workload is $W_{k} = \infty$, indicating that no completions can occur for an empty queue. 

Let $\xi_{k}$ denote exogenous noise in the environment, which consists of the sampled inter-arrival times and workloads for resetting the time of a completed event,
\[
\xi_{k} = (T_{k}, W_{k}) \in \R_{+}^{2n}.
\]
We finally arrive at the stated goal of describing the transition dynamics of $s_{k} = (x_{k}, \aux_{k})$ in terms of a function $f(s_{k}, u_{k}, \xi_{k+1})$. Notably, all the stochasticity is captured by $\xi_k$'s, which are independent of the states and actions. 

It is worth mentioning a few features of this discrete-event representation.
\begin{itemize}[itemsep=0pt]
    \item While auxiliary data $\aux_{k} = (\tau^{A}_{k}, w_{k})$ is necessary for $s_{k} = (x_{k}, \aux_{k})$ to be a valid Markovian system descriptor, this information is typically not available to the controller. We assume the controller only observes the queue lengths, i.e., the control policy $\pi$ only depends on $x_k$.
    \item The representation can flexibly accommodate non-stationary and non-exponential event-time distributions, i.e., $F_{j}^{A}$'s and $F_{j}^{S}$'s can be general and time-varying.
    \item This model enables purely data-driven simulation, as it only requires samples of the event times $\xi_{k}$. One does not need to know the event time distributions $F_{j}^{A}$'s and $F_{j}^{S}$'s to simulate the system if data of these event times are available.
    \item The matrix-vector representation enables GPU parallelism, which can greatly speed up the simulation of large-scale networks. 
    %It also enables efficient evaluation of several trajectories in parallel.
    \item As we will explain later, this representation enables new gradient estimation strategies.
\end{itemize}

\paragraph{Queuing Network Examples}
As a concrete illustration, we show how a few well-known queuing networks are described as discrete-event dynamical systems.
% \hntodo{Split the figure and put it in each example via wrapfigure}
\vspace{1em}

\begin{example}
\label{example:mm1}
The $M/M/1$ queue (see Figure \ref{fig:MM1}) with arrival rate $\lambda>0$ and
service rate $\mu\geq\lambda$ features a single queue $n=1$ and a single server $m=1$, and exponentially distributed inter-arrival times and workloads, i.e., $T_k \sim \mathsf{Exp}\left(\lambda\right)$ and
$W_k \sim \mathsf{Exp}(1)$ respectively. The network topology is $M=[1]$, the service rate is $\mu$, and the routing matrix is $R =[-1]$, indicating that jobs leave the system after service completion.
The scheduling policy is work-conserving, the server always serves the queue when it is non-empty, i.e. $u_{k}=1\{x_{k}>0\}$. The state update is,
\begin{align*}
 x_{k+1}  &=x_{k}+\left[\begin{array}{cc}
1 & -1
\end{array}\right]^{\top}\event_{k+1}\\
\event_{k+1} 
&=\arg\min\left\{ \tau_{k}^{A},\tau_{k}^{S} \right\} =\arg\min\left\{ \tau_{k}^{A},\frac{\servicek}{\mu \cdot 1\{x_{k}>0 \}} \right\} \in \{0,1\}^{2} \\
\tau^{*}_{k+1} &= \min\left\{ \tau_{k}^{A},\tau_{k}^{S} \right\} \\
\left[
\begin{array}{c}
\tau_{k+1}^{A} \\
w_{k+1}
\end{array}
\right]
&= 
\left[
\begin{array}{c}
\tau_{k}^{A} \\
w_{k}
\end{array}
\right]
- \tau_{k+1}^{*}\left[
\begin{array}{c}
1 \\
\mu 1\{ x_{k} > 0\}
\end{array}
\right] +
\left[
\begin{array}{c}
T_{k+1} \\
W_{k+1}
\end{array}
\right] \odot e_{k+1}.
\end{align*}
%The event matrix is  $D=[1,-1]$, indicating that the queue increases by 1 when an arrival occurs and decreases by 1 when a job completion occurs. 
%Note that since $(T_{k}, W_{k})$ are independent exponential random variables, $(\tau_{k}^{A}, w_{k})$ are as well for all $k$.
\end{example}

\begin{wrapfigure}{L}{0.3\textwidth}
  \begin{center}
  \includegraphics[width=0.2\textwidth]{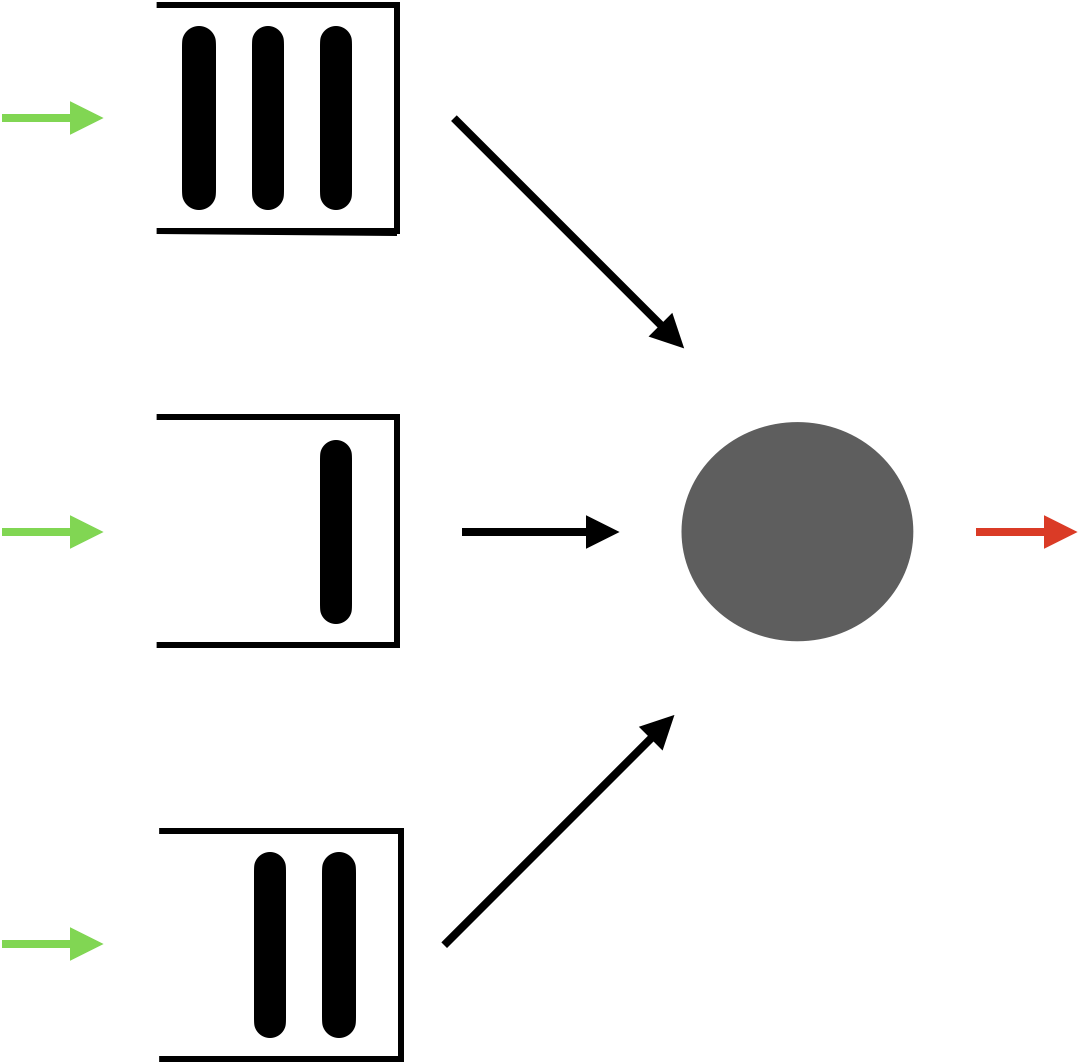}
  \end{center}
  \caption{Multi-class, single-server queue.} \label{fig:multi-class}
\end{wrapfigure}

\begin{example}
\label{example:multiclass}
The {\bf multi-class singer-server queue} features an $n$ queues and a single server $m=1$ (see Figure \ref{fig:multi-class}). While the inter-arrival times and workloads, i.e., $(T_{k}, W_{k})$'s are usually exponentially distributed, they can also follow other distributions. The network topology is $M=[1,...,1] \in \R^{n}$, the service rates are $\mu = [\mu_{1},...,\mu_{n}]$, and the routing matrix is $R =[-1,...,-1]\in \R^{n}$, indicating that jobs leave the system after service completion. 
A well-known scheduling policy for this system is the $c\mu$-rule, a static priority rule.
%based on the holding costs and service rate and is known to be optimal for this network. 
Let $h = (h_{1},...,h_{n}) \in \R^{n}$ denote the holding costs. The $c\mu$-rule sets
\[
    u_{k}=\argmax_{j\in [n]} \{ h_{j}\mu_{j}1\{x_{j}>0\} \}\in \{0,1\}^{n}.
\]
%where $\argmax$ maps a vector into a one-hot vector indicating the index of the arg max. 
The state update is,
\begin{align*}
 x_{k+1}  &=x_{k}+\left[\begin{array}{cc}
\ones_{n} & -\ones_{n}
\end{array}\right]^{\top}\event_{k+1}\\
\event_{k+1} 
&=\arg\min\left\{ \tau_{k,1}^{A},...,\tau_{k,n}^{A},\tau_{k,1}^{S},...,\tau_{k,n}^{S} \right\} 
=\arg\min\left\{ \tau_{k,1}^{A},...,\tau_{k,n}^{A},
\frac{w_{k,1}}{\mu_{1} u_{k,1}},...
\frac{w_{k,n}}{\mu_{n} u_{k,n}}\right\} \\
\tau^{*}_{k+1} &= \min\left\{ \tau_{k,1}^{A},...,\tau_{k,n}^{A},\tau_{k,1}^{S},...,\tau_{k,n}^{S} \right\} \\
\left[
\begin{array}{c}
\tau_{k+1}^{A} \\
w_{k+1}
\end{array}
\right]
&= 
\left[
\begin{array}{c}
\tau_{k}^{A} \\
w_{k}
\end{array}
\right]
- \tau_{k+1}^{*}\left[
\begin{array}{c}
\ones_{n} \\
\mu \odot u_{k}
\end{array}
\right] +
\left[
\begin{array}{c}
T_{k+1} \\
W_{k+1}
\end{array}
\right] \odot e_{k+1}.
\end{align*}
%where $\odot$ is element-wise multiplication. 
\end{example}

\begin{example}
\label{example:criss-cross}
The {\bf criss-cross network}~\cite{harrison1990scheduling} features $n=3$ queues and $m=2$ servers (see Figure~\ref{fig:networks}). External jobs arrive to queues 1 and 3. The first server can serve queues 1 and 3 with service rates $\mu_{11}$ and $\mu_{13}$ respectively, while the second server is dedicated to serving queue 2 with service rate $\mu_{22}$. After jobs from queue 1 are processed, they are routed to queue 2; jobs from queues 2 and 3 exit the system after service completion. The inter-arrival times and workloads, i.e., $(T_{k}, W_{k})$'s, can follow general distributions.
The network topology $M$, service rate matrix, and the routing matrix $R$ are:
\[
M = \left[ \begin{array}{ccc}
1 & 0 & 1\\
0 & 1 & 0 \\
\end{array} \right], \qquad
\mu = \left[ \begin{array}{ccc}
\mu_{11} & 0 & \mu_{13}\\
0 & \mu_{22} & 0 \\
\end{array} \right], \qquad
R = \left[ \begin{array}{ccc}
-1 & 0 & 0\\
1 & -1 & 0 \\
0 & 0 & -1
\end{array} \right]
\]
%At every step $k$, the action $u_{k}\in \{0,1\}^{m \times n}$ is constrained so that $u_{k,11} + u_{k,13}\leq 1$, $u_{k,22}\leq 1$. 
\citet{harrison1990scheduling} develop a work-conserving threshold policy for this system. For a threshold $a \in \N_{+}$, server 1 prioritizes jobs in queue 1 if the number of jobs in queue 2 is below $a$. Otherwise, it prioritizes queue 3. This gives the scheduling action
\[
u_{k,11} = 1\{x_{k,2} \leq a \}, \quad
u_{k,22} = 1\{x_{k,2} > 0 \}, \quad 
u_{k,13} = (1 - u_{k,11})1\{x_{k,3} > 0\},
\]
and the transition dynamics 
\begin{align*}
 x_{k+1}  &=x_{k}+\left[\begin{array}{cc}
I_{3} & R
\end{array}\right]
\event_{k+1}\\
\event_{k+1} &=\arg\min\left\{ 
\tau_{k,1}^{A}, \infty, \tau_{k,3}^{A}, 
\tau_{k,1}^{S},
\tau_{k,2}^{S},
\tau_{k,3}^{S}
\right\}  
=\arg\min\left\{ 
\tau_{k,1}^{A}, \infty, \tau_{k,3}^{A}, 
\frac{w_{k,1}}{\mu_{11}u_{k,11}},
\frac{w_{k,2}}{\mu_{22}u_{k,22}}, 
\frac{w_{k,3}}{\mu_{13}u_{k,13}}\right\}  \\
\tau^{*}_{k+1} &= \min\left\{ 
\tau_{k,1}^{A}, \infty, \tau_{k,3}^{A}, 
\tau_{k,1}^{S},
\tau_{k,2}^{S},
\tau_{k,3}^{S}\right\} \\
\left[
\begin{array}{c}
\tau_{k+1}^{A} \\
w_{k+1}
\end{array}
\right]
&= 
\left[
\begin{array}{c}
\tau_{k}^{A} \\
w_{k}
\end{array}
\right]
- \tau_{k+1}^{*}\left[
\begin{array}{c}
\ones_{3} \\
\servicemapk
\end{array}
\right] +
\left[
\begin{array}{c}
T_{k+1} \\
W_{k+1}
\end{array}
\right] \odot e_{k+1}.
\end{align*}
Here, $\tau_{k,2}^{A} = \infty$ since queue 2 has no external arrivals. 
%While usually $(T_{k}, W_{k})$ are independent exponential random variables, this need not be the case.
%Only jobs from queue 1 that have finished processing are routed to queue 2.
% Jobs leave the system after processing so $R=-1$, and the event matrix is  $D=[1,-1]$. 
\end{example}

\section{Gradient Estimation}
\label{section:gradient}
% \hntodo{Provide more context on the main results to expect in this section}

% \hntodo{I think we should explicitly highlight early on, in previous sections (maybe intro) that our smoothing approach is able to compute accurate gradient estimators even in long horizons, and list the extensive nature of our upcoming empirical validation (number policies, envs etc etc).}
In this section, we introduce our proposed approach for estimating the gradient of the objective \eqref{eq:param-cost}, $\nabla J_{N}(\theta)$. We start with a brief discussion of existing methods for gradient estimation, including their advantages and limitations. We then outline the main challenges for computing pathwise derivatives in multi-class queuing networks, and introduce our strategy for overcoming these challenges. Finally, we formally define our gradient estimation framework and discuss its computational and statistical properties. Later in section~\ref{section:gradient_eval}, we perform a comprehensive empirical study and find that our gradient estimation framework is able to overcome many of the limitations of existing methods in that (1) it is capable of estimating gradients for general queuing networks, (2) it provides stable gradient estimations over very long horizons ($>10^5$ steps), (3) it provides greater estimation accuracy than model-free policy gradient methods with 1000x less data, %across 10,800 parameter instances, 
and (4) when applying to policy optimization, it drastically improves the performance of the policy gradient algorithm for various scheduling and admission control tasks. 

Our goal is to optimize the parameterized control problem (\ref{eq:param-cost}). A standard optimization algorithm is (stochastic) gradient descent, which has been considered for policy optimization and reinforcement learning~\cite{sutton1999policy, baird1998gradient}. The core challenge for estimating policy gradient $\nabla J_{N}(\theta) = \nabla\E\left[J(\theta;\xi_{1:N})\right]$ from sample paths of the queuing network is that the sample path cost $J(\theta, \xi_{1:N})$ is in general not differentiable in $\theta$. As a consequence, one cannot change the order of differentiation and expectation, i.e.,
\[
\nabla J_{N}(\theta)=\nabla\E\left[J_{N}(\theta;\xi_{1:N})\right]\ne\mathbb{E}\left[\nabla J_{N}(\theta;\xi_{1:N})\right],
\]
where $\nabla J_{N}(\theta;\xi_{1:N})$ is not even well-defined.
The non-differentiability of these discrete-event dynamical systems emerges from two sources. First, actions $u_{1:N}$ are discrete scheduling decisions, and small perturbations in the policy can result in large changes in the scheduling decisions produced by the policy. Second, the actions affect the dynamics through the event times. The ordering of events is based on the `$\mathsf{argmin}$' of the residual event times, which is not differentiable.
%This makes it difficult to estimate $\nabla J_{N}(\theta)$ from sample paths.

% \hntodo{We need a 1-2 line math introduction to IPA and LR here. There is no context on what these methods actually do.}
In the stochastic simulation literature, there are two popular methods for gradient estimation: {\bf infinitesimal perturbation analysis} (IPA) and generalized {\bf likelihood ratio} (LR) gradient estimation.
To illustrate, consider abstractly and with a little abuse of notation a system following the dynamics $s_{k+1} = f(s_{k}, \theta, \xi_{k+1})$, where $s_k \in \R$ is the state, $\theta \in \R$ is the parameter of interest,  $\xi_{k}$ is exogenous stochastic noise, and $f$ is a differentiable function. Then, the IPA estimator computes a sample-path derivative estimator
by constructing a derivative process $D_{k}=\partial s_{k} / \partial\theta$ via the recursion:
\[
    D_{k+1} = \frac{\partial}{\partial \theta} f(s_{k}, \theta, \xi_{k+1}) + \frac{\partial}{\partial s_{k}} f(s_{k}, \theta, \xi_{k+1}) \cdot D_{k}
    \tag{IPA}
\]
Likelihood-ratio gradient estimation on the other hand uses knowledge of the distribution of $\xi_{k}$ to form the gradient estimator. Suppose that $s_{k}$ is a Markov chain for which the transition kernel is parameterized by $\theta$, i.e., $s_{k+1} \sim p_{\theta}(\cdot |s_{k})$. %Let $p_{\theta}(s_{k},...,s_{0})=\prod_{j=1}^{k-1} p_{\theta}(s_{j+1}|s_{j})$. 
For a fixed $\theta_{0}$, let 
\[
\frac{\partial}{\partial \theta}\E_{\theta}[s_{k}] = \frac{\partial}{\partial \theta}\E_{\theta_{0}}[s_{k}L_{k}(\theta)] = \E_{\theta_{0}}\left[s_{k}\frac{\partial}{\partial \theta}L_{k}(\theta)\right]
~~\mbox{where}~~L_{k}(\theta) := \frac{\prod_{j=1}^{k-1} p_{\theta}(s_{j+1}|s_{j})}{\prod_{j=1}^{k-1} p_{\theta_0}(s_{j+1}|s_{j})}.
\]
This allows one to obtain the following gradient estimator:
\[
D_{k} = s_{k} \sum_{j=1}^{k-1} 
\frac{\frac{\partial}{\partial \theta}p_{\theta}(s_{j+1}|s_{j})}
{p_{\theta_{0}}(s_{j+1}|s_{j})}L_{k}(\theta), \mbox{ where } s_{j+1} \sim p_{\theta_{0}}(\cdot|s_{j}),\forall j \leq k.
\tag{LR}
\]

Despite their popularity, there are limitations to applying these methods to general multi-class queuing networks. While IPA has been proven efficient for simple queuing models, such as the $G/G/1$ queue through the Lindley recursion, it is well-known that unbiased IPA estimates cannot be obtained for general queuing networks~\cite{cao1987first,fu1994smoothed,fu2012conditional}. The implementation of LR gradient estimation hinges on precise knowledge of the system's Markovian transition kernel ~\cite{glynn1987likelilood}. This requires knowledge of the inter-arrival time and workload distributions, and even with this knowledge, it is non-trivial to specify the transition kernel of the queue lengths and residual event times in generic systems. Modifications to IPA \cite{fu1994smoothed, fu2012conditional} also require precise knowledge of event time distributions and often involve analyzing specific ordering of events which must be done on a case-by-case basis. As a result, none of these methods can reliably provide gradient estimation for complex queuing networks under general scheduling policies and with possibly unknown inter-arrival and service time distributions. Yet, the ability to handle such instances is important to solve large-scale problems arising in many applications.
% While these methodologies have shown to be successful in specific small-scale queueing networks, to the best of our knowledge, neither method can reliably provide gradient estimation for general queueing networks, i.e., under general routing policies with general and possibly unknown inter-arrival and service time distributions. Yet, the ability to handle such instances is important to solve large-scale problems arising in many applications.

Due to the challenges discussed above, existing reinforcement learning (RL) approaches for queueing network control mainly rely on model-free gradient estimators, utilizing either the $\reinforce$ estimator and/or $Q$-function estimation. As we will discuss shortly, these methods do not leverage the structural properties of queuing networks and may be highly sample-inefficient, e.g., requiring a prohibitively large sample for gradient estimation.

To address the challenges discussed above, we propose a novel gradient estimation framework that can handle general, large-scale multi-class queuing networks under any differentiable scheduling policy, requiring only samples of the event times rather than knowledge of their distributions. Most importantly, our approach streamlines the process of gradient estimation, leveraging auto-differentiation libraries such as PyTorch~\citep{PaszkeGrChChYaDeLiDeAnLe17} or Jax~\citep{jax2018github} to automatically compute gradients, rather than constructing these gradients in a bespoke manner for each network as is required for IPA or LR. As shown in Figure~\ref{fig:code}, computing a gradient in our framework requires only a few lines of code. To the best of our knowledge, this is the first scalable alternative to model-free methods for gradient estimation in queuing networks.

% At the same time, by leveraging the known dynamics of queuing networks, we can greatly improve sample efficiency over model-free policy gradient methods, foregoing the need for various hyperparameters and implementation details, such as return normalization, which can often be crucial for the performance and introduce complexity~\cite{engstrom2020implementation}. 

% Despite the discrete nature of the dynamics, this framework leverages knowledge of the queueing network dynamics and provides a simple way to compute derivatives directly through the dynamics, using auto-differentiation libraries such as . To the best of our knowledge, this is the first scalable alternative to $\reinforce$, and as we will demonstrate in the following sections, it can greatly improve upon $\reinforce$ in terms of sample efficiency.

\subsection{The standard approach: the $\reinforce$ estimator}
% \hntodo{Nit: let's use $\what{\cdot}$ instead of $\hat{\cdot}$ throughout the paper}

% \hntodo{Explicitly justify why you use REINFORCE as the main baseline. Use soft language like 
% this is a basic proof of concept, so as to not give the impression that we're claiming we're uniformly better than all RL approaches. We can do a bit of this in the abstract + intro.
% }

Considering the lack of differentiability in most reinforcement learning environments, the standard approach for gradient estimation developed in model-free RL is the score-function or $\reinforce$ estimator~\cite{williams1992simple,sutton1999policy}. This serves as the basis for modern policy gradient algorithms such as Trust-Region Policy Optimization (TRPO)~\cite{schulman2015trust} or Proximal Policy Optimization (PPO)~\cite{schulman2017proximal}. As a result, it offers a useful and popular baseline to compare our proposed method with.

 The core idea behind the $\reinforce$ estimator is to introduce a randomized policy $\pi_{\theta}$ and differentiate through the action probabilities induced by the policy. Under mild regularity conditions on $\pi_{\theta}$ and $c(x_k,u_k)$, the following expression holds for the policy gradient:
\[
\nabla J_{N}(\theta) = \mathbb{E}\left[\sum_{t=0}^{N-1} \left(\sum_{k=t}^{N-1}c(x_{k},u_{k})\tau^{*}_{k+1}  \right) \nabla_{\theta}\log \pi_{\theta}(u_{t}|x_{t})\right],
\]
%The term inside the expectation can be estimated pathwise, 
which leads to the following policy gradient estimator:
\begin{equation}
\label{eq:reinforce}
\widehat{\nabla}^{\mathsf{R}} J_{N}(\theta; \xi_{1:N}) = \sum_{t=0}^{N-1} \left(\sum_{k=t}^{N-1}c(x_{k},u_{k})\tau^{*}_{k+1}  \right) \nabla_{\theta}\log \pi_{\theta}(u_{t}|x_{t}).
\hspace{-3em}
\tag{$\reinforce$}
\end{equation}

While being unbiased, the $\mathsf{REINFORCE}$ estimator is known to have a very high variance~\cite{weaver2013optimal}. The variance arises from two sources. First, the cumulative cost $\sum_{k=t}^{N-1}c(x_{k},u_{k})\tau^{*}_{k+1}$ can be very noisy, as has been observed for queuing networks~\cite{dai2022queueing}. Second, 
%when the probabilities $\pi_{\theta}(u_{t}|x_{t})$'s are small, which naturally occurs 
as the policy converges to the optimal policy, the score function $\nabla_{\theta} \log \pi_{\theta}(u_{t}|x_{t})$ can grow large, magnifying the variance in the cost term.  Practical implementations involve many algorithmic add-ons to reduce variance, e.g., adding a `baseline' term~\cite{weaver2013optimal} which is usually (an estimate of) the value function $V_{\pi_{\theta}}(x_{k})$,
\begin{equation}
\label{eq:reinforce_baseline}
\widehat{\nabla}^{\mathsf{RB}} J_{N}(\theta; \xi_{1:N}) = \sum_{t=0}^{N-1} \left(\sum_{k=t}^{N-1}c(x_{k},u_{k})\tau^{*}_{k+1}  
 - V_{\pi_{\theta}}(x_{k})\right) \nabla_{\theta}\log \pi_{\theta}(u_{t}|x_{t}).
 \hspace{-1em}
 \tag{$\mathsf{BASELINE}$}
\end{equation}
These algorithmic add-ons have led to the increased complexity of existing policy gradient implementations~\cite{huang2022cleanrl} and the outsized importance of various hyperparameters~\cite{huang2022implementation}. It has even been observed that seemingly small implementation ``tricks" can have a large impact on performance, even more so than the choice of the algorithm itself~\cite{engstrom2020implementation}.

% To address this challenge, in most applications, the $\mathsf{REINFORCE}$ estimator is averaged across a large number of trajectories. Various variance reduction techniques have also been introduced. One of the common variance reduction strategies is introducing a baseline term, such as value functions estimated using the same sample path. In addition, modern implementations such as Proximal Policy Optimization (PPO)~\cite{schulman2017proximal} or Trust-Region Policy Optimization (TRPO)~\cite{schulman2015trust} put constraints on how far each iterate can move, in large part due to the unreliability of the gradients.\hntodo{This sentence will be too abstract for some readers. Need to specify what "iterates" mean etc. }
% These modifications introduce additional complexity into the gradient estimation process, with standard code implementations of PPO requiring at least 10,000 lines of code split across multiple files~\cite{huang2022cleanrl}. Moreover, the performance of standard policy gradient algorithms can be highly sensitive to  auxiliary implementation details. For example, the clipping ranges and normalization for the rewards and observations having an even larger impact on performance than the choice of RL algorithm itself~\cite{engstrom2020implementation}.  As a result of all these factors, model-free reinforcement learning is often sample-inefficient, computationally-intensive, and fragile~\cite{engstrom2020implementation, ilyas2018closer, huang2022cleanrl, huang2022implementation}.

\subsection{ Our approach: Differentiable Discrete-Event Simulation}

% \hntodo{I think this should be a different section}

%To mitigate these challenges and reliably optimize the control problem \eqref{eq:param-cost}, we develop an alternative framework for gradient estimation. 
% Unlike typical RL settings, where the dynamics of the environment are treated as a black box, queuing networks have structured and known dynamics as described in section \ref{section:model}.
% While the discrete-event representation of the queuing network is primarily used in order to allow for more general inter-arrival and service times.

We can view the state trajectory as a repeated composition of the transition function $s_{k+1} = f(s_{k},u_{k}, \xi_{k+1})$, which is affected by exogenous noise $\xi_{1:N}$, i.e., stochastic inter-arrival and service times. If the transition function were differentiable with respect to the actions $u_{k}$, then under any fixed trace $\xi_{1:N}$, one could compute a \emph{sample-path} derivative of the cost $J(\theta;\xi_{1:N})$ using auto-differentiation frameworks such as PyTorch~\cite{PaszkeGrChChYaDeLiDeAnLe17} or Jax~\cite{jax2018github}. Auto-differentiation software computes gradients efficiently using the chain rule. To illustrate, given a sample path of states, actions, and noise $(s_{k},u_{k},\xi_{k+1})_{k=0}^{N-1}$, we can calculate the gradient of $s_{3}$ with respect to $u_{1}$ via
\[
\frac{\partial s_{3}}{\partial u_{1}} = \frac{\partial s_{3}}{\partial s_{2}} \frac{\partial s_{2}}{\partial u_{1}} =
\frac{\partial f(s_{2},u_{2}, \xi_{3})}{\partial s_{2}} \frac{\partial f(s_{1},u_{1}, \xi_{2})}{\partial u_{1}}.
\]
This computation is streamlined through a technique known as backpropagation, or reverse-mode auto-differentiation. The algorithm involves two steps. The first step, known as the \emph{forward pass}, evaluates the main function or performance metric (in the example, $s_{i}$'s) and records the partial derivatives of all intermediate states relative to their inputs (e.g. $\partial s_{2} / \partial u_{1}$). 
This step constructs a computational graph, which outlines the dependencies among variables. The second step is a \emph{backward pass}, which traverses the computational graph in reverse. It sequentially multiplies and accumulates partial derivatives using the chain rule, propagating these derivatives backward through the graph until the gradient concerning the initial input (in this example, $u_{1}$) is calculated. Due to this design, gradients of functions involving nested compositions can be computed in a time that is linear in the number of compositions. By systematically applying the chain rule in reverse, auto-differentiation avoids the redundancy and computational overhead typically associated with numeric differentiation methods.

However, as mentioned before, the dynamics do not have a meaningful derivative due to the non-differentiability of actions and the $\mathsf{argmin}$ operation which selects the next event based on the minimum residual event time. Yet if we can utilize suitably differentiable surrogates, it would be possible to compute meaningful approximate sample-path derivatives using auto-differentiation.
% Yet, under the discrete-event dynamical system representation, the non-differentiability is isolated in a single operation: the $\mathsf{argmin}$ which selects the next event based off the minimum residual event time.
% \begin{align}
%     e_{k+1} & =\mathsf{argmin}\cup_{i=1}^{n}\{\tau_{k}^{A_{i}},\tau_{k}^{S_{i}}/u_{i}^{\top}\mu_{i}\}\tag{Event Selection}
% \end{align}

\subsubsection{Capacity sharing relaxation} 
% \hntodo{You may get a lot of questions on this relaxation. Discuss in detail whether it's restricting or not restricting. Is it only for learning a policy? Do you present all results in this new formulation so you've effectively changed the name of the game? Does a naive rounding affect performance?}

First, we address the non-differentiability of the action space. Recall that $u_{k} \in \{0,1\}^{m \times n}$ are scheduling decisions, which assign jobs to servers. Since $u_{k}$ lies in a discrete space, a small change in the policy parameters can produce a jump in the actions. To alleviate this, we consider the transportation polytope as a continuous relaxation of the original action space~\eqref{eq:action_space}:
% \begin{equation}
% \label{eq:bar_action_space}
% u_{k} \in \bar{\mathcal{U}} := \left\{u \in [0,1]^{m\times n}: \sum_{i=1}^{m}u_{ij} = 1, \sum_{j=1}^{n}u_{ij} = 1, u \leq M
% \right\}
% \end{equation}
\begin{equation}
\label{eq:bar_action_space}
\overline{\mathcal{U}}:=\left\{ u\in [0,1]^{m\times n}:u\ones_{n} = \ones_{m}, \ones_{m}^{\top} u=\ones_{n},u\leq M\right\}.
\end{equation}
The set of extreme points of $\overline{\mathcal{U}}$ coincide with the original, integral action space $\mathcal{U}$. %However, it is unclear how to interpret 
For a fractional action $u_{k}\in \overline{\mathcal{U}}$, %in the dynamics of the queuing network. We propose a relaxation motivated from fluid models of queuing networks \cite{chen1994hierarchical}: fractional routing decisions represent 
we can interpret it as servers splitting their capacity among multiple job classes motivated by the fluid approximation of queues \cite{chen1994hierarchical}. As a relaxation, it allows servers to serve multiple jobs simultaneously. The effective service rate for each job class is equal to the fraction of the capacity allocated to the job class multiplied by the corresponding service rate. 
%The less capacity allocated to a job, the longer it will take to complete.

As a result, instead of considering stochastic policies over discrete actions, we approach this problem as a \emph{continuous} control problem and consider
\emph{deterministic} policies over continuous actions, i.e., the fractional scheduling decisions. Under this relaxation, the processing times are differentiable in the (fractional) scheduling decision. Finally, it is worth mentioning that we only use this relaxation when training policies. For policy evaluation, we enforce that actions are integral scheduling decisions in $\mathcal{U}$. To do so, we treat the fractional action as a probability distribution and use it to sample a discrete action. 

% \hntodo{Say this earlier, at the beginning of this subsection. Assume that people will view this framework with skepticism, and actively mitigate it by having these sentences up front and center.}
%To reiterate,
\begin{definition}
\label{def:capacity-sharing}
Under the {\bf capacity sharing relaxation}, the service rate for queue $j$ under the routing decision $u\in\overline{\mathcal{U}}$ is $\mu_{j}^\top u_{j} \equiv \sum\nolimits_{i=1}^{m} \mu_{ij}u_{ij}$.
Thus, given workload $w_{j}$, the processing time of the job will be
\begin{equation}
\label{eq:capacity-sharing}
\tau^{S}_{j} =\frac{w_{j}}{\sum_{i=1}^{m}u_{ij}\mu_{ij}} = \frac{w_{j}}{\servicemap_{j,j}}.
\end{equation}
\end{definition}
Note that this is identical to the original definition of the processing times in~\eqref{eqn:processing_time}. The only difference is that we now allow fractional routing actions, under which a server can serve multiple jobs at the same time.

For a concrete example, consider a single server $i$ compatible with two job classes 1 and 2 with service rates $\mu_{i1} = 9$ and $\mu_{i2} = 15$ respectively. Suppose it splits its capacity between job classes 1 and 2 according to $u_{i1} = 1/3$ and $u_{i2} = 2/3$. Then for residual workloads $w_1$ and $w_2$, the corresponding processing times are $\tau^{S}_{1}=w_1/3$ and $\tau^{S}_{2}=w_2/10$. If $u_{i1}=0$ and $u_{i2}=1$ instead, then the corresponding processing times are $\tau^{S}_{1}=w_1/0 = \infty$ and $\tau^{S}_{2}=w_2/15$. %Note that when $u_{k}$ is integral, departure times under the capacity sharing relaxation~\eqref{eq:capacity-sharing} perfectly coincide with departure times under the original, discrete action space.

% Above all, unlike the standard model-free RL approach, which considers stochastic policies $\pi_{\theta}$ that draw actions from the discrete combinatorial action space $\mathcal{U}$, 

% $\pi_{\theta}\in \bar{U}$ that produce fractional routing decisions that determine the effective service rates in a continuous way. 

%{\color{blue} Add an example here.}

\subsubsection{Differentiable event selection}
% \hntodo{Nits: jacobian $\Rightarrow$ Jacobian, iid $\Rightarrow$ i.i.d.}

To determine the next event type, the $\mathsf{argmin}$ operation selects the next event based on the minimum residual event time. This operation does not give a meaningful gradient.

\paragraph{Pitfalls of `naive' smoothing}
In order to compute gradients of the sample path, we need to smooth the $\mathsf{argmin}$ operation. There are multiple ways to do this. A naive approach is to directly replace $\mathsf{argmin}$ with a differentiable surrogate. %Instead of returning a one-hot vector $\{0, 1\}^{2n}$ corresponding to the index of the minimum residual event time, the 
One such popular surrogate is $\softmin$. With some inverse temperature $\beta > 0$, $\softmin_\beta$ applied to the vector of residual event times $\tau \in \R_{+}^{2n}$ returns a vector in $\R_{+}^{2n}$, 
which we use to replace the event selection operation $\event_{k+1}$: 
\begin{equation}
\label{eq:direct_smoothing}
\tilde{\event}_{k+1} = \mathsf{softmin}_{\beta}(\tau_{k})
~~~\mbox{where}~~~\mathsf{softmin_{\beta}(\tau)}_{j}=e^{-\beta \tau_{j}} / \sum_{l=1}^{2n} e^{-\beta \tau_{l}}.
\hspace{-1em}
\tag{Direct Smoothing}
\end{equation}
As $\beta \to \infty$, $\mathsf{softmin}_{\beta}$ converges to $\mathsf{argmin}$. Thus, one may expect that for large $\beta$, $\mathsf{softmin}_{\beta}$ would give a reliable differentiable surrogate. 

\begin{figure}[t]
\centering
    \includegraphics[height = 2.2in]{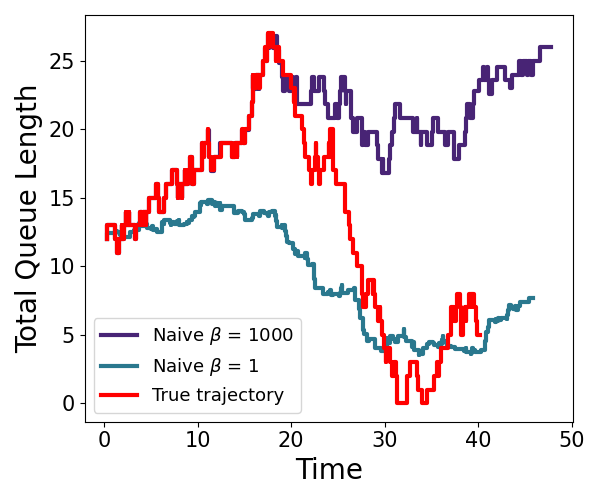}
    % \hspace{0.1em}
    % \includegraphics[height = 1.6in]{plot/gradient/direct_smoothing_vectors.png}
    \hspace{0.1em}
    \includegraphics[height = 2.2in]{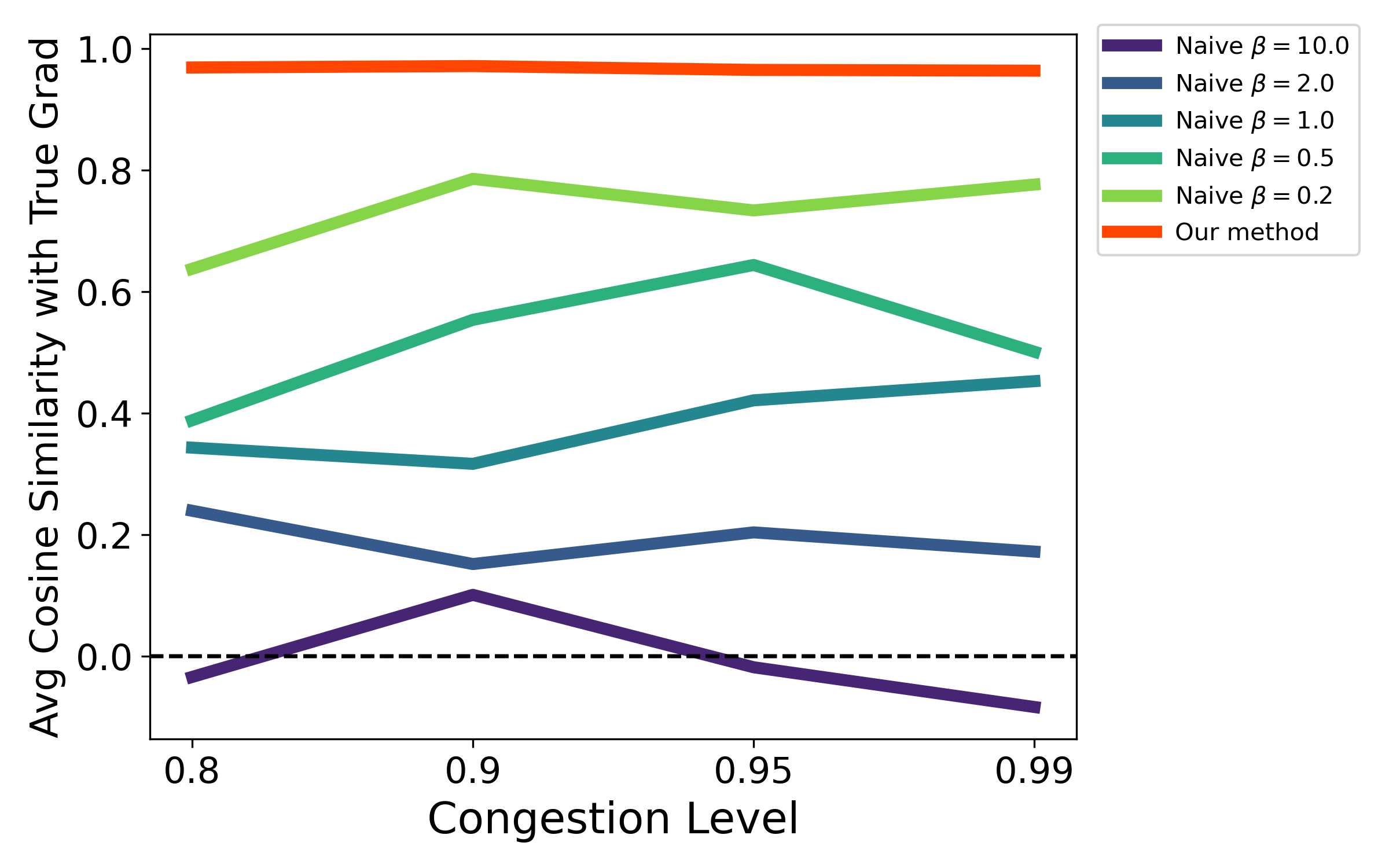}
    
    \caption{Failure modes of `naive' smoothing. (Left) Comparison of sample paths under the original dynamics and direct smoothing with $\beta = 1$ and $\beta = 1000$. Criss-cross network under a randomized backpressure policy with identical event times in each path, for $N = 200$ steps. 
    Even under high inverse temperature $\beta = 1000$, the trajectory veers off from the original trajectory after only a hundred steps. 
    % (Middle) Samples of policy gradient estimators for the randomized backpressure policy in the criss-cross network estimated along a horizon of $N = 1000$. The true gradient (in black) is estimated by averaging the $\mathsf{REINFORCE}$ estimator over $10^{7}$ trajectories. 
    % Direct smoothing for $\beta = 1$ has low variance but is biased. Direct smoothing with $\beta = 1000$ has a prohibitively high variance. Our proposed method greatly reduces bias while maintaining a low variance.
    (Right) Comparison of average cosine similarity (higher is better) of gradient estimators using direct smoothing with $\beta  \in \{0.2, 0.5, 1, 2, 10\}$ for the criss-cross network under a randomized MaxWeight policy for $N = 1000$ steps. 
    The gradient estimators either suffer from high bias or high variance and are unable to achieve a high cosine similarity with the true gradient.
    }
    \label{fig:naive_smoothing}
\end{figure}

However, queuing networks involve a unique challenge for this approach: one typically considers very \emph{long} trajectories when evaluating performance in queuing networks, as one is often interested in long-run average or steady-state behavior. Thus, even if one sets $\beta$ to be very large to closely approximate $\argmin$, the smoothing nonetheless results in `unphysical', real-valued queue lengths instead of integral ones, and small discrepancies can accumulate over these long horizons and lead to entirely different sample paths. 
This can be observed concretely in the left panel of Figure~\ref{fig:naive_smoothing}, which displays the sample paths of the total queueing length processes for a criss-cross queueing network (in Example~\ref{example:criss-cross}) under the original dynamic and under direct smoothing, using the same inter-arrival and service times.  We observe that when setting the inverse temperature $\beta = 1$, the sample path under direct smoothing is completely different from the original one, even though all of the stochastic inputs are the same. 
%The sample path is much smoother. 
Even when setting a very high inverse temperature, i.e., $\beta = 1000$, for which $\mathsf{softmin}_{\beta}$ is almost identical to $\mathsf{argmin}$, the trajectory veers off after only a hundred steps. 

This can greatly affect the quality of the gradient estimation. 
%The middle panel of Figure~\ref{fig:naive_smoothing} displays samples of the smoothed gradient estimator for a policy gradient task. Direct smoothing under $\beta = 1$ leads to a low-variance but biased estimator. Yet, if we set $\beta = 10$, the variance can be large and the gradient estimator does not reliably point in any direction. This points to a trade-off that prevents reliable gradient estimation: low inverse temperatures bias the gradient estimates whereas high inverse temperatures lead to numerical instabilities. Even if this hyper-parameter is carefully tuned, the estimators either have too high a bias or variance rather than reaching a meaningful tradeoff. For example,
We observe in the right panel of Figure~\ref{fig:naive_smoothing} that across a range of inverse temperatures, the average cosine similarity between the surrogate gradient and the true gradient (defined in \eqref{eqn:similarity}) are all somewhat low.
%the estimators either have too high a bias or variance rather than reaching a meaningful tradeoff. 
In the same plot, we also show the average cosine similarity between our proposed gradient estimator, which we will discuss shortly, and the true gradient. Our proposed approach substantially improves the gradient estimation accuracy, i.e., the average cosine similarity is close to $1$, and as we will show later, it does so across a wide range of inverse temperatures. %Overall this illustrates the unique challenges that emerge for constructing differentiable simulators of discrete dynamical environments. 
% In the machine learning literature, smoothing techniques for gradient estimation have typically been applied to tasks such as density estimation or structured prediction~\cite{maddison2016concrete,jang2017categorical}, which are one-period problems that do not involve repeated smoothing across a trajectory. 
% Recently, \cite{petersen2021learning} develop differentiable relaxations for algorithms (such as sorting or shortest path) by applying differentiable relaxations for discrete operations. 
% Most relevant for our work, \cite{andelfinger2021differentiable, andelfinger2023towards} propose differentiable agent-based simulators based on differentiable relaxations, but find that while they helped achieve strong performance on optimization tasks, these relaxations also created unpredictable discrepancies with the original dynamics. Reliable simulation involved carefully balancing the trade-off in the level of smoothing in order to balance fidelity to the original environment with reliability of the gradient estimates. These considerations are especially crucial for queuing networks, as reliably estimating steady-state quantities requires rolling out the dynamics for horizons ranging from $N = 10,000$ to even $N= 100,000$ steps.

\paragraph{Our approach: `straight-through' estimation}

The failure of the direct smoothing approach highlights the importance of preserving the original dynamics, as errors can quickly build up even if the differentiable surrogate is only slightly off. We propose a simple but crucial adjustment to the direct smoothing approach, which leads to huge improvements in the quality of gradient estimation. 
% \hntodo{Notice how you don't mention how you pick $\beta$ until the very end of this section. Again, assume by default that people will be skeptical of your highly novel and innovative empirical framework. Actively take measures to mitigate concerns. I would put the paragraph on insensitive against $\beta$ and the corresponding empirical plots front and center, right after you introduce "our method"}

Instead of replacing the $\mathsf{argmin}$ operation with $\mathsf{softmin}_{\beta}$ when generating the sample path, we preserve the original dynamics as is, and only replace the Jacobian of $\mathsf{argmin}$ with the Jacobian of $\softmin_\beta$ when we query gradients. In short, we introduce a gradient operator $\widehat{\nabla}$ such that
\begin{equation}
\event_{k+1}=\mathsf{argmin}(\tau_{k}), \quad \quad
\widehat{\nabla} \event_{k+1}= \nabla \softmin_{\beta}(\tau_{k}). 
\end{equation}
where $\nabla$ is respect to the input $\tau$.
%while $\widehat{\nabla}$ is equivalent to standard gradient operator for all other differentiable functions.
% Plot of maybe MM1 (or something more complicated). Our method gets the right gradient, but direct smoothing gets something crazy.
%  Our approach is to preserve the $\mathsf{argmin}(\tau)$ in the forward pass of dynamics, when we roll out the dynamics on a particular sample path, but replace with the $\mathsf{softmin_{\beta}(\tau)}=e^{\beta \tau_{j}} / \sum_{j=1}^{2n} e^{\beta \tau_{j}}$ operation when computing
% To overcome this issue, we apply a differentiable surrogate in a careful manner. We maintain the $\mathsf{argmin}$ operation when generating the sample path to preserve the original dynamics exactly as is. It is only when we query gradients, that we then replace the jacobian of $\mathsf{argmin}$ with the jacobian of the $\softmin$ operation, defined as
% $\mathsf{softmin_{\beta}(\tau)}=e^{-\beta \tau_{j}} / \sum_{i=1}^{2n} e^{-\beta \tau_{i}}$, so that:
% \begin{equation}
% \event_{k+1}=\argmin(\nu(\tau_{k}, u_{k})), \quad \quad
% \nabla \event_{k+1}= \nabla \softmin_{\beta}(\nu(\tau_{k}, u_{k})) 
% \end{equation}
This is known as the `straight-through' trick in the machine learning literature and is a standard approach for computing approximate gradients in discrete environments~\cite{bengio2013estimating, van2017neural}. To the best of our knowledge, this is the first application of this gradient estimation strategy for discrete-event dynamical systems. Using this strategy, we can use the chain rule to compute gradients of performance metrics that depend on the event selection. 
% We denote $\widehat{\nabla}$ to be the corresponding gradient operator. 
Consider any differentiable function $g$ of $e_{k+1}$, 
%(e.g. a holding cost for the queue lengths which is updated by the event selection):
\[
\widehat{\nabla} g(\event_{k+1}) = \frac{\partial g(\event_{k+1})}{\partial \event_{k+1}} \widehat{\nabla} \event_{k+1} \nabla \tau_{k}= \frac{\partial g(\event_{k+1})}{\partial \event_{k+1}} \nabla \softmin_{\beta}(\tau_{k}) \nabla \tau_{k}
\]
In contrast, direct smoothing involves the derivative $\partial g(\tilde{e}_{k+1})/\partial \tilde{e}_{k+1}$ where $\tilde{e}_{k+1} = \softmin_{\beta}(\tau_{k})$.  Evaluating the gradient of $g$ at $g(\tilde{e}_{k+1})$ is a cause of additional bias.
% \hntodo{I also often see that because OR/MS does not have a well-developed empirical rigor standard (e.g., in contrast to top-tier LLM research these days), there is a default belief that anything empirical must be viewed with skepticism and that most researchers are not thorough (in some cases that they are sometimes even dishonest). We'd do well to explicitly highlight many times throughout the paper (starting in the introduction and abstract) how systematic and comprehensive your evaluations are. Think of concrete numbers and figures you can say, e.g., number of environments, number of policy params etc etc.}

%\subsubsection{Come up with a title for this}
With these relaxations, the transition function of the system is differentiable. We can now compute a gradient of the sample path cost $J_{N}(\theta; \xi_{1:N})$, using the chain rule on the transition functions. Given a sample path of states $s_{k} = (x_{k}, \aux_{k})$, actions $u_{k} = \pi_{\theta}(x_{k})$, and $\xi_{k} = (T_{k}, W_{k})$, the pathwise gradient of the sample path cost is $J_{N}(\theta; \xi_{1:N})$ with respect to an action $u_{k}$ is,
\[
\widehat{\nabla}_{u_{k}} J_{N}(\theta; \xi_{1:N})
= \underbrace{\nabla_{u_{k}}c(x_{k},u_{k})}_{\text{current cost}} + \sum_{t=k+1}^{N} 
\left[  \underbrace{ \nabla_{x}c(x_{t}, u_t) + \nabla_{u} c(x_{t}, u_t)\nabla_{x_{t}}\pi_{\theta}(x_{t})}_{\text{future costs}} \right]
\nabla_{u_{k}}x_{t}
\]
The gradient consists of the sensitivity of the current cost with respect to the action as well the sensitivity of future costs via the current action's impact on future states. The policy gradient with respect to $\theta$ can then be computed as
\begin{equation}
\label{eq:pathwise-pg}
\widehat{\nabla}_{\theta} J_{N}(\theta; \xi_{1:N}) = \sum_{k=1}^{N} \left. \widehat{\nabla}_{u_{k}} J_{N}(\theta; \xi_{1:N}) \right|_{u_{k} = \pi_{\theta}(x_{k})} \nabla_{\theta} \pi_{\theta}(x_{k}).
\hspace{-3em}
\tag{$\pathwise$}
\end{equation}
As a result of the straight-through trick, we do not alter the event selection operation $\event_{k}$'s and thus the state trajectory $\{x_{k} \}_{k=1}^{N}$ is unchanged.

% \hntodo{We should highlight this point way earlier, starting in Section 1, but also throughout Sections 2-3. We need to accentuate the computational nature of our regimen. Until very recently, taking these gradients required hand-coding them in, but now we have extremely effective ways to do auto-differentiation. This is precisely why the reader had to go through the cumbersome notation in the entirety of Section 2. Once you have the entire problem in matrix-vector notation, we basically have a "pseudo-code" for our method. This point should come out in Sections 1 and 2.} 

We refer to the gradient estimator \eqref{eq:pathwise-pg} as the $\mathsf{PATHWISE}$ {\bf policy gradient estimator}. Although this formula involves iterated products of several gradient expressions, these can be computed efficiently through reverse-mode auto-differentiation using libraries such as PyTorch~\cite{PaszkeGrChChYaDeLiDeAnLe17} or Jax~\cite{jax2018github} with $O(N)$ time complexity in the time horizon $N$. This time complexity is of the same order as the forward pass, i.e., generating the sample path itself, and is equivalent to the time complexity of $\mathsf{REINFORCE}$. The policy gradient algorithm with $\mathsf{PATHWISE}$ gradient is summarized in Algorithm \ref{alg:pathwise-pg}. In Section~\ref{section:gradient_eval}, we perform a careful empirical comparison of $\pathwise$ and $\reinforce$, and find that $\pathwise$ can lead to orders of magnitude improvements in sample efficiency. 
% Although the straight-through method preserves the sample path on which we evaluate the gradient, it is still unclear whether this will result in an effective gradient estimator, since we are still using a surrogate gradient.
%inputting a surrogate gradient in-place of the uninformative gradient of the $\argmin$ at every step. 
% Later in section~\ref{section:gradient_eval}, we empirically investigate the statistical properties of the gradients, namely whether
% \[
% \E[\widehat{\nabla}_{\theta} J_{N}(\theta; \xi_{1:N})]
% \approx \nabla_{\theta} J_{N}(\theta) \quad \text{and} \quad 
% \var(\widehat{\nabla}_{\theta} J_{N}(\theta; \xi_{1:N})) < \var(\widehat{\nabla}^{\mathsf{R}}_{\theta} J_{N}(\theta; \xi_{1:N}))
% \]
% \hntodo{I don't think this is the right algo box to state. I recommend coming up with a visual scheme of iteratively applying $f$'s and illustrating the straight-through in the same figure. }

\begin{algorithm}[t]
  \caption{\label{alg:pathwise-pg} $\pathwise$ Policy Gradient ($\mathsf{PathPG}$) }
  \begin{algorithmic}[1]
    \State \textsc{Input: Policy $\pi_{\theta}$, Number of Iterates $T$, horizon $N$, trace $\xi_{1:N}$, step-size $\alpha>0$}.
    \For{each $t \in 1, \ldots, T$}
    \State Compute $J_{N}(\theta;\xi_{1:N}) = \sum_{k=1}^{N} c(x_{k}, \pi_{\theta}(x_{k})) \tau_{k+1}^{*}$ from trace $\xi_{1:N}$.
    \State Compute gradient $\widehat{\nabla}_{\theta} J_{N}(\theta; \xi_{1:N})$ via~\eqref{eq:pathwise-pg}
    \State Update policy parameters,
    \[
    \theta_{t+1} \leftarrow \theta_{t} - \alpha \widehat{\nabla}_{\theta} J_{N}(\theta; \xi_{1:N})
    \]
    \EndFor 
    \State \Return $\pi_{\theta_{T}}$
  \end{algorithmic}
\end{algorithm}

% \hntodo{Again, from the skepticism view, I would err on the side of stating this early on. I also found this paragraph quite confusing because it is unclear what the difference between evaluating a policy vs. model is.}
It is important to re-emphasize that when computing the gradient, we evaluate the policy differently than we would for $\reinforce$. Instead of drawing a random discrete action $u_{k} \sim \pi_{\theta}(x)$, we use the probabilities output by the policy directly as a fractional routing matrix in $\overline{\mathcal{U}}$,
\begin{align*}
\reinforce:& \quad u_{k} \sim \pi_{\theta}(x_{k}),\qquad u_{k} \in \mathcal{U} \\
\pathwise:& \quad u_{k} = \pi_{\theta}(x_{k}),\qquad u_{k} \in \overline{\mathcal{U}}.
\end{align*}
However, this is only for gradient computation. When we evaluate the policy, we draw $u_{k} \sim \pi_{\theta}(x_{k})$.

\begin{figure}[t]
    \includegraphics[height = 2.3in]{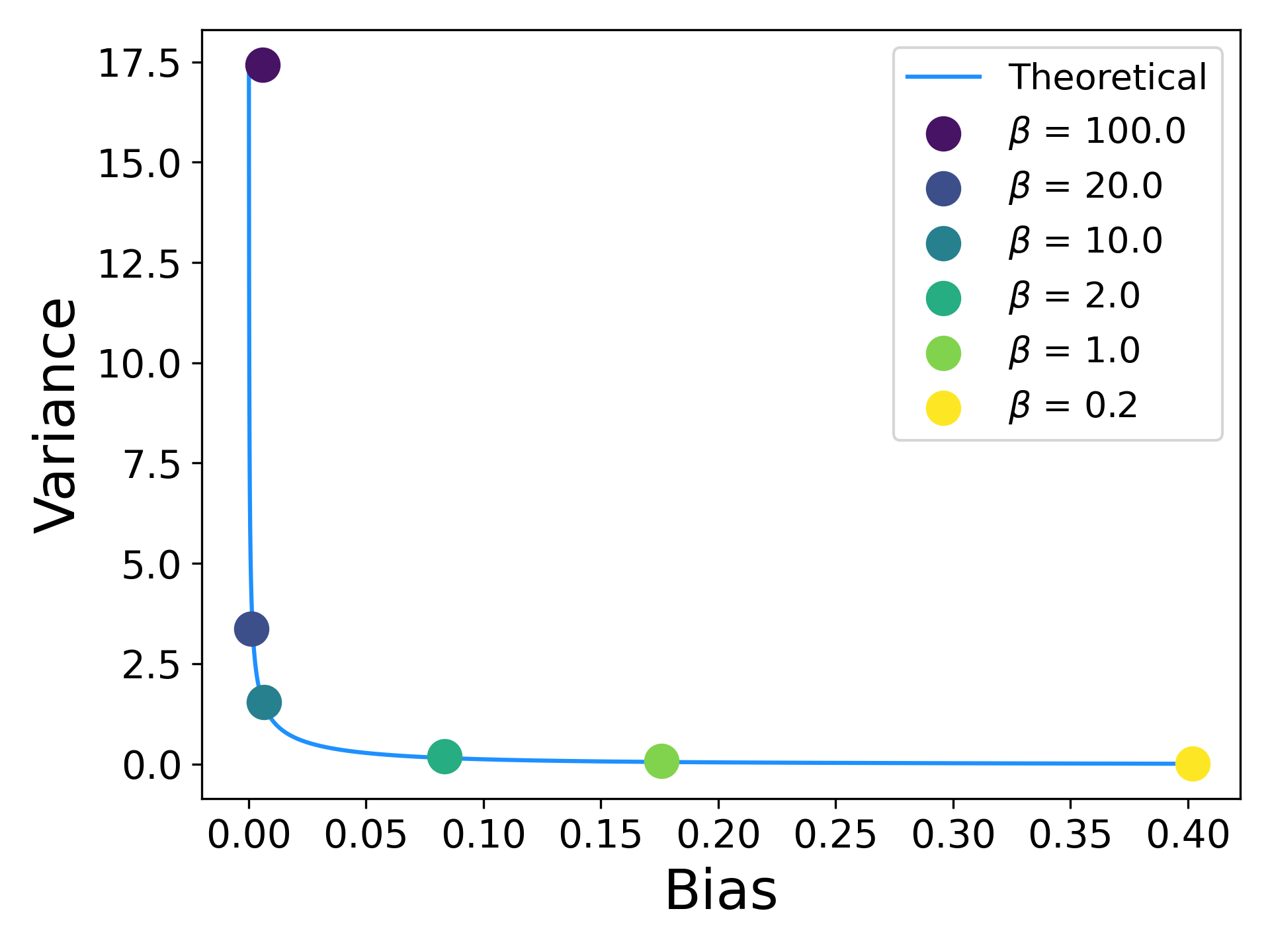}
    \includegraphics[height = 2.3in]{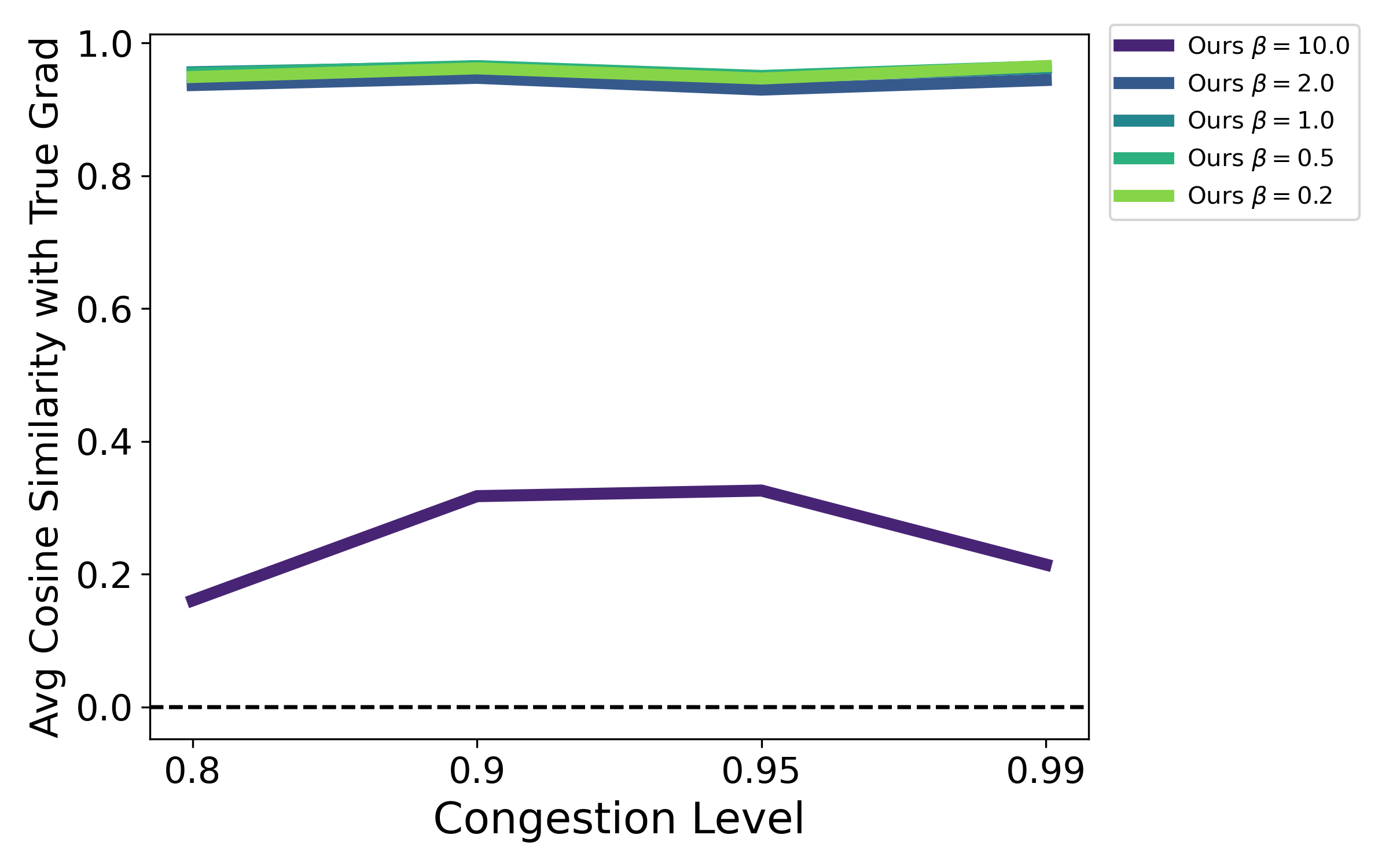}
    
    \caption{Bias-variance trade-off for inverse temperature $\beta$. (Left) Bias and variance of pathwise gradient estimator across a range of inverse temperatures $\beta$ for the one-step change $\widehat{\nabla}_{\mu}(x_{k+1} - x_{k})$ compared to true gradient $\nabla_{\mu}\E[x_{k+1} - x_{k}] = \nabla_{\mu}\frac{\lambda - \mu}{\lambda + \mu}$. Blue line specifies the theoretical values for the bias and variance. (Right) Average cosine similarity (see~\eqref{eqn:similarity} for the definition) of $\pathwise$ policy gradients with the true gradient for a randomized max-weight policy (see~\eqref{eqn:soft_policies}) in a 6-class reentrant network (see Figure~\ref{fig:networks}) across a range of inverse temperatures computed from a single trajectory $B = 1$ for $N = 1000$ steps. True gradient is computed by averaging $\reinforce$ over $10^{6}$ trajectories. $\pathwise$ gradient has almost a perfect cosine similarity $\approx 1$ across a wide range of inverse temperatures.}
    \label{fig:bias-variance} 
\end{figure}

\paragraph{Bias-variance trade-off for inverse temperature: One-step analysis}
The inverse temperature $\beta$ is a key hyperparameter that determines the fidelity of the $\softmin_{\beta}$ approximation to $\argmin$. The choice of $\beta$ poses a bias-variance trade-off, with a higher $\beta$ leading to a smaller bias but a higher variance and a smaller $\beta$ incurring a higher bias but a lower variance. 

In general, it is difficult to assess the bias since we  often do not know the true gradient. However, for some simple examples, we can evaluate the true gradient explicitly. We next analyze the gradient of the one-step transition of the $M/M/1$ queue with respect to the service rate $\mu$,
\[
\nabla_{\mu}\E[x_{k+1} - x_{k}]=  \nabla_{\mu} \E[D e_{k+1}] =\nabla_{\mu}\frac{\lambda - \mu}{\lambda + \mu} = -2\frac{\lambda}{(\lambda + \mu)^{2}}.
\]
This permits an exact calculation of the mean and variance of our proposed pathwise gradient estimator. Although we can derive analytical expressions for these quantities, we present the leading order asymptotics as $\beta \to \infty$ for conciseness of presentation. While it is straightforward to see that almost surely $\softmin_{\beta}(\tau)\to \argmin(\tau)$ as $\beta \to \infty$, it is much less clear whether the gradient converges, i.e., whether $\E[\nabla \softmin_{\beta}(\tau)] \to \nabla \E[\argmin(\tau)]$. Since $\argmin$ has a gradient of zero almost everywhere, the expectation and gradient operators cannot be interchanged. 
%Using a pointwise convergence argument would lead to an incorrect gradient of zero. 
Instead, we analyze the expectations directly using properties of the exponential distribution.
\begin{theorem}
\label{thm:mm1-smoothing}
Let $\widehat{\nabla}_\mu (x_{k+1} - x_{k}) =  \widehat{\nabla}_\mu De_{k+1}$ denote the $\mathsf{PATHWISE}$ gradient estimator of the one-step transition of the $M/M/1$ queue with respect to $\mu$.  For $x_k\geq 1$, as $\beta \to \infty$,
\begin{align*}
\E[\widehat{\nabla}_\mu De_{k+1}]
-\nabla_\mu \E[ De_{k+1}]
&= \beta^{-2} \cdot \frac{\pi^{2}\lambda (\mu^{2} - \lambda^{2} + 2\mu \lambda)}{6(\lambda + \mu)^{2}}
+ o \left( \beta^{-2} \right), \\
\var(\widehat{\nabla}_\mu De_{k+1})
&= \beta \cdot \frac{4\lambda}{\mu (\lambda + \mu)^{2}} + o(\beta).
\end{align*}
\end{theorem}
See section~\ref{section:mm1-smoothing-proof} for the proof.
As $\beta \to \infty$, the bias is $O(1/\beta^{2})$ while the variance is $O(\beta)$. This means that one can significantly reduce the bias with only a moderate size of $\beta$. The left panel of Figure~\ref{fig:bias-variance} shows the bias-variance trade-off of $\widehat{\nabla}_\mu (x_{k+1} - x_{k})$ for various inverse temperatures $\beta$. The blue line is based on the analytical expression for the bias-variance trade-off curve. We observe that with the inverse temperature $\beta \in [0.5, 2]$, both the bias and the variance are reasonably small.
%the bias can be made to be small with only a mild increase in variance

%The favorable bias-variance trade-off translates into sample efficiency of the gradient estimator, as can be seen in Corollary~\ref{cor:one-step-mse}.
\begin{corollary}
\label{cor:one-step-mse}
Suppose we compute the sample average of $B$ iid samples of $\widehat{\nabla}_{\mu} De_{k+1}$, which are denoted as $\widehat{\nabla}_{\mu} De_{k+1, i}$, $i=1,\dots, K$. In particular, the estimator takes the form $\frac{1}{B}\sum_{i=1}^{B} \widehat{\nabla}_{\mu} De_{k+1,i}$. The choice of $\beta$ that minimizes the mean-squared error (MSE) of the estimator 
%$\text{MSE}(\beta) := \text{Bias}(\beta)^{2} + \text{Variance}(\beta)/B$ 
is $\beta^{*} = O(B^{1/5})$ and $\text{MSE}(\beta^*)=O(B^{-4/5})$.
\end{corollary}

The $\pathwise$ estimator provides a more statistically efficient trade-off than other alternatives. As an example, a standard gradient estimator is the finite-difference estimator in which one evaluates the one-step transition at $\mu - h$ and $\mu + h$ for some small $h \in (0,\infty)$, and the estimator is constructed as
\[
\frac{1}{B}\sum_{i=1}^{B} \frac{De_{k+1,i}(\mu + h) - De_{k+1,i}(\mu - h)}{2h},
\]
where $De_{k+1,i}(\mu + h)$'s are iid samples of $De_{k+1}(\mu + h)$.
If we set $h = 1/\beta$, it is well-known that the bias scales as $O(1/\beta^{2})$ while the variance scales as $O(\beta^{2})$. The choice of $\beta$ that minimizes the MSE is $\beta^{*} = O(B^{1/6})$ and $\text{MSE}(\beta^*)=O(B^{-1/3})$.

% \hntodo{State before theory, say many times in the introduction and subsequent sections that our approach is practically insensitive to smoothing parameter so long as you use straight-through. Reiterate and forward reference to empirical validations to come.}
While this analysis is restricted to the one-step transition of the $M/M/1$ queue, these insights hold for more general systems and control problems. The right panel of Figure~\ref{fig:bias-variance} displays the average cosine similarity (defined in ~\eqref{eqn:similarity}) between the $\pathwise$ gradient estimator and the true gradient for a policy gradient task in a 6-class reentrant network across different congestion levels and for different inverse temperatures. We observe that for a wide range of inverse temperatures, $\beta \in \{0.2, 0.5, 1, 2\}$, the estimator has near-perfect similarity with the true gradient, while a very large inverse temperature suffers due to high variance. This indicates that while there is a bias-variance trade-off, the performance of the $\pathwise$ gradient estimator is not sensitive to the choice of the inverse temperature within a reasonable range. In our numerical experiments, we %fix $\beta = 10$ or $\beta = 1$ and 
find that one can get good performance using the same inverse temperature across different settings without the need to tune it for each setting. 
%{\color{blue} We also conduct a case study comparing the sample efficiency of $\reinforce$ and $\pathwise$ for the $M/M/1$ queue in section \ref{section:case_study}.}

% \hntodo{From the viewpoint of mitigating skepticism, this figure is dangerously misleading. We've been differentiating between Direct vs ours so far, yet it appears that we now use "Direct" to mean ours. We should always denote "Ours $\beta$" vs. "Direct $\beta$" to differentiate between straight-through, even in single step problems where the distinction doesn't matter. It is easy for casual readers to miss these things.

% Moreover, it is genuinely difficult for people to grasp that $\beta = 10$ is already a pretty extreme value, and that $\beta = 100$ is an incredibly large number. Casual readers (unfortunately sometimes even referees) will think that if you plot 6 lines without much explicit context and see half of them suck, then the method is sensitive and problematic.}
% One can contrast this with the standard discrete-time MDP representation, in which the transition function is determined by a draw from a discrete probability distribution: $s_{t+1} \sim P(\cdot | s_{t},u_{t})$.

\section{Empirical Evaluation of the $\pathwise$ Gradients}
\label{section:gradient_eval}

In the previous section, we introduced the $\pathwise$ gradient estimator for computing gradients of queuing performance metrics with respect to routing actions or routing policy parameters. In this section, we study the statistical properties of these gradient estimators and their efficacy in downstream policy optimization tasks. We use $\reinforce$ as the baseline gradient estimator. First, in section~\ref{sec:gradient_efficiency} we empirically study the estimation quality across a range of queuing networks, traffic intensities, and policies. After that, in section~\ref{section:learn_cmu}, we investigate their performance in a scheduling task: learning the $c\mu$ rule in a multi-class queuing network. Finally, we demonstrate the applicability of our framework beyond scheduling: we investigate the performance of the $\pathwise$ gradient estimator for admission control tasks in section~\ref{section:admission}. 

% \hntodo{Hyperref to subsection names, and demarcate between this and next section clearly.}

\subsection{Gradient Estimation Efficiency}
\label{sec:gradient_efficiency}
In general, it is challenging to theoretically compare the statistical properties of different gradient estimators, and very few results exist for systems beyond the $M/M/1$ queue (see section \ref{section:case_study} for a theoretical comparison between $\reinforce$ and $\pathwise$ for the $M/M/1$ queue). For this reason, we focus on numerical experiments across a range of environments and queuing policies typically considered in the queuing literature. Specifically, we will be comparing the statistical properties of 
$\pathwise$ estimator with the baseline estimator $\reinforce$. While $\pathwise$ introduces bias into the estimation, we find in our experiments that this bias is small in practice and remains small even over long time horizons. At the same time, the $\pathwise$ estimator delivers dramatic reductions in variance, achieving greater accuracy with a single trajectory than $\reinforce$ with $10^3$ trajectories.

% \hntodo{In Figure~\ref{fig:gradient_comparison}, have a label for the color "cosine similarity to ground-truth" or something like that.}

\begin{figure}[t]
\centering
\hspace{-4em}
\includegraphics[height = 3.2in]{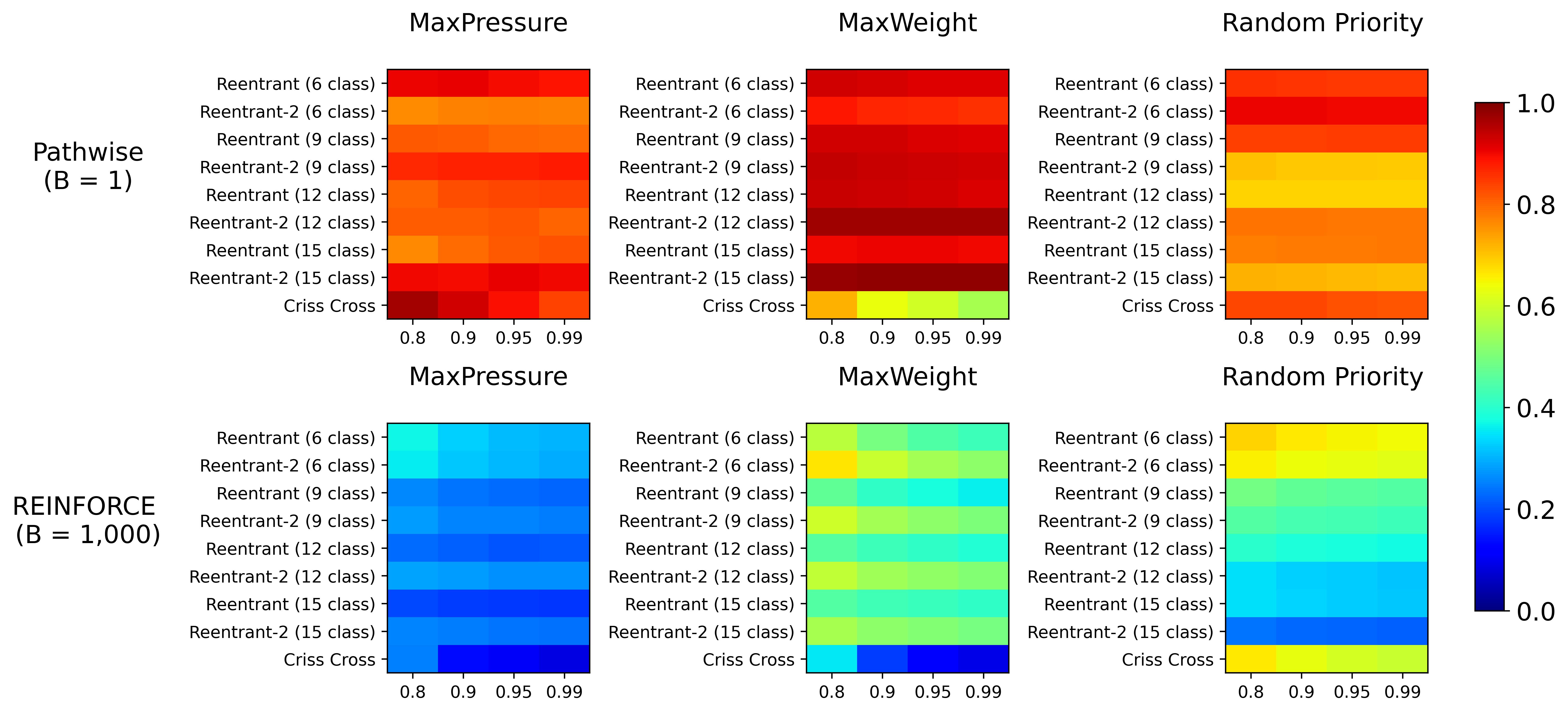}
    
    \caption{Comparison of estimation quality between $\pathwise$ and $\reinforce$ across several settings. We perform this comparison for 3 policies (soft MaxPressure, soft MaxWeight, soft Priority in~\eqref{eqn:soft_policies}), 9 network settings, and 4 levels of traffic intensity for each network. For each cell, we randomly draw $100$ parameters $\theta \in \R^{n}$. For each parameter, we estimate the true gradient by averaging the $\reinforce$ estimator over $10^{6}$ trajectories (with horizon $N = 1000$). We then compute the $\pathwise$ estimator with $B =1$ trajectory along with the $\reinforce$ estimator averaged across $B = 1000$ trajectories. In each instance, we draw 100 samples of these estimators to estimate $\cossim$, i.e., the average cosine similarity with the true gradient. In total, we perform this comparison across $10,800$ parameters. We find that across all of these diverse settings, $\pathwise$ delivers much higher fidelity to the true gradient, in many cases achieving an average cosine similarity close to the maximum value of 1, despite using orders of magnitude less data, whereas the cosine similarity of $\reinforce$ remains around $0.2$-$0.6$ even with 1000 trajectories.}
    \label{fig:gradient_comparison} 
\end{figure}

First, recall that a policy $\pi(x)$ maps queue-lengths $x$ to assignment between servers and queues, represented by an $m \times n$ matrix in $\overline{\mathcal{U}}$ (allowing for fractional routing matrices). We visit three classical queuing policies: priority policies~\cite{cox1961queues}, MaxWeight~\cite{tassiulas1990stability}, and MaxPressure~\cite{dai2005maximum}. Each of these methods selects the routing that solves an optimization problem. This means that the routing generated by the policy is deterministic given the state and is not differentiable in the policy parameters. In order to apply either $\reinforce$ or the $\pathwise$ gradient estimator to compute a policy gradient, we require differentiable surrogates of these policies. To this end, we define softened and parameterized variants of these policies, denoted as soft priority ($\mathsf{sPR}$), soft MaxWeight ($\mathsf{sMW}$), and soft MaxPressure ($\mathsf{sMP}$),
\begin{equation}
\label{eqn:soft_policies}
\pi^{\mathsf{sPR}}_{\theta}(x)_{i}= \softmax(\theta_{j} \cdot \mu_{i}), \quad
\pi^{\mathsf{sMW}}_{\theta}(x)_{i}= 
\softmax(\theta_{j}x_{j} \cdot \mu_{i}), \quad 
\pi^{\mathsf{sMP}}_{\theta}(x)_{i} = \softmax((\mu \odot R(\theta x))_{i})
\end{equation}
where $\theta \in \R_{+}^{n}$ are a vector of costs/weights for each queue, $\mu$ denotes the matrix of service rates with $\mu_{i} \in \R_{+}^{n}$ denoting the service rates associated with server $i$. 
The operation $\odot$ refers to element-wise multiplication and the
$\softmax$ operation maps a vector $a \in \R^{n}$ into a set of probabilities $\softmax(a)_{i} = e^{a_{i}} / \sum_{j=1}^{n} e^{a_{j}}$.

We are interested in identifying the parameter $\theta$ that minimizes long-run average holding cost where $c(x,u) = h^{\top} x$. 
%Note that when $\theta = h$, these correspond with the standard definitions of the above policies. 
We use the objective 
$J_{N}(\theta) = \E\left[\sum_{k=0}^{N-1}c(x_{k},\pi_{\theta}(x_{k}))\tau^{*}_{k+1}\right]$
where $N$ is a large enough number to approximate the long-run performance, and the goal of the gradient estimation is to estimate $\nabla J_{N}(\theta)$.

We consider the following environments, which appear throughout our computational experiments and serve as standard benchmarks for control policies in multi-class queuing networks. We describe the network structure in detail in Figure~\ref{fig:networks}.
\begin{itemize}
\item {\bf Criss-cross:} The network introduced in Example~\ref{example:criss-cross} (see Figure~\ref{fig:networks} (c)).
\item {\bf Re-entrant 1 ($n$ classes):} We consider a family of multi-class re-entrant networks with a varying number of classes, which was studied in~\cite{bertsimas2014robust,dai2022queueing}. The network is composed of several layers and each layer has 3 queues. Jobs processed in one layer are sent to the next layer. Arrivals to the system come to queues 1 and 3 in the first layer while queue 2 receives re-entered jobs from the last layer (see Figure~\ref{fig:networks} (a) for a two-layer example).
\item {\bf Re-entrant 2 ($n$ classes):} We consider another family of re-entrant network architecture that was studied in~\cite{bertsimas2014robust}. It also consists of multiple layers with 3 queues in each layer. It differs from the Re-entrant 1 environment in that only queue 1 receive external arrivals while queues 2 and 3 receive re-entered jobs from the last layer (see Figure~\ref{fig:networks} (b) for a two-layer example).
\end{itemize}

For a gradient estimator $\hat{g}$, the main performance metric we evaluate is $\cossim(\hat{g})$, which is the expected cosine similarity with the ground-truth gradient,
\begin{equation}
\label{eqn:similarity}
\cossim(\hat{g}) \equiv \E[\mathsf{cos} \left( \hat{g}, \nabla J_{N}(\theta) \right)] = \E \left[ \frac{ \langle \hat{g}, \nabla J_{N}(\theta) \rangle}{\norm{\hat{g}} \norm{ \nabla J_{N}(\theta)}} \right] \in [-1,1]
\end{equation}
where the expectation $\E$ is over randomness in $\hat{g}$. The higher the similarity is, the more aligned $\hat{g}$ is to the direction of $\nabla J_{N}(\theta)$. This metric incorporates both bias and variance of the gradient estimator. If the gradient estimator is unbiased but has a high variance, then each individual realization of $\hat{g}$ is likely to have low correlation with the true gradient, so the average cosine similarity will be small even if $\E[\hat{g}]=\nabla J_{N}(\theta)$. At the same time, if the gradient estimator has a low variance but a high bias, then the $\cossim(\hat{g})$ could still be small if $\mathsf{cos} \left(\E[\hat{g}], \nabla J_{N}(\theta) \right)$ is small. We focus on this metric, because it directly determines how informative the gradient estimates are when applying various gradient descent algorithms.  
For our experiments, we evaluate (a close approximation of) the ground-truth gradient $\nabla J_{N}(\theta)$
by using the unbiased $\reinforce$ gradient estimator over exceedingly many trajectories (in our case, $10^{6}$ trajectories). 

We compare the similarity of $\pathwise$ with that of $\reinforce$. 
%where $\reinforce$ is averaged over multiple trajectories. 
We denote $B$ as the number of trajectories we use to calculate each $\pathwise$ or $\reinforce$ gradient estimator.
\begin{align}
\underbrace{\hat{\nabla}_{\theta} J_{N}(\theta; \xi^{(1)}_{1:N})}_{\pathwise\text{ with }B=1} \qquad
\underbrace{\hat{\nabla}^{\mathsf{R}}_{\theta} J_{N,B}(\theta; \xi_{1:N}) := \frac{1}{B} \sum_{b = 1}^{B} \hat{\nabla}^{\mathsf{R}}_{\theta} J_{N}(\theta; \xi^{(b)}_{1:N})}_{\reinforce\text{ with }B\text{ trajectories}}
\end{align}
We compute the $\pathwise$ gradient with only $B = 1$ trajectory, while $\reinforce$ gradient is calculated using $B=10^3$ trajectories.
%to test the improvements in efficiency achieved by the estimator. 
For each policy and setting, we compute these gradients for $100$ different randomly generated values of $\theta$, which are drawn from a $\mathsf{Lognormal}(0,1)$ distribution (as the parameters must be positive in these policies). In total, we compare the gradients in $10,080$ unique parameter settings, and each gradient estimator is computed $100$ times to evaluate the average cosine similarity. When computing the policy gradient, we consider a time horizon of $N = 10^3$ steps. 
% \hntodo{This is great---you should repeat this in abstract, intro, and subsequent sections. This adds credibility.}
% {\color{blue} to approximate the long-run average cost?}.

Figure~\ref{fig:gradient_comparison} compares the $\pathwise$ estimator with $B = 1$ trajectory with the $\reinforce$ estimator averaged over $B = 10^3$ trajectories. For the $\reinforce$ estimator, costs are computed with a discount factor $\gamma = 0.999$, as using a lower discount rate introduced significant bias in the estimation. For $\pathwise$, we use an inverse temperature $\beta = 1$ for the $\softmin$ relaxation across all settings. 
% \hntodo{This part is uber important. See also my comments in the previous section. You need to say this at every section including the abstract, and really highlight that you did not tune this. It needs to come out that you chose $\beta = 1$ because it's a nice round number.} 
Each cell in Figure~\ref{fig:gradient_comparison} corresponds to a (policy, network, traffic-intensity) and the cell value is the average expected cosine similarity of the estimator averaged across the $100$ randomly drawn $\theta$ values. We observe that across these diverse settings, the $\pathwise$ estimator consistently has a much higher average cosine similarity with the true gradient despite using only a single trajectory. In fact, for {\bf 94.5\%} of the 10,800 parameter settings, $\pathwise$ has a higher average cosine similarity with 99\% confidence than $\reinforce$ with $B = 1000$ trajectories. In most cases, the cosine similarity of $\pathwise$ is close to 1, indicating almost perfect alignment with the true gradient even under high congestion. $\reinforce$ on the other hand suffers greatly from high variance. 
% We also observe in the left panel of Figure~\ref{fig:gradient_cossim_eval}, which displays the histogram of the similarity metric for each of the $10,080$ parameter settings, that this improvement in efficiency occurs across almost all parameter settings (with the exception of a few outliers). 
Overall, this demonstrates that $\pathwise$ is able to deliver greater estimation accuracy with an order of magnitude fewer samples.

% {\color{blue} talk about the dial plot}
% \begin{figure}[t]

% \includegraphics[height = 2.2in]{plot/gradient_eval/hist_cosine_sim.png}
% \includegraphics[height = 2.2in]{plot/gradient_eval/cossim_rho.jpg}
    
%     \caption{Cosine similarity of gradient estimators with the true gradient. (Left) Histogram of average cosine similarity for $\reinforce$ ($B = 1000$) and $\pathwise$ ($B = 1$) across $10,800$ parameter settings. $\pathwise$ consistently attains high cosine similarity close to the maximum value of 1. On the other hand, the average cosine similarity of the $\reinforce$ estimator is centered around $[0.25 , 0.5]$. (Right) Average cosine similarity with the true policy gradient for a parameterized max-pressure policy in the criss-cross network. This displays how average cosine similarity degrades with higher traffic intensity. While the cosine similarity of $\pathwise$ with only $B = 1$ trajectory degrades from $\approx 1$ to $\approx 0.8$, $\reinforce$ with $B = 10,000$ trajectories experiences a sharper degradation, from $0.6$ to $0.3$. $\reinforce$ with fewer trajectories does poorly across all traffic intensities.}
%     \label{fig:gradient_cossim_eval} 
% \end{figure}

% \subsection{Optimization tasks}

% Given the strong improvements in estimation efficiency, we now see how they translate to downstream optimization tasks.

\subsection{Learning the $c\mu$ rule}
\label{section:learn_cmu}
Given the strong improvements in estimation efficiency, we turn to evaluate how these translate to a downstream optimization task.
In single-server multi-class queues, it is well-known that the $c\mu$-rule minimizes the long-run average holding cost~\cite{cox1961queues}. We assess whether gradient descent with the $\reinforce$ or $\pathwise$ gradients is capable of converging to the $c\mu$-rule, without knowing the holding costs $h$ or $\mu$ and only using feedback from the environment. Despite its simplicity, it has been observed in prior work that this is a difficult learning task, particularly under heavy traffic~\cite{tran2022finding}.

We revisit the soft priority policy mentioned before, but with only the parameters $\theta\in \R^{n}$, i.e., 
\begin{equation}
\label{eq:soft_priority}
\pi^{\mathsf{sPR}}_{\theta}(x)_{i}= \softmax(\theta_{i})
\end{equation}
We also modify the policy to ensure that it is work-conserving, i.e., not assigning the server to an empty queue (see section~\ref{section:optimization} for further discussion).

We consider a family of multi-class single-server queues with $n$ queues. Holding costs are identically $h_{j} = 1$. Inter-arrival and service times are exponentially distributed, the service rates are $\mu_{1j} = 1 + \epsilon j$, for some $\epsilon>0$, and the arrival rates are identical $\lambda_{j} = \lambda$ and $\lambda$ are set such that the traffic intensity $\sum_{j=1}^{n} \frac{\lambda}{\mu_{1j}} = \rho$
%for $\rho = 0.95$
for some $\rho \in (0,1)$. Note that in this case, the $c\mu$-rule prioritizes queues with higher indices $j$. We consider a grid of gap sizes $\epsilon \in \{1.0, 0.5, 0.1, 0.05, 0.01 \}$ to adjust the difficulty of the problem; the smaller $\epsilon$ is, the harder it is to learn.

We compare $\pathwise$ with $B=1$ trajectory and $\reinforce$ with $B = 100$ trajectories for trajectories of $N = 1000$ steps. In order to isolate the effect of the gradient estimator from the optimization scheme, for both estimators we use an identical stochastic gradient descent scheme with normalized gradients (as these two estimators may differ by a scale factor). That is, for gradient estimator $\hat{g}$, the update under step-size $\alpha$ is
\[
\theta_{t+1} = \theta_{t} - \alpha \frac{\hat{g}}{\norm{\hat{g}}}.
\]
We run $T$ gradient descent steps for each gradient estimator. To allow for the fact that different estimators may have different performances across different step sizes, we consider a grid of step sizes $\alpha \in \{0.01, 0.1, 0.5, 1.0 \}$. Gradient normalization may prevent convergence, so we use the averaged iterate $\bar{\theta}_{T}$ for $T$. We then evaluate the long-run average holding cost under a strict priority policy determined by $\bar{\theta}_{T}$, i.e., $\pi_{\bar{\theta}_{T}}(x)_{i} = \argmax \bar{\theta}_{T,j}$. 
%to compare with the $c \mu$ rule.

% \begin{figure}[t]
% \centering

% \includegraphics[height = 2.3in]{plot/gradient_eval/mm1_buffer_control_sgn.png}

%     \caption{Admission control.}
%     \label{fig:admission} 
% \end{figure}

\begin{figure}[t]
\centering
\includegraphics[height = 2.1in]{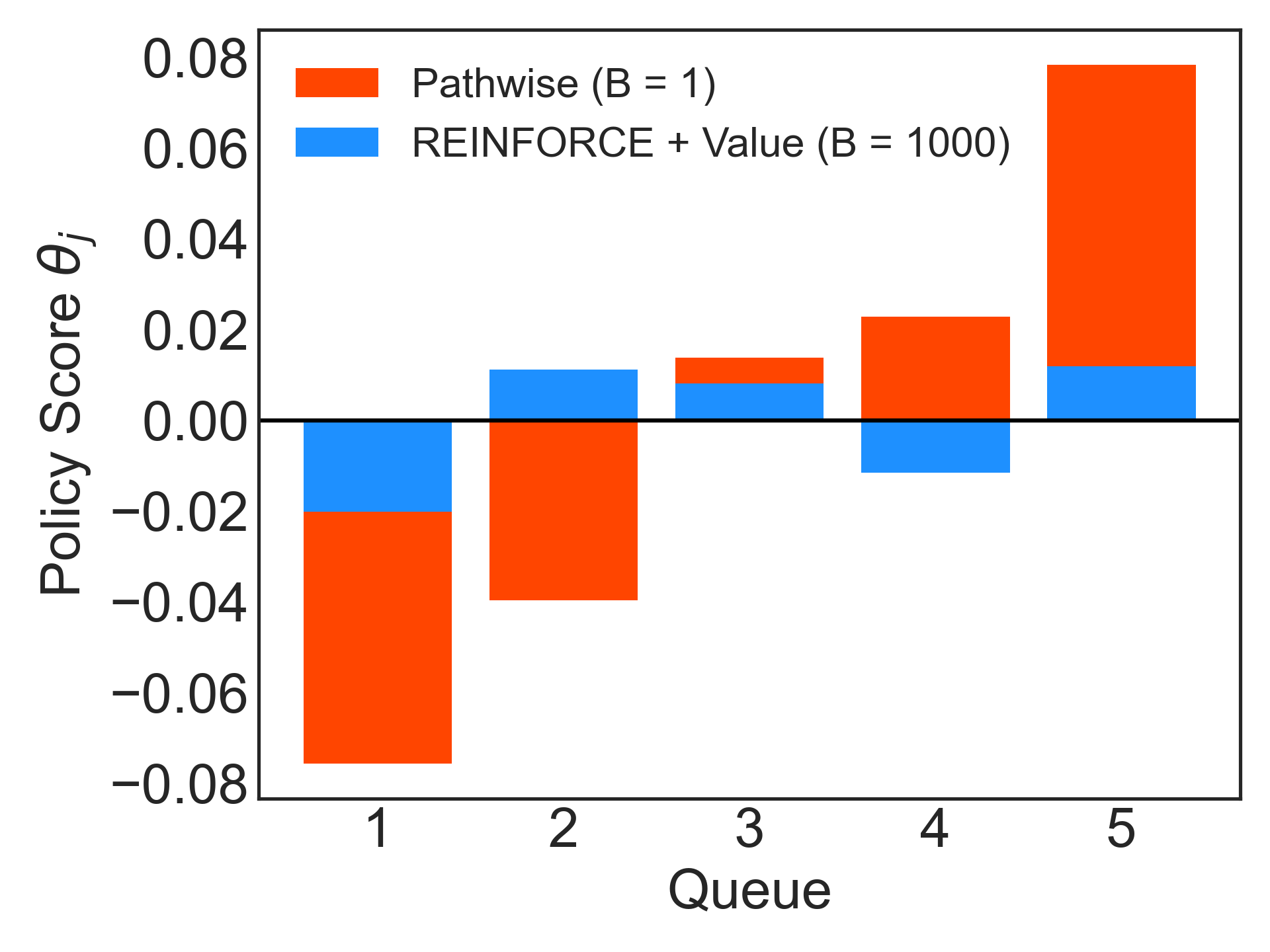}
\includegraphics[height = 2.1in]{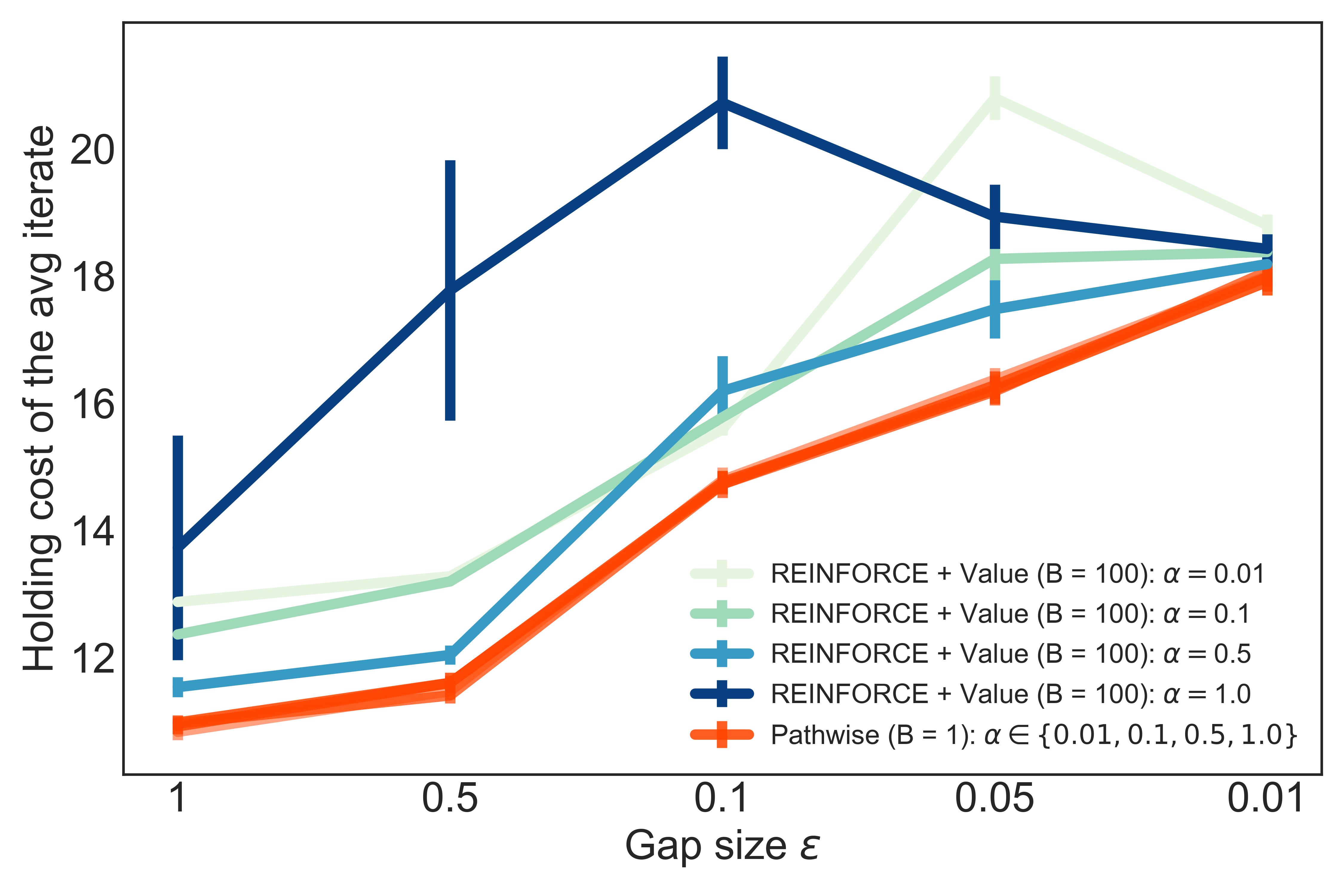}
    
    \caption{Learning the $c\mu$ rule. (Left) Averaged iterate of the policy parameters after 50 gradient steps with $\pathwise$ ($B = 1$) and $\reinforce$ ($B = 100$) gradients for a 5-class queue with traffic intensity $\rho = 0.99$ and gap-size $\epsilon = 0.1$. The scores obtained by $\pathwise$ are increasing in the queue index, which matches the ordering of the $c\mu$ index in this instance. The scores obtained by $\reinforce$ do not achieve the correct ordering despite using more trajectories. (Right) Average holding cost of the average iterate after 20 steps of gradient descent in a 10-class queue with $\rho = 0.95$. This figure reports the results averaged over 50 separate runs of gradient descent, across a grid of step-sizes (denoted as $\alpha$). Remarkably, the optimization performance of the $\pathwise$ estimator is highly similar across step-sizes, and \emph{uniformly} outperforms $\reinforce$ with different step-sizes $\alpha$. 
    %The improvement is higher in more difficult settings (corresponding to a smaller $\epsilon$), although for very small $\epsilon$ both methods have trouble learning the $c\mu$-rule.
    % {\color{blue} zoom into the orange line to see that it does the same across all step sizes}
    }
    \label{fig:cmu} 
\end{figure}

% {\color{blue} mention that pathwise the performance is basically the same across all step-sizes}

The left panel of Figure~\ref{fig:cmu} displays the values of $\bar{\theta}_{T}$ after $T=50$ gradient iterates for $\pathwise$ and $\reinforce$ with $n=5$, $\epsilon = 0.1$, and $\rho = 0.99$. We observe that while $\pathwise$ sorts the queues in the correct order (it should be increasing with the queue index), $\reinforce$ even with $B = 100$ trajectories fails to prioritize queues with a higher $c \mu$ index. Remarkably, we observe in the right panel of the same figure that $\pathwise$ with just a single trajectory achieves a lower average holding cost than $\reinforce$ {\bf uniformly} across various step sizes and difficulty levels, whereas the performance of $\reinforce$ varies greatly depending on the step size. This indicates that the improvements in gradient efficiency/accuracy of $\pathwise$ make it more robust to the step-size hyper-parameter. 
It is also worth mentioning that when gap size $\epsilon$ becomes smaller, it is more difficult to learn. At the same time, since $\mu_{1j}$'s are more similar to each other, the cost difference between different priority rules also diminishes.
%It is worth mentioning that while the cost reduction from $\pathwise$ is significant for most gap sizes $\epsilon$, for very small $\epsilon$ it becomes difficult to either method to identify the queue with the highest index, and the performance benefits from doing so also diminish.

% {\color{blue} mention that performance is the same for all $\epsilon$ because the problem gets too hard. and practically theres no difference}

\subsection{Admission Control}
\label{section:admission}

While we focus mainly on scheduling tasks in this work, our gradient estimation framework can also be applied to admission control, which is another fundamental queuing control task~\cite{naor1969regulation, chr1972individual, ccil2009effects, ghosh2007optimal, koccauga2010admission}. To manage congestion, the queuing network may reject new arrivals to the network if the queue lengths are above certain thresholds. The admission or buffer control problem is to select these thresholds to balance the trade-off between managing congestion and ensuring sufficient resource utilization.

Under fixed buffer sizes $\buff = \{\buff_{j}\}_{j=1}^{n}$, new arrivals to queue $j$ are blocked if $x_{j} = \buff_{j}$. As a result, the state update is modified as follows,
\begin{equation}
x_{k+1} = \min\{x_{k} + De_{k+1}, \buff\}.
\end{equation}
% \hntodo{Let's not use $\beta$ for the buffer; overlaps with the inverse temp. }
While a small $\buff$ can greatly reduce congestion, it can impede the system throughput. To account for this, we introduce a cost for rejecting an arrival to the network. Let $o_{k} \in \{0,1\}^{n}$ denote whether an arrival is \emph{overflowed}, i.e., an arrival is blocked because the buffer is full,
\begin{equation}
o_{k+1} = De_{k+1} \cdot 1\{
x_{k} + De_{k+1} > \buff \}.
\end{equation}
Given a fixed routing policy, the control task is to choose the buffer sizes $\buff$ to minimize the holding and overflow costs:
\begin{align}
%c(x, \buff) &= h^{\top}x + b^{\top}z \\
J_{N}(\buff;\xi_{1:N}) &=  \sum_{k=0}^{N-1}  (h^{\top}x_{k})\tau_{k+1}^{*} + b^{\top}o_{k}.
\end{align}
Similar to the routing control problem, despite the fact that overflow is discrete, our gradient estimation framework is capable of computing a $\pathwise$ gradient of the cost with respect to the buffer sizes, which we denote as $\widehat{\nabla}_{\buff} J_{N}(\buff;\xi_{1:N})$, i.e., we can evaluate gradients at integral values of the buffer size and use this to perform updates. Since the buffer sizes must be integral, 
%and Using the $\pathwise$ gradient $\widehat{\nabla}_{\buff} J_{N}(\buff;\xi_{1:N})$, 
we update the buffer sizes via sign gradient descent to preserve integrality:
\begin{equation}
\label{eq:sign_gd}
\buff_{t+1} = \buff_{t} - \mathsf{sign} \left( \widehat{\nabla}_{\buff} J_{N}(\buff;\xi_{1:N}) \right).
\end{equation}
% One could also optimize over purely integral values through sign gradient descent, but this achieves worse performance 
% \hntodo{In Figure~\ref{fig:admission}, explicitly write SGD stepsize instead of $\alpha$}

\begin{figure}[t]
\centering
\includegraphics[height = 2.3in]{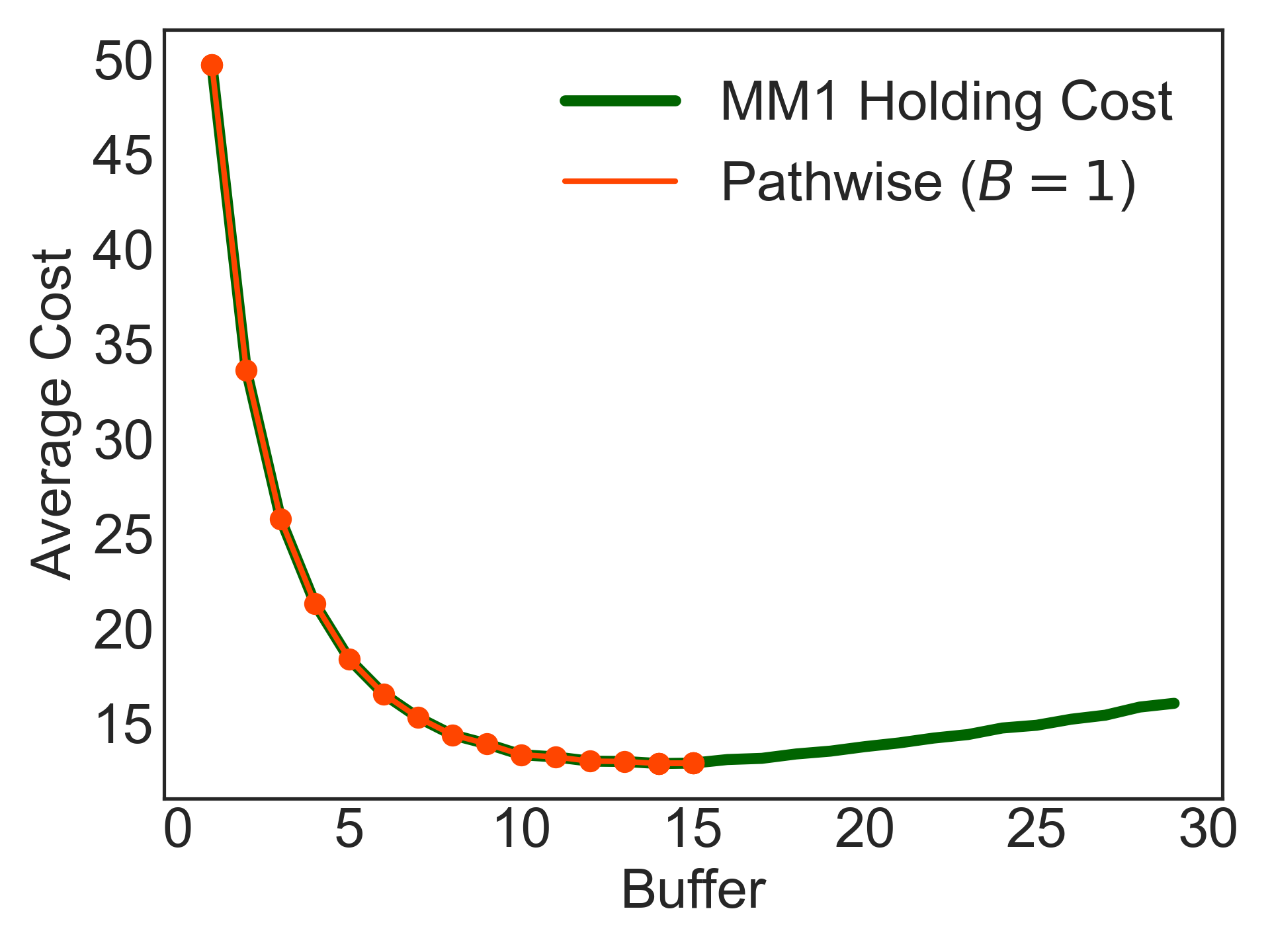}
\includegraphics[height = 2.5in]{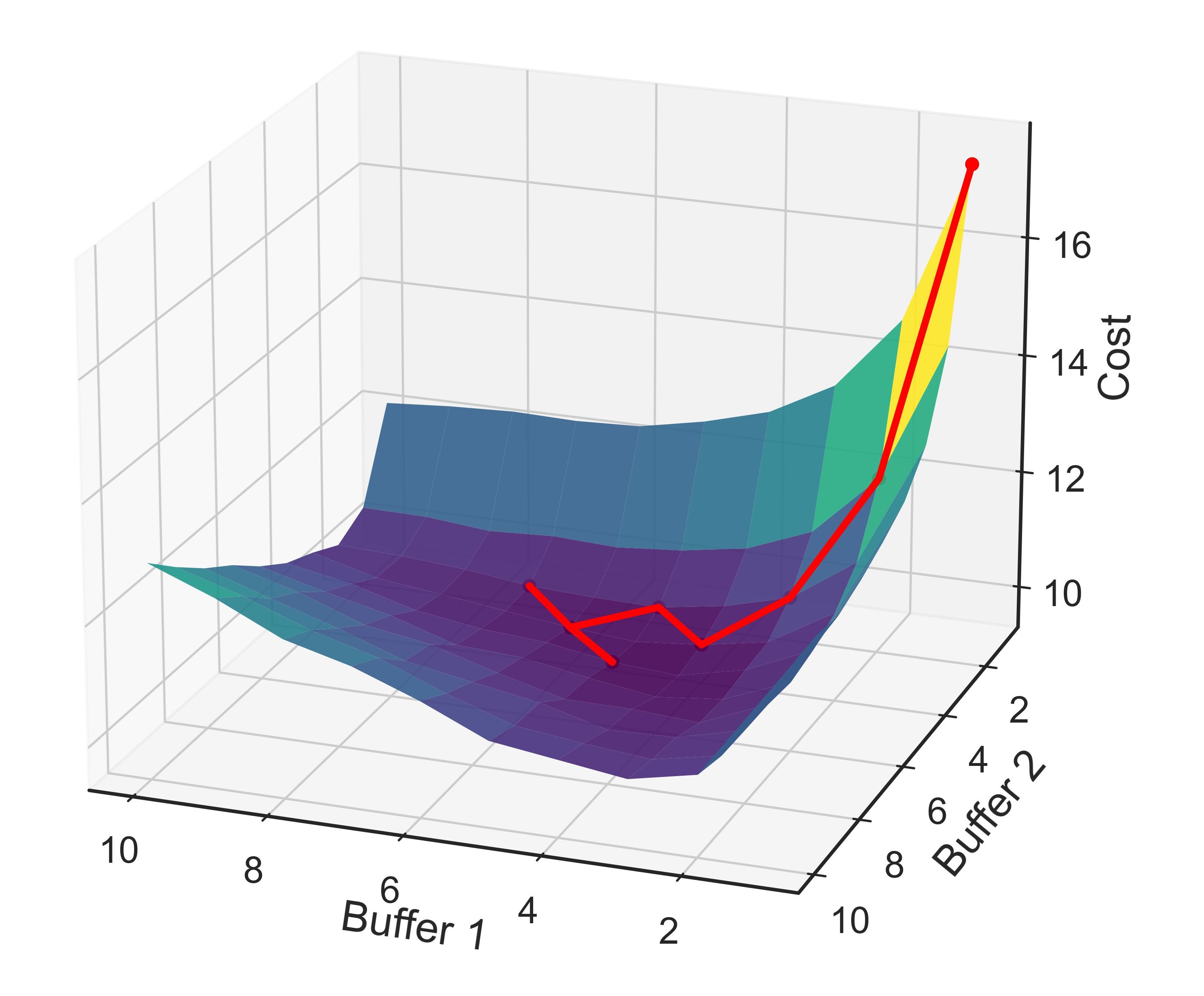}
    
    \caption{Gradient descent with $\pathwise$ for admission control tasks. (Left) Iterates of sign gradient descent with the $\pathwise$ estimator for the buffer size of an $M/M/1$ queue with holding cost $h = 1$ and overflow cost $b = 100$. Starting from $L_{0} = 1$, the iterates quickly converge to the optimal level of $L^{*} = 14$. (Right) buffer sizes obtained by sign gradient descent with the $\pathwise$ estimator for a 2 queue 1 server network with holding cost $h = 1$ and overflow cost $b = 20$. Starting from $L_{0} = (1,1)$, the iterates quickly converge to the basin of the loss surface. }
    \label{fig:admission} 
\end{figure}

Learning for admission control has been studied in the queuing and simulation literature~\cite{cassandras2002perturbation, cassandras2003perturbation, ho1983new, cohen2024learning}. While exact gradient methods are possible in fluid models~\cite{cassandras2002perturbation, cassandras2003perturbation}, the standard approach for discrete queuing models is finite perturbation analysis~\cite{ho1983new}, given the discrete nature of the buffer sizes.
Randomized finite-differences, which is also known as Simultaneous Perturbation Stochastic Approximation (SPSA)~\cite{spall1992multivariate, fu1997optimization}, is a popular optimization method for discrete search problems. This method forms a finite-differences gradient through a random perturbation. Let $\eta \sim \mathsf{Rademacher}(n, 1/2) \in \{-1,1 \}^{n}$ be a random $n$-dimensional vector where each component is an independent $\mathsf{Rademacher}$ random variable, taking values in $\{-1,1 \}$ with equal probability. For each perturbation $\eta$, we evaluate the objective at $\buff \pm \eta$, i.e., $J_N(\buff + \eta; \xi_{1:N})$ and $J_N(\buff - \eta; \xi_{1:N})$, using the same sample path for both evaluations to reduce variance. For improved performance, we average the gradient across a batch of $B$ perturbations, i.e., $\eta^{(b)}$ for $b=1,...,B$, drawing a new sample path $\xi^{(b)}_{1:N}$ for each perturbation. The batch SPSA gradient is 
\begin{equation}
\widehat{\nabla}_{\buff}^{\mathsf{SPSA},B} J_{N}(\buff) = \frac{1}{B} \sum_{b=1}^{B} \frac{1}{2}\left( J_N(\buff + \eta^{(b)}; \xi^{(b)}_{1:N}) - J_N(\buff - \eta^{(b)}; \xi^{(b)}_{1:N}) \right) \eta^{(b)}.
\end{equation}
We update the buffer sizes $\buff$ according to the same sign gradient descent algorithm as in \eqref{eq:sign_gd}. 
%to isolate the effect of the gradient estimator.

In comparison with existing works in the queuing literature (e.g.~\cite{naor1969regulation, ccil2009effects, ghosh2007optimal}), which derive analytical results for simple single-class or multi-class queues, we consider admission control tasks for large, re-entrant networks with multiple job classes. Each job class has its own buffer, resulting in a high-dimensional optimization problem in large networks. Moreover, the buffer size for one job class affects downstream congestion due to the re-entrant nature of the networks. For our experiments, we fix the scheduling policy to be the soft priority policy $\pi^{\mathsf{sPR}}_{\theta}(x)$ in~\eqref{eq:soft_priority} due to its simplicity and strong performance in our environments. We emphasize however that our framework can be applied to buffer control tasks under any differentiable routing policy, including neural network policies. For each gradient estimator, we perform $T=100$ iterations of sign gradient descent, and each gradient estimator is computed from trajectories of length $N = 1000$. For SPSA, we consider batch sizes of $B = \{10 , 100, 1000\}$ whereas we compute $\pathwise$ with only $B=1$ trajectory. When evaluating the performance, we calculate the long-run average cost with the buffer size determined by the last iterate with a longer horizon $N = 10^4$ and over 100 trajectories. We also average the results across $50$ runs of sign gradient descent.

The left panel of Figure~\ref{fig:admission} displays iterates of the sign gradient descent algorithm with  $\pathwise$ for the $M/M/1$ queue with holding cost $h = 1$ and overflow cost $b = 100$. We observe that sign gradient descent with $\pathwise$ (computed over a horizon of $N=1000$ steps) quickly reaches the optimal buffer size of $L^{*} = 14$ and remains there, oscillating between $L = 14$ and $15$. The right panel shows the iterates for a simple 2-class queue with 1 server, $h = 1$, and $b = 20$ under a soft priority policy. We again observe that sign gradient descent with $\pathwise$  quickly converges to a near-optimal set of buffer sizes.

\begin{figure}[t]
\centering
\includegraphics[height = 2.3in]{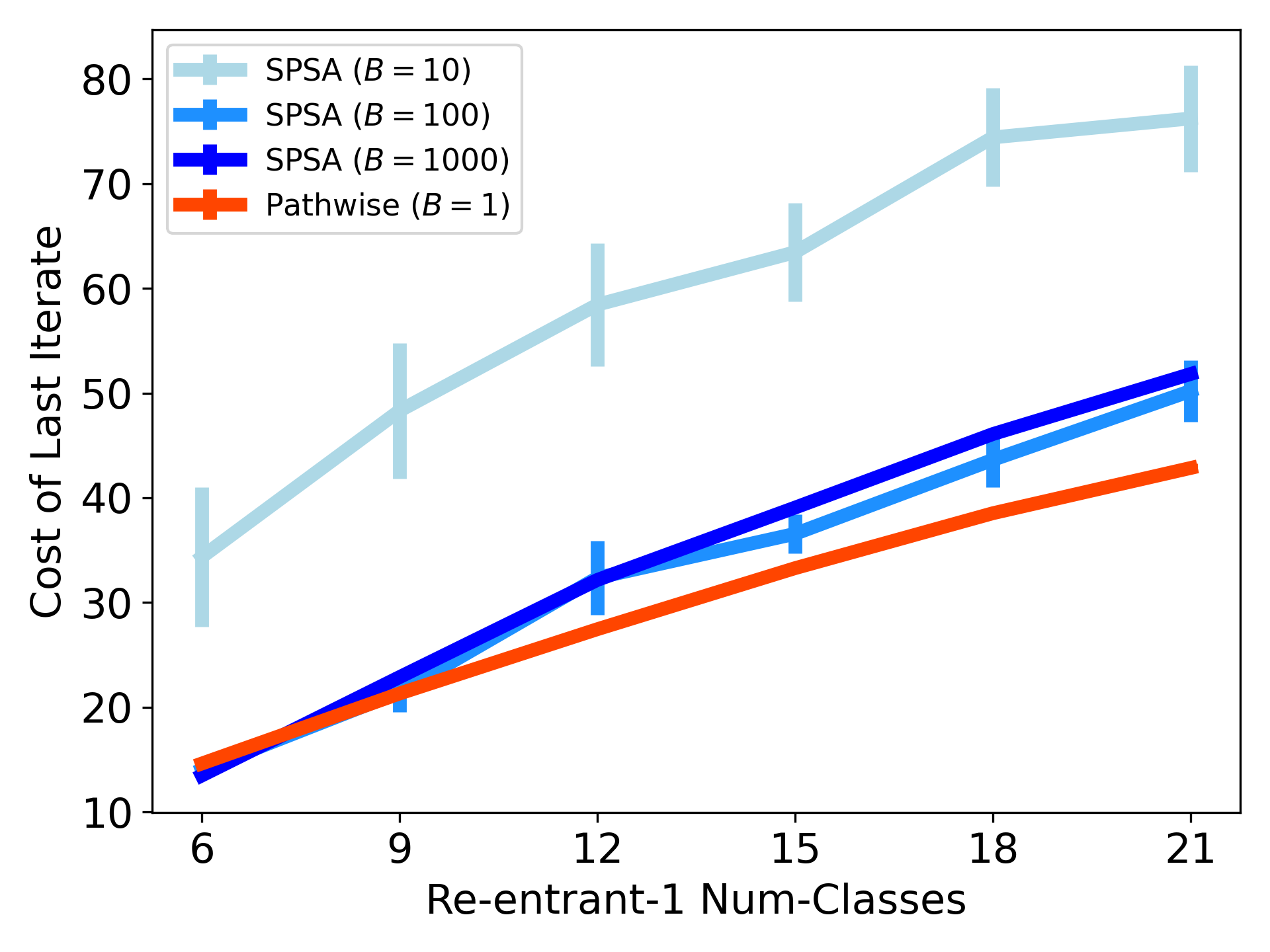}
\includegraphics[height = 2.3in]{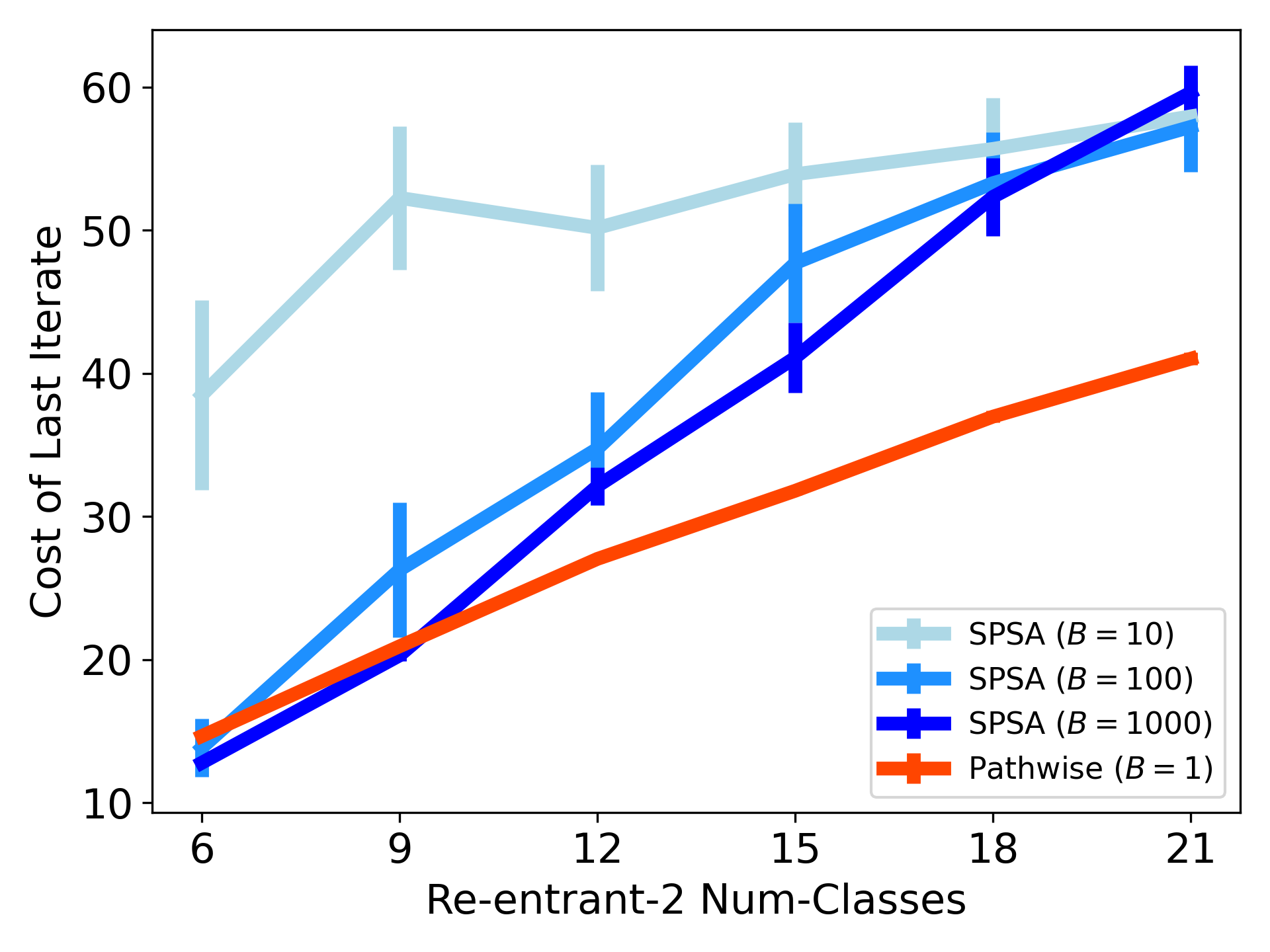}
    
\caption{Last iterate performance of $\pathwise$ ($\beta = 1$) and SPSA~\cite{spall1992multivariate} on admission control for re-entrant networks. (Left) Average cost of the last iterate of sign gradient descent (100 iterations) for the Re-entrant 1 networks, across different numbers of job classes. While SPSA averaged across $B = 100$ trajectories achieves good performance, using a smaller batch $B = 10$ leads to instabilities that lead to much higher costs. $\pathwise$ with only 1 trajectory is able to consistently achieve a smaller cost, especially for larger networks. (Right) Average cost of the last iterate of sign gradient descent (100 iterations) for the Re-entrant 2 networks. Even with $B = 1000$ trajectories, SPSA is unable to effectively optimize the buffer sizes for larger networks, reaching the same cost as SPSA with $B = 10$. This illustrates that the performance of finite-difference methods such as SPSA degrade in higher-dimensional problems, whereas $\pathwise$ performs well in these larger instances using much less data.}
    \label{fig:pathwise_spsa} 
\end{figure}

To see how the estimator performs in larger-scale problems, we consider the Re-entrant 1 and Re-entrant 2 networks introduced in Section~\ref{sec:gradient_efficiency} with varying number of job classes (i.e., varying number of layers). 
%Since the control problem involves a buffer size for each job class, larger networks involve higher-dimensional control problems. 
Figure~\ref{fig:pathwise_spsa} compares the last iterate performance of SPSA and $\pathwise$ for these two families of queuing networks with instances ranging from $6$-classes to $21$-classes. Holding costs are $h = 1$ and overflow costs are $b = 1000$ for all queues. We observe that $\pathwise$ with only a single trajectory is able to outperform SPSA with $B = 1000$ trajectories for larger networks. Sign gradient descent using SPSA with only $B = 10$ trajectories is much less stable, with several of the iterations reaching a sub-optimal set of buffer sizes that assign $L_{j} = 0$ to several queues. This illustrates the well-known fact that for high-dimensional control problems, zeroth-order methods like SPSA must sample many more trajectories to cover the policy space and their performance can scale sub-optimally in the dimension. Yet $\pathwise$, which is an approximate first-order gradient estimator, exhibits much better scalability with dimension and is able to optimize the buffer sizes with much less data.

\begin{figure}[t]
\centering
\includegraphics[height = 2.5in]{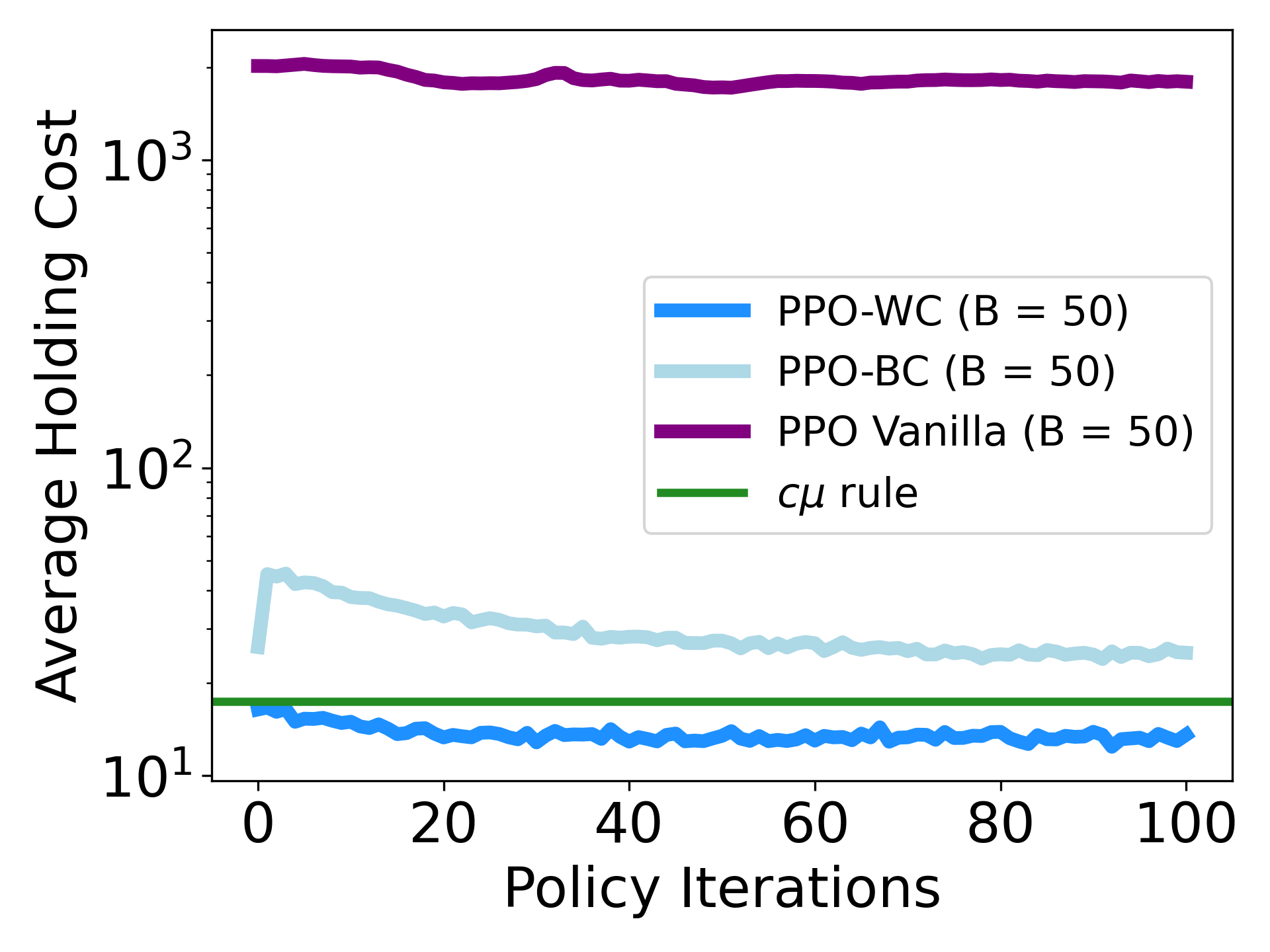}

    \caption{Performance of PPO with and without work-conserving softmax for the Re-entrant 1 network with 6 classes. Average holding cost of PPO without any modifications, PPO initialized from a behavior cloned policy (PPO BC), and PPO with the work-conserving softmax (PPO WC). Average cost of the $c\mu$-rule added for reference. Without any modification, PPO is unable to stabilize the queues, resulting in an average queue-length of $10^{3}$ and does not improve over time. Behavior cloning provides a much better initialization, and the policy improves with training. However, it fails to improve over the $c\mu$-rule. With the work conserving softmax, even the randomly initialized policy is capable of stabilizing the network -- achieving an equivalent cost as the $c\mu$-rule -- and is able to outperform the $c\mu$ over the course of training.}
    \label{fig:work-conserving} 
\end{figure}

\section{Policy Parameterization}
\label{section:optimization}

% While our gradient estimation framework offers a sample-efficient alternative for learning from the environment, there is another practical issue that degrades the performance of learning algorithms for queuing control: stability. It has been observed that standard reinforcement learning algorithms, which are usually initialized at a random policy, can fail to stabilize queuing networks. As a result, researchers have proposed modifications to ensure stability, including behavior cloning of a stabilizing policy~\cite{dai2022queueing}, switching to a stabilizing policy if the queue-lengths exceed a specified level~\cite{liu2022rl}, or shaping the costs to be stability-aware. 
While our gradient estimation framework 
offers a sample-efficient alternative for 
learning from the environment, there is another practical 
issue that degrades the performance of learning algorithms for 
queuing network control: instability. Standard model-free RL algorithms are based on the `tabula rasa' principle, which aims to search over a general and unstructured policy class in order to find an optimal policy. However, it has been observed that this approach may be unsuitable for queuing network control. Due to the lack of structure, the policies visited by the algorithm often fail to stabilize the network,
which prevents the algorithm from learning and improving.
As a result, researchers have proposed structural modifications to ensure stability, 
including behavior cloning of a stabilizing policy to find a good initialization~\cite{dai2022queueing}, 
switching to a stabilizing policy 
if the queue lengths exceed some finite thresholds~\cite{liu2022rl}, 
or modifying the costs to be stability-aware~\cite{pavse2024learning}. 
% \hntodo{CITE}

We investigate the source of instability in various queuing scheduling problems and find a possible explanation. 
We note that many policies obtained by model-free RL algorithms are not {\bf work-conserving} and often allocate servers to empty queues. A scheduling policy is work-conserving if it always keeps the server(s) busy when there are compatible jobs waiting to be served.
%Work-conservation is a well-studied propertyin the queuing literature, and is usually a pre-condition for a policy to induce stability in the queuing network. 
Standard policies such as the $c\mu$-rule, MaxPressure, and MaxWeight are all work-conserving, which partly explains their success in stabilizing complex networks. We treat work conservation as an `inductive bias'
% \hntodo{Most queueing and applied probability people will have no idea what this word means. Spend 2-3 sentences providing background on how this is the key modeling lever in ML and has been responsible for most practical success.} 
and consider a simple modification to the policy architecture that guarantees this property without sacrificing the flexibility of the policy class. 

The de-facto approach for parameterizing policies in deep reinforcement learning is to consider a function $\nu_{\theta}(x)$,
which belongs to a general function family, such as neural networks, and outputs real-valued scores.
These scores are then fed into a $\softmax$ layer, which converts the scores to probabilities over actions.
Naively, the number of possible routing actions can grow exponentially in the number of queues and servers.
Nonetheless, one can efficiently sample from the action space by having the output of
$\nu_{\theta}(x) \in \R^{m\times n}$ be a matrix where row $i$, denote as $\nu_{\theta}(x)_{i}$, contains the scores for matching server $i$ to different queues. Then by applying the $\softmax$ for row $i$, i.e., $\softmax(\nu_{\theta}(x)_{i})$, we obtain the probability that server $i$ is assigned to each queue. We then sample the assignment independently for each server to obtain an action in $\mathcal{U}$.
For the purpose of computing the $\pathwise$ estimator, $\softmax(\nu_{\theta}(x)_{i})$ also gives a valid fractional routing in $\overline{\mathcal{U}}$. We let $\softmax(\nu_{\theta}(x))\in \overline{\mathcal{U}}$ denote the matrix formed by applying the softmax to each row in $\nu_{\theta}(x)$.

Under this `vanilla' softmax policy, the probability $\pi_{\theta}(x)_{ij}$ that server $i$ is routed to queue $j$ 
(or alternatively, the fractional capacity server $i$ allocated to $j$) is given by
\begin{equation}
\label{eq:vanilla_softmax}
\pi_{\theta}(x)_{ij} = \softmax(\nu_{\theta}(x))_{ij} = \frac{e^{\nu_{\theta}(x)_{ij}}}{\sum_{j=1}^{n} e^{\nu_{\theta}(x)_{ij}}}. \hspace{-3em}
\tag{Vanilla Softmax}
\end{equation}
Many of the policies mentioned earlier can be defined in this way, such as the soft MaxWeight policy, $\nu_{\theta}(x)_{i} = \{ \theta_{j}x_{j}\mu_{ij}\}_{j=1}^{n}$. This parameterization is highly flexible
and $\nu_{\theta}(x)$
can be the output of a neural network. However, for a general $\nu_{\theta}(x)$,
there is no guarantee that $\pi(\theta)(x)_{i,j} = 0$ if $x_{j} = 0$. This means that such policies
may waste service capacity by allocating capacity to empty queues even when there are non-empty queues that 
server $i$ could work on.

We propose a simple fix, which reshapes the actions produced by the policy.
We refer to this as the {\bf work-conserving $\softmax$},
% \[
% \bar{\pi}_{\theta}(x)_{i,j} \proto \pi_{\theta}(x)_{i,j}1{x_{j} > 0} \wedge \epsilon \propto e^{\nu_{\theta}(x)_{i,j}}1{x_{j} > 0} \wedge \epsilon
% \]
\begin{equation}
\label{eq:work-conserving}
    \pi_{\theta}^{\mathsf{WC}}(x)_{ij} = \mathsf{softmaxWC}(\nu_{\theta}(x))_{ij} \equiv \frac{e^{\nu_{\theta}(x)_{ij}}1\{x_{j} > 0\}\wedge \epsilon}
{\sum_{l=1}^{n} e^{\nu_{\theta}(x)_{il}}1\{x_{l} > 0\}\wedge \epsilon},
\hspace{-3em}
\tag{WC-Softmax}
\end{equation}
where $\wedge$ is the minimum and $\epsilon$ is a small number to prevent
division by zero when the queue lengths are all zero.

This parameterization is fully compatible with deep reinforcement learning approaches.
$\nu_{\theta}(x)$ can be a neural network and critically, the work-conserving $\softmax$
preserves the differentiability of $\pi^{\mathsf{WC}}_{\theta}(x)$ with respect to $\theta$.
As a result, $\reinforce$ and $\pathwise$ estimators can both be computed under this parameterization, since $\nabla_{\theta} \log \pi_{\theta}^{\mathsf{WC}}(x)$ and $\nabla_{\theta} \pi_{\theta}^{\mathsf{WC}}(x)$
both exist.

This simple modification delivers substantial improvements in performance. Figure~\ref{fig:work-conserving} compares the average holding cost across policy iterations for PPO without any modifications, PPO initialized with a policy trained to imitate MaxWeight, and PPO with the work-conserving $\softmax$. Despite its empirical success in many other reinforcement learning problems, PPO without any modifications fails to stabilize the network and incurs an exceedingly high cost. It performs much better under an initial behavioral cloning step, which achieves stability but still underperforms the $c\mu$-rule. On the other hand, with the work-conserving $\softmax$, even the randomly initialized policy stabilizes the network and outperforms the $c\mu$-rule over the course of training. This illustrates that an appropriate choice of policy architecture, motivated by queuing theory, is decisive in enabling learning-based approaches to succeed. As a result, for all of the policy optimization experiments in sections~\ref{section:learn_cmu},~\ref{section:admission}, and~\ref{section:results}, we equip the policy parameterization with the work-conserving $\softmax$.

\begin{figure}[t]
\centering
\includegraphics[height = 1.3in]{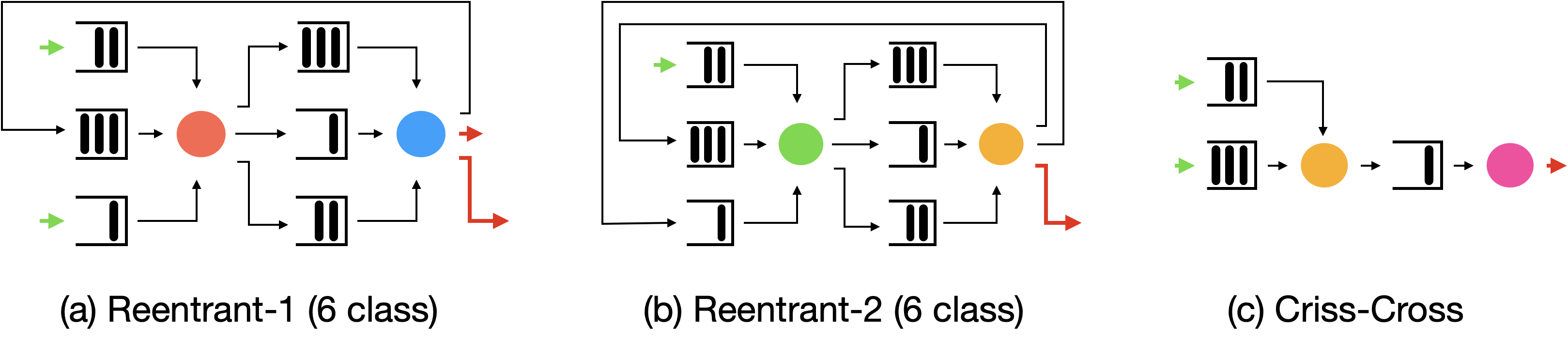} 
        \caption{Multi-class queuing networks. This figure displays the architectures for the networks considered in sections~\ref{section:gradient_eval} and~\ref{section:results}, and have appeared in previous works (see~\cite{dai2022queueing, bertsimas2014robust}). This displays the re-entrant networks with 6 job classes but we also consider networks with a larger number of job classes/queues. For both of these network architectures, the network with $n$ job classes has $n/3$ servers, and each job must be processed sequentially by all servers. Each server serves 3 job classes, and some of the jobs processed by the last server are fed back into the front of the re-entrant network.}

    \label{fig:networks} 
\end{figure}

\section{Scheduling for Multi-Class Queuing Networks: Benchmarks}
\label{section:results}

We now benchmark the performance of the policies obtained by $\pathwise$ policy gradient (Algorithm~\ref{alg:pathwise-pg}) with standard queuing policies and policies obtained using state-of-the-art model-free reinforcement learning algorithms.
We consider networks displayed in Figure~\ref{fig:networks}, which were briefly described in section~\ref{sec:gradient_efficiency} and appeared in previous works~\cite{dai2022queueing, bertsimas2014robust}. \citet{dai2022queueing} used Criss-cross and Re-entrant-1 networks to show that $\ppo$ can outperform standard queuing policies. \citet{bertsimas2014robust} consider the Re-entrant-2 network, but did not include any RL baselines. We consider networks with exponential inter-arrival times and workloads in order to compare with previous results. We also consider hyper-exponential distributions to model settings with higher coefficients of variation, as has been observed in real applications~\cite{green2007coping}. The hyper-exponential distribution $X\sim \mathsf{HyperExp}(\lambda_{1},\lambda_{2},p)$ is a mixture of exponential distributions:
\[
X \stackrel{d}{=} Y\cdot E_{1} + (1 - Y) \cdot E_{2},
\]
for $Y\sim\mathsf{Bernoulli}(p)$, $E_{1} \sim \mathsf{Exp}(\lambda_{1})$, $E_{2} \sim \mathsf{Exp}(\lambda_{2})$, and all are drawn independently of each other. We calibrate the parameters of the hyper-exponential distribution to have the same mean as the corresponding exponential distribution, but with a 1.5x higher variance.
\begin{figure}[t]
\centering
\includegraphics[height = 2.2in]{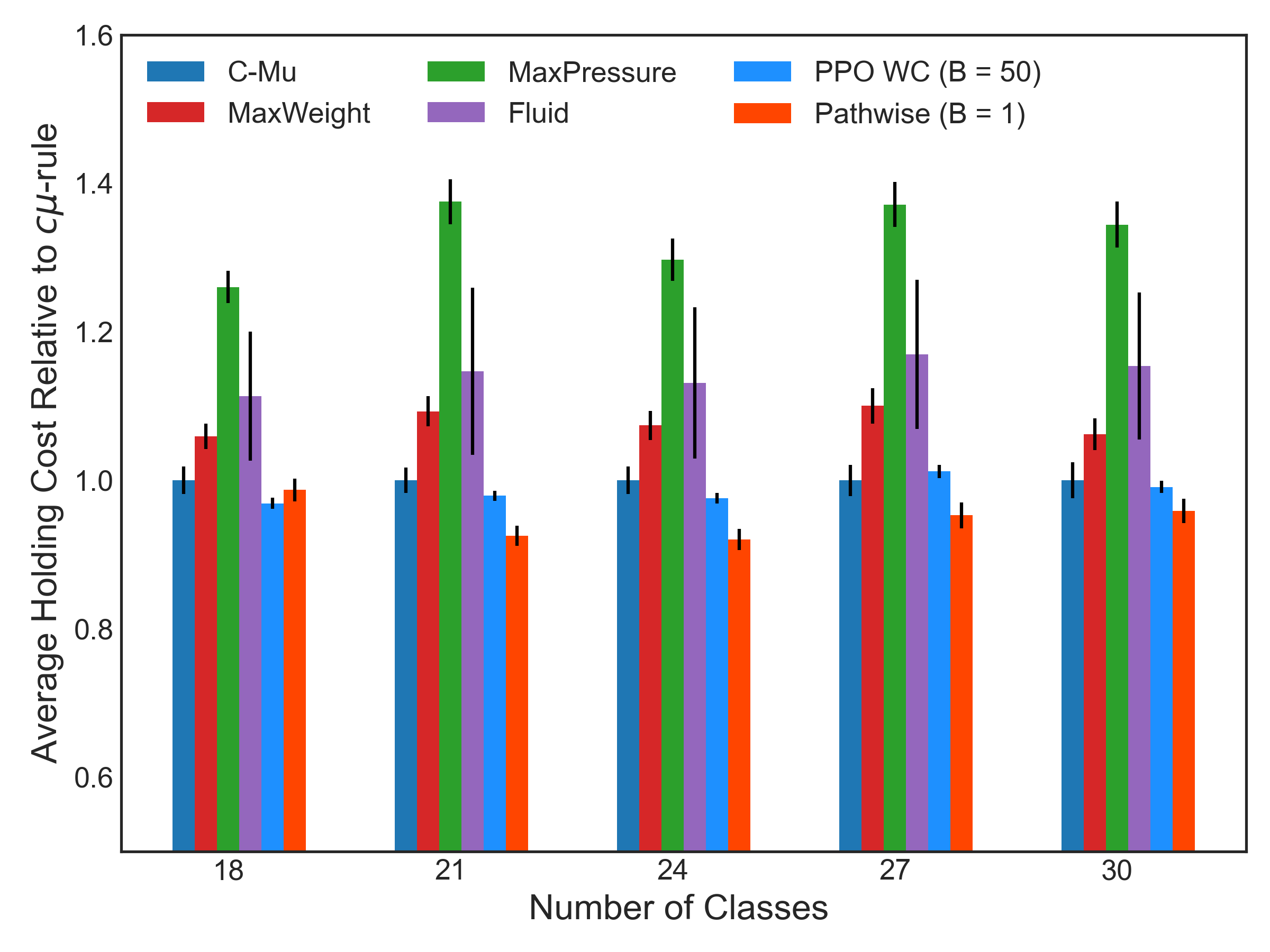}
\includegraphics[height = 2.2in]{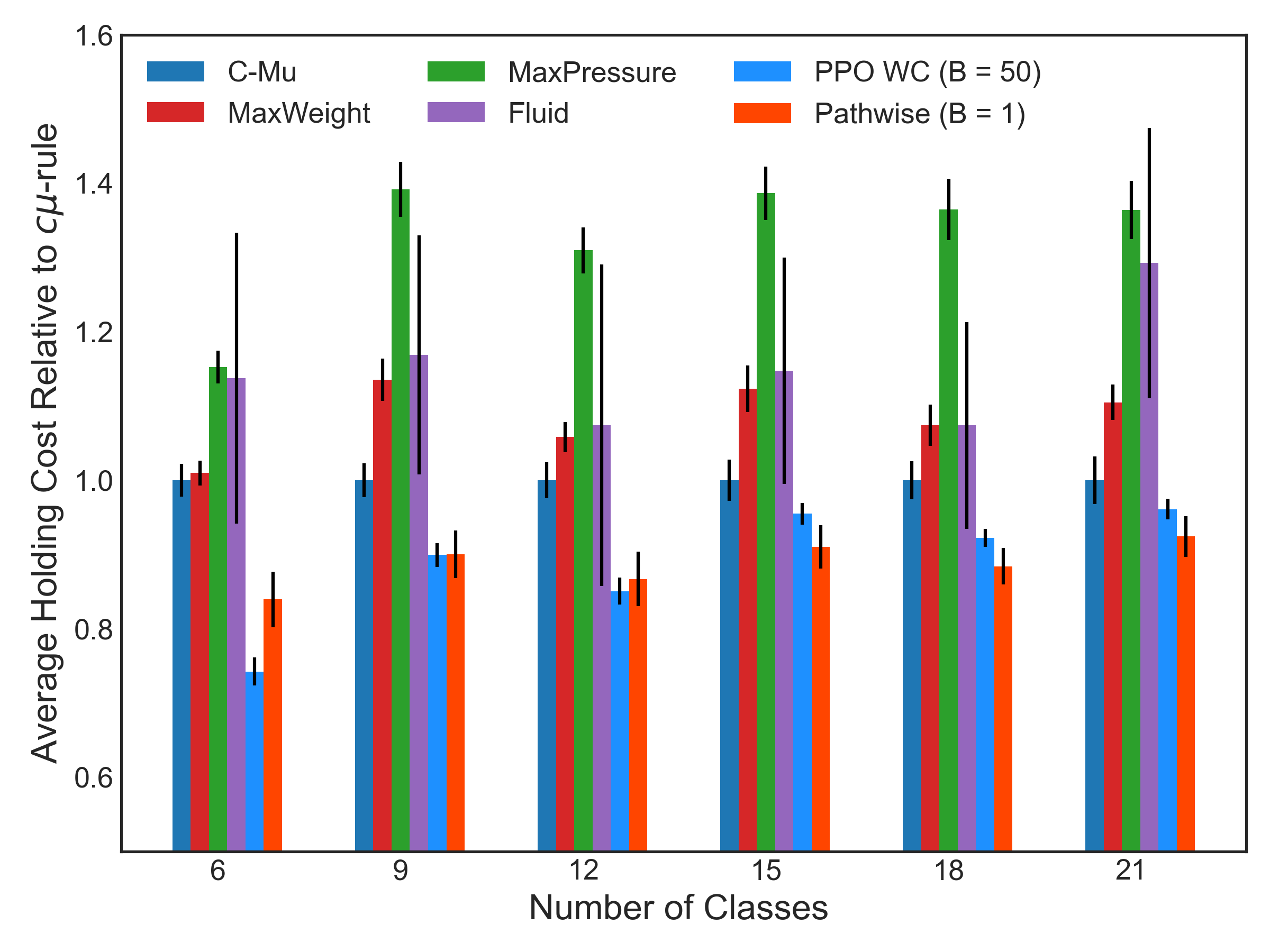}
% \includegraphics[height = 1.6in]{plot/results/reentrant_5.png}
% \includegraphics[height = 1.6in]{plot/results/reentrant_6.png}
% \includegraphics[height = 1.6in]{plot/results/reentrant_7.png}

% \includegraphics[height = 1.6in]{plot/results/reentrant_8.png}
% \includegraphics[height = 1.6in]{plot/results/reentrant_9.png}
% \includegraphics[height = 1.6in]{plot/results/reentrant_10.png}

% \caption{Comparison of policy iterations for $\ppowc$ and $\pathwise$ policy gradient (Algorithm~\ref{alg:pathwise-pg}) across several Re-entrant-1 networks. From top-left to bottom-right: Re-entrant-1 networks with 15, 18, 21, 24, 27, and 30 job classes. We run 100 policy iterations for both algorithms. $\pathwise$ uses a single trajectory for gradient estimation, whereas $\ppowc$ uses all trajectories drawn from $B = 50$ parallel actors. We plot the average holding cost of the $c\mu$-rule for comparison.}

\caption{Comparison of average holding cost of queuing control policies relative to $c\mu$-rule, including $\ppowc$ (ours) and $\pathwise$ (Algorithm~\ref{alg:pathwise-pg}). (Left) Average holding cost for the Re-entrant-1 networks with 18, 21, 24, 27, and 30 job classes. (Right) Average holding cost for the Re-entrant-1 networks with hyper-exponential inter-arrival and service times with 6, 9, 12, 14, 18, 21 job classes. $\pathwise$ outperforms $\ppowc$ for large networks in particular.}

    \label{fig:policy_iterations} 
\end{figure}
Our empirical validation goes beyond the typical settings studied in the reinforcement learning for queuing literature, and is enabled by our discrete-event simulation framework.

\begin{table*}[h!]
\centering \caption{Criss Cross}
\vspace{1em}
\label{table:criss-cross} %
\begin{tabular}{ccccccccc}
\toprule 
{\footnotesize{}Noise } & {\footnotesize{}$c\mu$ } & {\footnotesize{}MaxWeight } & {\footnotesize{}MaxPressure} & {\footnotesize{}Fluid } & {\footnotesize{}$\mathsf{PPO}$-$\mathsf{DG}$~\cite{dai2022queueing}} & {\footnotesize{}$\mathsf{PPO}$-$\mathsf{WC}$} & {\footnotesize{}$\mathsf{PATHWISE}$} &
{\footnotesize{}\% improve}\tabularnewline
\midrule 
{\footnotesize{}$\mathsf{Exp}$ } & {\footnotesize{}$17.9\pm0.3$ } & {\footnotesize{}$17.8\pm0.3$ } & {\footnotesize{}$19.0\pm0.3$ } & {\footnotesize{}$18.2\pm2.7$} & {\footnotesize{}$\mathbf{15.4\pm0.1}$ } & {\footnotesize{}$\mathbf{15.4\pm0.2}$ } & {\footnotesize{}$\mathbf{15.2\pm0.4}$} &
{\footnotesize{}$17.8\%$}
\tabularnewline
\midrule 
{\footnotesize{}$\mathsf{HyperExp}$} & {\footnotesize{}$28.4\pm0.5$ } & {\footnotesize{}$28.3\pm0.8$ } & {\footnotesize{}$28.1\pm0.8$ } &
{\footnotesize{}$27.1\pm4.7$ }& {\footnotesize{}N/A} & 
{\footnotesize{}$24.0\pm0.7$ } & 
{\footnotesize{}$\mathbf{22.5\pm0.7}$ } &
{\footnotesize{}$20.4\%$}
\tabularnewline
\bottomrule
\end{tabular}
\end{table*}

We now describe the standard queuing policies considered in this section, which can all be expressed in the form
\[
    \pi(x)=\argmax_{u\in \mathcal{U}} \sum_{i\in[m],j\in[n]} \rho_{ij}(x)u_{ij}
    \]
for some index $\rho\in \R^{m \times n}$ that differs per method.
\begin{itemize}[itemsep=0pt]
    \item $c\mu$-rule~\cite{cox1961queues}: $\rho_{ij} = h_{j}\mu_{ij}1\{x_{j} > 0 \}$. Servers prioritize queues with a higher holding cost and a larger service rate.
    %Servers are allocated to the queue with the highest $c\mu$ index
    \item MaxWeight~\cite{tassiulas1990stability, stolyar2004maxweight}: $\rho_{ij}(x) = h_{j}\mu_{ij}x_{j}$. Servers prioritize queues that are longer, with a higher holding cost, and a larger service rate.
    %Servers are allocated to the queue with the longest queue, weighted by the cost and service rate.
    \item MaxPressure~\cite{tassiulas1990stability,dai2005maximum}: $\rho_{ij}= \sum_{\ell=1}^{n}\mu_{ij}R_{j\ell}h_{\ell}x_{\ell}$. MaxPressure is a modification of MaxWeight, in the sense that it takes workload externality within the network into account through the $R_{jl}$ terms, e.g., processing a class $j$ job may generate a new class $j'$ job. 
    %Servers are allocated to the queue with longest queue, weighted by the cost and service rate.
    \item Fluid~\cite{bauerle2000asymptotic}: The scheduling policy is based on the optimal sequence of actions in the fluid relaxation, which approximates the average evolution of stochastic queue lengths by deterministic ordinary differential equations. We aim to solve the continuous-time problem
    % \begin{align*}
    % \min_{\bar{u}}\text{ } &\sum_{s=1}^{H} (h^{\top}x_{s}) \Delta t \\
    % \text{s.t. } & x_{s+1} = x_{s} + (\lambda - R(\mu \odot \bar{u}_{s}))\Delta t, \quad \forall s\in [H] \\
    % & x_{s} \geq 0, \quad \forall s\in[H] \\
    % & \bar{u}_{s} \in \mathcal{U}, \quad \forall s\in[H]
    % \end{align*}
    \begin{align*}
    \min_{\bar{u}}\text{ } & \int_{0}^{T} h^\top x(t)dt \\
    \text{s.t. } & \dot{x}(t) = \lambda - R(\mu \odot \bar{u}(t)), \quad  \forall t\in [0,T]\\
    & x(t) \geq 0, \quad \forall t\in [0,T] \\
    & \bar{u}(t) \in \overline{\mathcal{U}}, \quad \forall t\in [0,T].
    \end{align*}
    For tractability, we discretize the problem with time increment $\Delta t > 0$ and horizon $H=T/\Delta t$, and solve as a linear program.
    We then set $u_{k} = \overline{u}(t_{k})$. The linear program is re-solved periodically to improve fidelity with the original stochastic dynamics.
\end{itemize}
We next describe the deep reinforcement learning methods considered in this section:
\begin{itemize}[itemsep=0pt]
    \item $\ppo$-$\mathsf{DG}$~\cite{dai2022queueing}: $\ppo$ is a standard model-free policy gradient method~\cite{schulman2017proximal, huang2022implementation}. ~\citet{dai2022queueing} implement $\ppo$ for multi-class queuing networks and show that the policies obtained outperform several standard queuing policies. 
    Their implementation includes an initial behavioral cloning for stability and a carefully designed variance-reducing policy gradient estimation. We report the results from their paper, although in our experiments we include several problem instances not evaluated in their work.
    \item $\ppo$-$\mathsf{WC}$ (ours): Given the empirical success of $\ppo$ with the work-conserving $\softmax$, we use this algorithm as the main RL benchmark. For this policy, we use the same neural network architecture, hyper-parameters, and variance reduction methods as $\ppo$-$\mathsf{DG}$ ~\cite{dai2022queueing}.
    \item $\pathwise$ policy gradient (Algorithm~\ref{alg:pathwise-pg}): Trains a neural network policy with work-conserving $\softmax$ using the $\pathwise$ policy  gradient estimator. We use an inverse temperature of $\beta = 10$ for all experiments in this section.
\end{itemize}
While value-based methods have also been considered for queuing network control~\cite{liu2022rl,wei2024sample}, our focus in this work is on benchmarking policy gradient algorithms. 

For the reinforcement learning policies, we train each method over $100$ episodes, each consisting of $N = 50,000$ environment steps. Following~\citet{dai2022queueing}'s implementation, $\ppo$-$\mathsf{WC}$ was trained with $B = 50$ actors, while $\pathwise$ was trained only with $B = 1$ actor, which means that $\ppo$-based methods used 50x more trajectories than $\pathwise$. See Appendix~\ref{sec:training-details} for more details on the training process.

To evaluate each scheduling policy, we run $100$ parallel episodes starting from empty queues $x_{0} = \mathbf{0}_{n}$ with a long horizon $N$ to estimate the long-run average holding cost (typically $N =200,000$ steps). As in previous works (e.g.~\cite{dai2022queueing, bertsimas2014robust}), we consider holding costs $h = \ones_{n}$ in which case the holding cost is equivalent to the total queue length. To reiterate, for each policy $\pi$ we estimate the following quantity 
\begin{equation}
    J_{N}(\pi) = 
  \E \left[\frac{1}{t_{N}}\sum_{k=0}^{N-1} (\ones_{n}^{\top}x_{k})\tau^{*}_{k+1} \right]
=  \E \left[\frac{1}{t_{N}} \int^{t_{N}}_{0} \sum_{i=1}^{n} x_{i}(t)dt \right]
\end{equation}
where $t_{N}$ is the time of the $N$th event. We measure the standard deviation across the 100 episodes to form 95\% confidence intervals. For the reinforcement learning policies, we report the average holding cost for the best policy encountered during training.

\vspace{1em}
\begin{table*}[h!]
\centering
 \caption{Re-entrant-1 ($\mathsf{Exp}$)}%
\begin{tabular}{ccccccccc}
\toprule 
{\footnotesize{}Classes} & {\footnotesize{}$c\mu$ } & {\footnotesize{}MaxWeight } & {\footnotesize{}MaxPressure} & {\footnotesize{}Fluid } & {\footnotesize{}$\mathsf{PPO}$-$\mathsf{DG}$~\cite{dai2022queueing} } & {\footnotesize{}$\ppowc$} & {\footnotesize{}$\mathsf{PATHWISE}$} & {\footnotesize{}\% improve}\tabularnewline
\midrule 
{\footnotesize{}6} & {\footnotesize{}$17.4\pm0.4$ } & {\footnotesize{}$17.5\pm0.4$ } & {\footnotesize{}$18.8\pm0.5$ } & {\footnotesize{}$16.8\pm4.3$} & {\footnotesize{}$14.1\pm0.2$ } & {\footnotesize{}$\mathbf{13.6\pm0.4}$ } & {\footnotesize{}$14.9\pm0.5$ } & {\footnotesize{}14.3\%}\tabularnewline
\midrule 
{\footnotesize{}9} & {\footnotesize{}$23.3\pm0.6$ } & {\footnotesize{}$26.1\pm0.5$ } & {\footnotesize{}$24.2\pm0.6$ } & {\footnotesize{}$27.7\pm4.4$} & {\footnotesize{}$23.3\pm0.3$ } & {\footnotesize{}${\bf 22.6\pm0.4}$ } & {\footnotesize{}${\bf 22.0\pm0.6}$ } & {\footnotesize{}4.3\%}\tabularnewline
\midrule 
{\footnotesize{}12 } & {\footnotesize{}$33.0\pm0.8$ } & {\footnotesize{}$34.0\pm1.0$ } & {\footnotesize{}$35.1\pm0.9$ } & {\footnotesize{}$40.6\pm4.8$} & {\footnotesize{}$32.2\pm0.6$ } & {\footnotesize{}$\mathbf{29.7\pm0.4}$ } & {\footnotesize{}$\mathbf{30.7\pm0.7}$ } & {\footnotesize{}7.0\%}\tabularnewline
\midrule 
{\footnotesize{}15 } & {\footnotesize{}$40.2\pm1.3$ } & {\footnotesize{}$43.6\pm1.1$ } & {\footnotesize{}$42.2\pm1.3$ } & {\footnotesize{}$49.8\pm5.1$} & {\footnotesize{}$39.3\pm0.6$ } & {\footnotesize{}$38.7\pm0.4$ } & {\footnotesize{}$\mathbf{36.2\pm0.8}$ } & {\footnotesize{}9.9\%}\tabularnewline
\midrule 
{\footnotesize{}18 } & {\footnotesize{}$ 48.5\pm1.0$ } & {\footnotesize{}$51.0\pm1.4$ } & {\footnotesize{}$52.4\pm1.6$ } & {\footnotesize{}$54.5\pm4.2$} & {\footnotesize{}$51.4\pm1.0$ } & {\footnotesize{}$47.5\pm0.5$ } & {\footnotesize{}$\mathbf{45.7\pm0.7}$ } & {\footnotesize{}$5.7\%$}\tabularnewline
\midrule 
{\footnotesize{}21 } & {\footnotesize{}$55.2\pm1.1$ } & {\footnotesize{}$59.5\pm1.6$ } & {\footnotesize{}$56.0\pm1.6$} & {\footnotesize{}$63.7\pm6.7$} & {\footnotesize{}$55.1\pm1.8$} & {\footnotesize{}$56.3\pm0.8$ } & {\footnotesize{}$\mathbf{52.8\pm1.2}$ } & {\footnotesize{}4.3\%}\tabularnewline
\midrule 
{\footnotesize{}24 } & {\footnotesize{}$64.9\pm1.4$ } & {\footnotesize{}$69.4\pm1.6$ } & {\footnotesize{}$66.6\pm2.2$ } & {\footnotesize{}$74.0\pm6.7$} & {\footnotesize{}N/A} & {\footnotesize{}$65.8\pm0.6$ } & {\footnotesize{}$\mathbf{60.2\pm0.9}$ } & {\footnotesize{}6.2\%}\tabularnewline
\midrule 
{\footnotesize{}27 } & {\footnotesize{}$71.1\pm1.5$ } & {\footnotesize{}$77.7\pm1.9$ } & {\footnotesize{}$72.0\pm2.5$ } & {\footnotesize{}$83.1\pm7.1$} & {\footnotesize{}N/A} & {\footnotesize{}$75.8\pm0.7$ } & {\footnotesize{}$\mathbf{67.7\pm1.3}$ } & {\footnotesize{}3.0\%}\tabularnewline
\midrule 
{\footnotesize{}30 } & {\footnotesize{}$87.7\pm2.5$ } & {\footnotesize{}$84.5\pm2.1$ } & {\footnotesize{}$106.8\pm2.5$ } & {\footnotesize{}$93.6\pm8.0$} & {\footnotesize{}N/A} & {\footnotesize{}$83.1\pm0.7$ } & {\footnotesize{}$\mathbf{77.8\pm1.3}$ } & {\footnotesize{}7.9\%}\tabularnewline
\bottomrule
\end{tabular}
\vspace{1em}
 \caption{Re-entrant-1 ($\mathsf{HyperExp}$)}
\begin{tabular}{cccccccc}
\toprule 
{\footnotesize{}Classes} & {\footnotesize{}$c\mu$ } & {\footnotesize{}MaxWeight } & {\footnotesize{}MaxPressure} & {\footnotesize{}Fluid } & {\footnotesize{}$\mathsf{PPO}$-$\mathsf{WC}$} & {\footnotesize{}$\mathsf{PATHWISE}$} & {\footnotesize{}\% improve}\tabularnewline
\midrule 
{\footnotesize{}6} & {\footnotesize{}$37.8\pm1.3$} & {\footnotesize{}$39.2\pm1.2$} & {\footnotesize{}$43.8\pm1.8$} & {\footnotesize{}$43.6\pm7.5$} & {\footnotesize{}$\mathbf{29.9\pm0.7}$} & {\footnotesize{}$32.2\pm1.4$ } & {\footnotesize{}6.9\%}\tabularnewline
\midrule 
{\footnotesize{}9} & {\footnotesize{}$50.2\pm1.9$} & {\footnotesize{}$55.5\pm2.2$} & {\footnotesize{}$68.7\pm2.7$} & {\footnotesize{}$59.2\pm8.2$} & {\footnotesize{}$\mathbf{47.5\pm0.8}$} & {\footnotesize{}$\mathbf{45.6\pm1.6}$ } & {\footnotesize{}9.2\%}\tabularnewline
\midrule 
{\footnotesize{}12 } & {\footnotesize{}$70.0\pm2.5$} & {\footnotesize{}$72.3\pm2.8$} & {\footnotesize{}$89.4\pm3.6$} & {\footnotesize{}$75.6\pm15.3$} & {\footnotesize{}$64.4\pm1.2$} & {\footnotesize{}$\mathbf{61.1\pm2.5}$ } & {\footnotesize{}12.7\%}\tabularnewline
\midrule 
{\footnotesize{}15 } & {\footnotesize{}$81.7\pm4.0$} & {\footnotesize{}$91.0\pm3.7$} & {\footnotesize{}$112.0\pm4.9$} & {\footnotesize{}$97.0\pm12.9$} & {\footnotesize{}$81.8\pm1.1$} & {\footnotesize{}${\bf 76.9\pm2.5}$ } & {\footnotesize{}5.9\%}\tabularnewline
\midrule 
{\footnotesize{}18 } & {\footnotesize{}$101.1\pm4.7$} & {\footnotesize{}$103.9\pm3.9$} & {\footnotesize{}$126.7\pm6.2$} & {\footnotesize{}$111.2\pm14.4$} & {\footnotesize{}$99.8\pm1.5$} & {\footnotesize{}${\bf 91.5\pm2.5}$ } & {\footnotesize{}9.4\%}\tabularnewline
\midrule 
{\footnotesize{}21 } & {\footnotesize{}$116.3\pm4.6$} & {\footnotesize{}$123.3\pm3.9$} & {\footnotesize{}$152.3\pm6.6$} & {\footnotesize{}$151.0\pm21.3$} & {\footnotesize{}$118.2\pm2.0$} & {\footnotesize{}${\bf 108.0\pm3.2}$ } & {\footnotesize{}7.2\%}\tabularnewline
\bottomrule
\end{tabular}
\end{table*}

Tables 1-5 display the results of our benchmarking across the problem instances discussed before. The column `\% improve' records the relative reduction in holding cost achieved by $\pathwise$ over the best standard queuing policy (either $c\mu$, MaxWeight, MaxPressure, or Fluid). Our main observations on the relative performance of the standard policies and policies obtained from reinforcement learning methods are summarized as follows.
\begin{itemize}[itemsep=0pt]
    \item {\bf $\ppowc$ is a strong reinforcement-learning benchmark.} $\ppo$ with our proposed work-conserving $\softmax$ is able to efficiently find policies that outperform standard policies across all problem instances, as well as the $\ppo$ featured in~\cite{dai2022queueing} (under the same policy network and hyper-parameters). This illustrates that simply ensuring work-conservation is a powerful inductive bias that delivers stability.
    \item {\bf $\pathwise$ policy gradient outperforms $\ppowc$ in larger networks, using 50x less data}. We observe that $\ppowc$ and $\pathwise$ achieve similar performances when the number of classes is small. However, when the number of job classes gets larger, $\pathwise$ consistently outperforms $\ppowc$, as seen in Figure~\ref{fig:policy_iterations}. This is likely due to the 
    %For smaller networks, the performance is equivalent. This is a practical demonstration of the improvement in 
    sample efficiency gained by the $\pathwise$ gradient, enabling the algorithm to find better policies with less data.
    \item {\bf $\pathwise$ achieves large performance gains over  $\ppowc$ for higher-variance problem instances, using 50x less data}. We observe that for the Re-entrant-1 networks with hyper-exponential noise, the reduction of holding cost of $\pathwise$ relative to $\ppowc$ is often equivalent or even larger than the cost reduction of $\ppowc$ relative to the $c\mu$-rule. This illustrates that even among optimized RL policies, there can be significant performance differences for difficult problem instances, and the sample efficiency of $\pathwise$ is particularly useful in noisier environments.
    \item {\bf While standard queuing methods work well, RL methods meaningfully improve performance in hard instances.} We observe in the `\% improve' column that $\pathwise$ achieves a 3-20\% improvement over the best standard queuing policy in each setting.
\end{itemize}

Altogether, these results illustrate that policy gradient with $\pathwise$ gradient estimator and work-conserving $\softmax$ policy architecture can learn effective queuing network control policies with substantially less data than model-free policy gradient algorithms for large networks with high-variance event times, mirroring real-world systems. 
In particular, the improved sample efficiency of the $\pathwise$ gradient estimator and the stability brought by work-conserving $\softmax$ policy architecture are the keys to enabling learning in large-scale systems with realistic data requirements.

\begin{table}[h!]
\centering
 \caption{Re-entrant-2 ($\mathsf{Exp}$)}%
\begin{tabular}{cccccccc}
\toprule 
{\footnotesize{}Classes} & {\footnotesize{}$c\mu$ } & {\footnotesize{}MaxWeight } & {\footnotesize{}MaxPressure} & {\footnotesize{}Fluid } & {\footnotesize{}$\mathsf{PPO}$-$\mathsf{WC}$ } & {\footnotesize{}$\mathsf{PATHWISE}$} & {\footnotesize{}\% improve}\tabularnewline
\midrule 
{\footnotesize{}6} & {\footnotesize{}$18.8\pm0.5$} & {\footnotesize{}$17.4\pm0.4$} & {\footnotesize{}$24.5\pm0.7$} & {\footnotesize{}$18.6\pm2.8$} & {\footnotesize{}$\mathbf{13.7\pm0.2}$} & {\footnotesize{}$14.7\pm0.6$} & {\footnotesize{}21.9\%}\tabularnewline
\midrule 
{\footnotesize{}9} & {\footnotesize{}$24.2\pm0.6$} & {\footnotesize{}$25.8\pm0.7$} & {\footnotesize{}$31.5\pm1.0$} & {\footnotesize{}$26.7\pm3.7$} & {\footnotesize{}$\mathbf{22.1\pm0.3}$} & {\footnotesize{}$\mathbf{21.6\pm0.6}$} & {\footnotesize{}10.7\%}\tabularnewline
\midrule 
{\footnotesize{}12 } & {\footnotesize{}$35.1\pm0.9$} & {\footnotesize{}$34.0\pm1.1$} & {\footnotesize{}$40.3\pm1.4$} & {\footnotesize{}$35.7\pm4.6$} & {\footnotesize{}$\mathbf{29.9\pm0.5}$} & {\footnotesize{}$\mathbf{29.8\pm0.7}$} & {\footnotesize{}15.1\%}\tabularnewline
\midrule 
{\footnotesize{}15 } & {\footnotesize{}$44.8\pm1.3$} & {\footnotesize{}$43.8\pm1.5$} & {\footnotesize{}$50.1\pm1.6$} & {\footnotesize{}$44.0\pm5.3$} & {\footnotesize{}$\mathbf{38.0\pm0.5}$} & {\footnotesize{}$\mathbf{36.2\pm1.0}$} & {\footnotesize{}19.2\%}\tabularnewline
\midrule 
{\footnotesize{}18 } & {\footnotesize{}$52.4\pm1.6$} & {\footnotesize{}$48.7\pm1.3$} & {\footnotesize{}$58.2\pm2.1$} & {\footnotesize{}$55.2\pm5.8$} & {\footnotesize{}$\mathbf{46.8\pm0.5}$} & {\footnotesize{}$\mathbf{45.6\pm0.8}$} & {\footnotesize{}13.0\%}\tabularnewline
\midrule 
{\footnotesize{}21 } & {\footnotesize{}$56.0\pm1.6$} & {\footnotesize{}$57.5\pm1.8$} & {\footnotesize{}$71.3\pm2.9$} & {\footnotesize{}$62.2\pm7.5$} & {\footnotesize{}$55.5\pm0.6$} & {\footnotesize{}$\mathbf{51.4\pm1.1}$} & {\footnotesize{}8.2\%}\tabularnewline
\midrule 
{\footnotesize{}24 } & {\footnotesize{}$66.6\pm2.2$} & {\footnotesize{}$69.0\pm1.7$} & {\footnotesize{}$76.0\pm3.0$} & {\footnotesize{}$70.8\pm7.7$} & {\footnotesize{}$63.2\pm0.7$} & {\footnotesize{}$\mathbf{59.8\pm1.4}$} & {\footnotesize{}10.2\%}\tabularnewline
\midrule 
{\footnotesize{}27 } & {\footnotesize{}$72.0\pm2.5$} & {\footnotesize{}$75.9\pm2.1$} & {\footnotesize{}$84.9\pm3.2$} & {\footnotesize{}$82.9\pm9.6$} & {\footnotesize{}${\bf 70.3\pm0.9}$} & {\footnotesize{}$\mathbf{68.4\pm1.6}$} & {\footnotesize{}5.0\%}\tabularnewline
\midrule 
{\footnotesize{}30 } & {\footnotesize{}$80.6\pm2.7$} & {\footnotesize{}$83.8\pm1.9$} & {\footnotesize{}$90.6\pm3.2$} & {\footnotesize{}$91.2\pm11.3$} & {\footnotesize{}$80.4\pm0.8$} & {\footnotesize{}$\mathbf{75.5\pm1.8}$} & {\footnotesize{}6.3\%}\tabularnewline
\bottomrule
\end{tabular}

 \caption{Re-entrant-2 ($\mathsf{HyperExp}$)}

\begin{tabular}{cccccccc}
\toprule 
{\footnotesize{}Classes} & {\footnotesize{}$c\mu$ } & {\footnotesize{}MaxWeight } & {\footnotesize{}MaxPressure} & {\footnotesize{}Fluid } & {\footnotesize{}$\ppowc$} & {\footnotesize{}$\mathsf{PATHWISE}$} & {\footnotesize{}\% improve}\tabularnewline
\midrule 
{\footnotesize{}6} & {\footnotesize{}$39.4\pm1.4$} & {\footnotesize{}$39.1\pm1.9$} & {\footnotesize{}$58.6\pm2.3$} & {\footnotesize{}$39.8\pm7.3$} & {\footnotesize{}$\mathbf{30.7\pm0.8}$} & {\footnotesize{}$\mathbf{30.9\pm1.4}$} & {\footnotesize{}21.6\%}\tabularnewline
\midrule 
{\footnotesize{}9} & {\footnotesize{}$52.4\pm2.1$} & {\footnotesize{}$58\pm2.7$} & {\footnotesize{}$67.1\pm3.0$} & {\footnotesize{}$55.5\pm9.4$} & {\footnotesize{}$\mathbf{45.7\pm0.8}$} & {\footnotesize{}$\mathbf{43.4\pm1.4}$} & {\footnotesize{}17.2\%}\tabularnewline
\midrule 
{\footnotesize{}12 } & {\footnotesize{}$70.9\pm2.7$} & {\footnotesize{}$77.0\pm3.7$} & {\footnotesize{}$93.0\pm4.0$} & {\footnotesize{}$72.1\pm14.4$} & {\footnotesize{}$\mathbf{61.1\pm1.3}$} & {\footnotesize{}$\mathbf{58.9\pm2.3}$} & {\footnotesize{}16.9\%}\tabularnewline
\midrule 
{\footnotesize{}15 } & {\footnotesize{}$81.5\pm2.7$} & {\footnotesize{}$90.5\pm3.9$} & {\footnotesize{}$109.0\pm5.4$} & {\footnotesize{}$84.2\pm18.1$} & {\footnotesize{}$78.0\pm1.7$} & {\footnotesize{}${\bf 73.8\pm3.6}$} & {\footnotesize{}9.4\%}\tabularnewline
\midrule 
{\footnotesize{}18 } & {\footnotesize{}$104.3\pm5.1$} & {\footnotesize{}$103.8\pm4.3$} & {\footnotesize{}$123.6\pm5.3$} & {\footnotesize{}$99.5\pm15.5$} & {\footnotesize{}$\mathbf{93.8\pm1.3}$} & {\footnotesize{}$\mathbf{92.1\pm2.7}$} & {\footnotesize{}11.7\%}\tabularnewline
\midrule 
{\footnotesize{}21 } & {\footnotesize{}$116.6\pm6.0$} & {\footnotesize{}$117.9\pm5.1$} & {\footnotesize{}$135.6\pm6.0$} & {\footnotesize{}$118.9\pm16.6$} & {\footnotesize{}$110.5\pm1.6$} & {\footnotesize{}$\mathbf{104.2\pm2.9}$} & {\footnotesize{}10.6\%}\tabularnewline
\bottomrule
\end{tabular}
\end{table}

\section{Why is $\reinforce$ Sample-Inefficient? A Theoretical Case Study for the $M/M/1$ Queue}
\label{section:case_study}

In this section, we provide a theoretical case study explaining how $\pathwise$
gradients are able to learn more from a single observed trajectory compared to
$\reinforce$. We focus on the special case of the $M/M/1$ queue to explain how
$\reinforce$ and its actor-critic variants utilizing baselines/advantages suffer from
sample inefficiency.
Although we are only able to analyze a substantially simpler setting than the control problems in general multi-class queueing networks we are interested in, 
our theoretical results illustrate the essential statistical benefits of pathwise gradient estimators.
We highlight that while $\reinforce$ applies to virtually any setting by relying on random exploration, %unlike $\pathwise$ 
it fundamentally struggles to assign credit to actions, especially in noisy environments. $\pathwise$, on the other hand, is much better at assigning credit to actions. This allows us to crystallize why we see such a large improvement in sample efficiency in sections~\ref{section:gradient_eval} and~\ref{section:results}. We also believe our result may be of broader interest in reinforcement learning, because it illustrates a practically relevant instance where $\reinforce$, even with an optimal baseline, is provably sub-optimal. 

We consider the $M/M/1$ queue with a fixed arrival rate $\lambda$ under service rate control $u=\mu > \lambda$. This setting permits an analytic approach to showing that the $\reinforce$ estimator has a sub-optimally large variance, particularly for congested systems when $\rho = \lambda/\mu \to 1$. On the other hand, the $\pathwise$ estimator achieves an order of magnitude improvement in estimation efficiency. 
In the $M/M/1$ queue setting, the $\pathwise$ approach
for general queuing networks 
reduces to IPA based on Lindley recursion. 
% \hntodo{You use statistical benefits and sample efficiency throughout, which gives the impression that our goal is to utilize data to its full potential. This is a bit orthogonal to how most applied probabilists think about simulation optimization. I recommend either clarifying what you mean early on in the paper or better yet, keeping to computational efficiency as the main framing.}

We consider a simple service-rate control problem where the cost is the steady-state average queue length in the $M/M/1$ queue
\[
Q(\mu) := \E_{\infty}[x(t)] = \frac{\lambda}{\mu - \lambda} = \frac{\rho}{1-\rho}.
\]
Given that the service rate is continuous, it is natural to consider a policy that randomizes over $[\mu - h, \mu]$. One such option is a Beta-distributed policy $\pi_{\theta}: A = \mu - hY$, where $Y \sim \text{Beta}(\theta,1)$ and $h>0$. As $\theta \to \infty$, the policy frequently sets service rates close to $\mu$ and as $\theta \to 0$, it concentrates more probability mass on service rates close to $\mu - h$. The task is to estimate the following policy gradient
\[
\nabla_{\theta} J(\theta) = \nabla \E_{A\sim \pi_{\theta}}[Q(A)].
\]
The $\mathsf{REINFORCE}$ estimator of $\nabla_{\theta}J(\theta)$ involves sampling a random service rate from the policy $\pi_{\theta}$, and then estimating the steady-state queue length from a trajectory under that service rate. For a trajectory with $N$ steps, we denote the steady-state queue length estimator as $\widehat{Q}_{N}(A)$. Then, the $\reinforce$ gradient estimator takes the form
\begin{equation}
\label{eq:mm1-reinforce}
\widehat{\nabla}^{\mathsf{R}} J_{N}(\theta; \xi_{1:N})
= \widehat{Q}_{N}(\mu - hY) \nabla_{\theta} \log \pi_{\theta}(Y) 
= \widehat{Q}_{N}(\mu - hY) \left( \log Y + \frac{1}{\theta} \right).
\end{equation}
The standard estimator for the steady-state queue length is simply the queue length averaged over a sample path: 
$\widehat{Q}_{N}(a) = \frac{1}{N}\sum_{k=1}^{N}x_{k}\tau_{k+1}^*$ when $A=a$. As long as $\mu - h > \lambda$, it is known that as $N\to \infty$, $\widehat{Q}_{N}(a) \to Q(a)$, which implies that $\widehat{\nabla}^{\mathsf{R}} J_{N}(\theta; \xi_{1:N}) \to \nabla J(\theta)$.

% \hntodo{$U$ is usually reserved for a uniform variable. Since we don't follow typical notational conventions (e.g., capital letters are RVs), it occasionally gets confusing. Concrete here, $\omega$ is now a uniform in the reparameterization trick, introducing some notational overhead.}
On the other hand, the $\pathwise$ estimator utilizes the structure of the single server queue. First, by inverse transform sampling, $Y \stackrel{d}{=} F_{\theta}^{-1}(\omega)$, where $\omega \sim \text{Uniform(0,1)}$ and $F_{\theta}^{-1}(\omega) = \omega^{1/\theta}$. Then, we can substitute $A = \mu - h\omega^{1/\theta}$. Since $Q(\mu)$ is differentiable and the derivative is integrable, we can change the order of differentiation and integration
\[
\nabla_{\theta} J(\theta) 
= \nabla_{\theta} \E_{\omega}[Q(\mu - h\omega^{1/\theta})]
= -\E_{\omega}[\nabla Q(\mu - h\omega^{1/\theta}) \cdot h \nabla_{\theta} \omega^{1/\theta}].
\]
The preceding display involves the gradient of the steady-state queue-length $Q(\mu)$ with respect to the service rate $\mu$, i.e., $\nabla Q(\mu)$. 

For the $M/M/1$ queue, there are consistent sample-path estimators of $\nabla Q(\mu)$. One such estimator uses the fact that by Little's law $Q(\mu) = \E_{\infty}[x(t)] = \lambda \E_{\infty}[w(t)] =: \lambda W(\mu)$ where $W(\mu)$ is the steady-state waiting time. The waiting time process $W_{i}$, which denotes the waiting time of the $i$th job arriving to the system, has the following dynamics, known as the Lindley recursion:
\begin{equation}
\label{eq:lindley}
 W_{i+1} = \left( W_{i} - T_{i+1} + \frac{S_{i}}{\mu} \right)^{+},
\end{equation}
where $T_{i+1}\stackrel{\text{iid}}{\sim}\text{Exp(1)}$ is the inter-arrival time of between the $i$th job and $(i+1)$th job, and $S_{i}\stackrel{\text{iid}}{\sim}\text{Exp(1)}$ is the workload of the $i$th job. Crucially, this stochastic recursion specifies how the service rate affects the waiting time along the sample path, which enables one to derive a pathwise derivative via the recursion:
\[
 \widehat{\nabla}W_{i+1} = 
 \left( -\frac{S_{k}}{\mu^{2}} + \widehat{\nabla} W_{i} \right)
 \mathbf{1} \left\{ W_{i+1} > 0 \right\},
\]
where $\widehat{\nabla} W_{i}$ together with $W_i$ form a Markov chain following the above recursion. By averaging this gradient across jobs and using Little's law, we have the following gradient estimator for $\nabla Q$, which we denote as $\widehat{\nabla} Q_{N}(\mu)$: 
%In particular, let $L_{N}$ be the number of arrivals that occur during a sample path with $N$ events. Then,
\[
\widehat{\nabla} Q_{N}(\mu) = \lambda \frac{1}{L_{N}} \sum_{i=1}^{L_{N}} \widehat{\nabla} W_{i},
\]
where $L_{N}$ is the number of arrivals that occur during a sample path with $N$ events.
Using this, the $\pathwise$ policy gradient estimator (a.k.a. IPA estimator) is
\begin{equation}
\label{eq:mm1-pathwise}
\widehat{\nabla}_{\theta} J_{N}(\theta; \xi_{1:N}) 
= h \cdot \widehat{\nabla} Q_{N}(\mu - h Y) \left( \frac{1}{\theta} Y \log Y \right).
\end{equation}
As long as $\mu - h > \lambda$, it has been established that $\widehat{\nabla} Q_{N}(\mu)$ is asymptotically unbiased, i.e., as $N\to \infty$, $\E[\widehat{\nabla} Q_{N}(\mu)] \to \nabla Q(\mu)$, which implies 
$\E[\widehat{\nabla}_{\theta} J_{N}(\theta; \xi_{1:N})] \to \nabla J(\theta)$. 

Since both the $\reinforce$ and $\pathwise$ gradient estimators give an asymptotically unbiased estimation of $\nabla J(\theta)$, we compare them based on their variances, which determines how many samples are needed to reliably estimate the gradient. Although the variances of the estimators are not precisely known for a finite $N$, $\widehat{Q}_{N}$ and $\widehat{\nabla}Q_{N}$ both satisfy the central limit theorem (CLT) with explicitly characterized asymptotic variances, which we denote as $\var_{\infty}(\widehat{Q})$ and $\var_{\infty}(\widehat{\nabla}Q)$ 
% \hntodo{Subscripts on $\sigma$ are cumbersome. How about $\var(\what{Q})$ instead?} 
respectively. This implies that the variance of $\widehat{Q}_{N}$ is approximately $\var_{\infty}(\widehat{Q})/N$. We define $\var_{\infty}(\widehat{\nabla} J_{N}(\theta; \xi_{1:N}))$ to be the variance of the gradient estimator when we approximate the variance of $\widehat{Q}_{N}$ and $\widehat{\nabla}Q_{N}$ using $\var_{\infty}(\widehat{Q})/N$ and $\var_{\infty}(\widehat{\nabla}Q)/N$ respectively.
%we plug in the asymptotic variance of the corresponding queueing estimator.

It is worth reiterating that the $\reinforce$ estimator only required an estimate of cost $Q(\mu)$, which does not require any domain knowledge, whereas the $\pathwise$ gradient required an estimate of $\nabla Q(\mu)$ which requires a detailed understanding of how the service rate affected the sample path dynamics. Utilizing this structural information can greatly improve the efficiency of gradient estimation. Since $\widehat{\nabla}_{\theta} J_{N}(\theta; \xi_{1:N}) = O(h)$ \emph{almost surely},
we have $\var(\widehat{\nabla}_{\theta}J_{N}(\theta; \xi_{1:N})) = O(h^{2})$. On the other hand, the variance of the $\reinforce$ estimator can be very large even if $h$ is small. To highlight this, consider the extreme case where $h = 0$ for which the policy gradient $\nabla J(\theta)$ is obviously zero since the policy deterministically sets the service rate to $\mu$ regardless of $\theta$. Strikingly, $\reinforce$ does not have zero variance in this case:

\begin{observation}
\label{obs:h_0}
If $h = 0$, then the variance of the estimators are
\[
\var_{\infty}\left(\widehat{\nabla} J_{N}(\theta; \xi_{1:N}) \right) = 0, \quad
\var_{\infty}\left(\widehat{\nabla}^{\mathsf{R}} J_{N}(\theta; \xi_{1:N}) \right)
= \Theta \left( N^{-1} (1 - \rho)^{-4} \right)
\]
\end{observation}
\noindent Note that even when $h=0$, the variance of the $\reinforce$ estimator can be quite high if the queue is congested, i.e. $\rho$ is close to $1$, while the pathwise estimator gives the correct estimate of zero with zero variance.

For non-trivial values of $h$, we focus on the so-called `heavy-traffic' asymptotic regime with $\rho=\lambda/\mu \to 1$, which is of major theoretical and practical interest in the study of queues. Estimating steady-state quantities becomes harder as the queue is more congested, so $(1 - \rho)^{-1}$ emerges as a key scaling term in the variance. 
%In addition, if $\theta \to 0$, then both estimators will diverge to infinity. 
We set $h$ such that $\frac{h}{\mu - \lambda}<c$ and $\frac{h}{\mu - \lambda}\to c \in (0,1)$ as $\rho \to 1$. This resembles the square-root heavy-traffic regime for capacity planning, where the service rate is set to be $\lambda + \beta \sqrt{\lambda}$ for some $\beta > 0$, and one considers the limit as $\lambda \to \infty$. In this case, if one were choosing a policy over the square-root capacity rules $A\in[\lambda + a \sqrt{\lambda}, \lambda + b \sqrt{\lambda}]$ for some $b>a>0$, this is equivalent to setting $\mu = \lambda + b\sqrt{\lambda}$ and $h = (b - a) \sqrt{\lambda} = O(\sqrt{\lambda})$. Note that if $c = 0$, the gradient is zero (identical to Observation~\ref{obs:h_0}), and if $c \geq 1$, the queue with service rate $\mu-h$ is unstable.

Within this regime, we have the following comparison between the gradient estimators, which utilizes recent results concerning the asymptotic variance of $\widehat{\nabla}Q_N(\mu)$~\cite{hu2023comparison}.
\begin{theorem}
\label{thm:mm1_variance}
Suppose $h = c(\mu - \lambda)$ for $c\in(0,1)$ as $\rho \to 1$. Under this scaling
$\nabla J(\theta) \sim (1-\rho)^{-1}$,
and
\begin{align}
\var_{\infty}\left(\widehat{\nabla} J_{N}(\theta; \xi_{1:N}) \right) 
&= O \Big( \underbrace{N^{-1}  \left(1 - \rho \right)^{-3}}_{\text{estimation noise}} + \underbrace{(1-\rho)^{-2}}_{\text{policy randomization}} \Big) \label{eq:V_R}\\ 
\var_{\infty}\left(\widehat{\nabla}^{\mathsf{R}} J_{N}(\theta; \xi_{1:N}) \right) &= \Theta 
\left( N^{-1} \left(1 -\rho\right)^{-4}
+ (1-\rho)^{-2}
\right) \label{eq:V_P}
\end{align}
\end{theorem}
\noindent See Appendix~\ref{sec:proof_mm1_variance} for the proof.

Overall, the $\pathwise$ estimator is much more sample efficient than the $\reinforce$ estimator as $\rho \to 1$, with the variance scaling as $(1-\rho)^{-3}$ compared to $(1-\rho)^{-4}$.
The first terms in \eqref{eq:V_R} and \eqref{eq:V_P} represent the variance occurring from the Monte Carlo estimation and becomes smaller if one generates a longer sample path (larger $N$), and it scales as $N^{-1}$. The second terms are the variance resulting from randomness in the service rate induced by the policy. 
%$\reinforce$ $(1-\rho)^{-1}$ times more samples in order to have the same sample variance as the pathwise estimator. 
% The pathwise estimator also improves upon the dependence on $\theta^{-1}$ in the second term as $\theta \to 0$. This is because the score function $\nabla_{\theta} \log \pi_{\theta}$, which appears in $\reinforce$ estimator, generally becomes poorly conditioned as $\pi_{\theta}$ converges to a deterministic policy.

Theorem \ref{thm:mm1_variance} illustrates that large improvements in statistical efficiency can be achieved by leveraging the structure of the system dynamics. An existing strategy for incorporating domain knowledge in $\reinforce$ is to subtract a baseline $b$ from the cost, which preserves un-biasedness:
\begin{equation}
\label{eq:mm1-reinforce-baseline}
\widehat{\nabla}^{\mathsf{RB}} J_{N}
(\theta, \xi_{1:N})
= (\widehat{Q}_{N}(\mu - hY) - b) \left( \log Y + \frac{1}{\theta} \right).
\end{equation}
%A well-chosen baseline can reduce the variance of the estimator. 
In this case, one can characterize the optimal variance-reducing baseline in closed form if one has knowledge of the true cost $Q(\mu)$. Under $h = c(\mu - \lambda)$,
\begin{align*}
b^{*} &= \frac{\E[Q(h-hY) \nabla_{\theta} \log \pi_{\theta}(Y)^{2}]}{\E[\nabla_{\theta}\log \pi_{\theta}(Y)^{2}]} \\
&= \frac{\lambda}{\mu - \lambda}\left[F^{2}_{1}\left(1,\theta,1+\theta,c\right) -2\theta^{2}\Phi\left(c,2,\theta\right) + 2c\theta^{3}\Phi\left(c,3,1+\theta\right)\right] \\
& = O((1-\rho)^{-1})
\end{align*}
where $F^{2}_{1}$ is the hypergeometric 2F1 function and $\Phi$ is the Lerch $\Phi$ transcendental. The optimal baseline $b^{*}$ is of the same order as $Q(\mu)$ as $\rho \to 1$.
%We compute the variance under the optimal baseline. See Appendix~\ref{sec:proof_reinforce_baseline} for the proof.
\begin{corollary}
\label{cor:reinforce-baseline}
Consider the $\reinforce$ estimator with the optimal baseline $b^{*}$. As $\rho \to 1$, the variance of the estimator scales as
\[
\var_{\infty}\left(\widehat{\nabla}^{\mathsf{RB}} J_{N}(\theta; \xi_{1:N}) \right) = \Theta 
\left( N^{-1} \left(1 -\rho\right)^{-4} 
+(1-\rho)^{-2}
\right)
\]
\end{corollary}

The proof of Corollary \ref{cor:reinforce-baseline} is provided in Appendix~\ref{sec:proof_reinforce_baseline}.
Simply, since the optimal baseline is a deterministic input, it is unable to improve upon the $(1-\rho)^{-4}$ dependence on $\rho$, which is driven by the statistical properties of $\widehat{Q}_{N}$. 
%The optimal baseline also does not fix the issue pointed out in Observation~\ref{obs:h_0} and still has a potentially large variance when $h = 0$.
This illustrates that the pathwise gradient estimator can offer an order of magnitude improvement in sample efficiency than the $\reinforce$ estimator even with an optimized baseline that requires knowledge of the true cost function (and thus precludes the need to estimate the cost in the first place). 

Intuitively, the $\reinforce$ estimator is inefficient because it is unable to leverage the fact that %at the \emph{sample-path-level}, 
$Q(\mu) \approx Q(\mu + \epsilon)$ when $\epsilon$ is small. After all, generic MDPs do not have such a structure; a slight change in the action can result in vastly different outcomes. The $\reinforce$ estimator cannot use the estimate of $\widehat{Q}_{N}(\mu)$ to say anything about $Q(\mu + \epsilon)$, and must draw a new sample path to estimate $Q(\mu + \epsilon)$. Meanwhile, using a \emph{single} sample path, the pathwise estimator can obtain an estimate for $Q(\mu + \epsilon)$ when $\epsilon$ is small enough via $\widehat{Q}_{N}(\mu + \epsilon)\approx \widehat{Q}_{N}(\mu) + \epsilon\widehat{\nabla}Q_{N}(\mu)$. In this sense, the pathwise estimator can be seen as a \emph{infinitesimal counterfactual} of the outcome under alternative---but similar---actions.

Even though we only study a single server queue here, %and we study the IPA estimator instead of our proposed $\pathwise$ estimator,
we believe the key observations may apply more broadly.
\begin{itemize}[itemsep=0pt]
\item Higher congestion ($\rho \to 1$) makes it more challenging to estimate the performance of queueing networks based on the sample path. This applies to both gradient estimators and baselines that could be used to reduce variance.
\item It is important to reliably estimate the effects of small changes in the policy, as large changes can potentially cause instability. Pathwise gradient estimators provide a promising way to achieve this. For general networks with known dynamics, their dynamics are often not differentiable, which requires the development of the $\pathwise$ estimator.
% \item While the queue lengths $x_{k}$ are discrete, the dynamics may look more continuous when viewing the system through the event times. Unfortunately, while general networks follow known dynamics, these dynamics are more complicated than the Lindley recursion~\eqref{eq:lindley}, and are not differentiable, which requires development of the $\pathwise$ estimator.
\end{itemize}

\section{Conclusion}
\label{section:conclusion}

In this work, we introduce a new framework for policy optimization in queuing network control. This framework uses a novel approach for gradient estimation in discrete-event dynamical systems. Our proposed $\pathwise$ policy gradient estimator is observed to be orders of magnitude more efficient than model-free RL alternatives such as $\reinforce$ across an array of carefully designed empirical experiments. In addition, we introduce a new policy architecture, which drastically improves stability while maintaining the flexibility of neural network policies. Altogether, these illustrate how structural knowledge of queuing networks can be leveraged to accelerate reinforcement learning for queuing control problems.

We next discuss some potential extensions of our approach:
\begin{itemize}[itemsep=0pt]
    %parallel server systems~\cite{harrison1998heavy}.
    \item We consider policies with preemption. Our proposed method can also handle non-preemptive policies by keeping track of the occupied servers as part of the state.
    %\item While we focus on scheduling and admission control separately, one can also solve these problems jointly using our methodology for improved performance. One can also consider state-dependent admission control policies, rather than fixed buffer sizes as we do in this work.
    \item We focus on scheduling and admission control problems in queuing network satisfying Assumptions \ref{ass:queue}, but the algorithmic ideas can be extended to more general queuing networks by utilizing a larger state space that contains the residual workloads of \emph{all} jobs in the network, rather than only the top-of-queue jobs as is done in this work. A higher dimensional state descriptor is required for more general networks as multiple jobs in the same queue can be served simultaneously.
    \item Beyond queuing network control, our methodology can be extended to control problems in other discrete-event dynamical systems. More explicitly, our methodology can handle systems that involve a state update of the form $x_{k+1} = g(x_{k}, e_{k+1})$ where $g$ is a differentiable function and $e_{k+1}$ is the selected event. Recall that in this work, the state update is linear in $x_{k}$ and $e_{k+1}$: $x_{k+1} = x_{k} + De_{k+1}$. We also require that $e_{k+1}$ is differentiable almost surely in the action $u_{k}$. 
\end{itemize}

% Acknowledgments---Will not appear in anonymized version

%% ========================== Bibliography =========================  = %%

\bibliography{./bib}

\bibliographystyle{abbrvnat}

\setlength{\bibsep}{.7em}

\newpage
\appendix

\section{Training Details}
\label{sec:training-details}

$\ppo$ WC was trained over $100$ episodes, each consisting of $50,000$ environment steps parallelized over $50$ actors. We closely follow the hyper-parameters and training setup as in~\cite{dai2022queueing}. We used a discount factor of $0.998$, a GAE~\cite{schulman2015high} parameter of $0.99$, and set the Kullback–Leibler divergence penalty as $0.03$. For the value network, we used a batch size of $2,500$, while for the policy network, we used the entire rollout buffer (batch size of $50,000$) to take one gradient step. We performed $3$ PPO gradient updates on the same rollout data. For all the experiments, we used the Adam optimizer with a cosine decaying warming-up learning rate scheduler. The learning rates were set to $3 \times 10^{-4}$ for the value network and $9 \times 10^{-4}$ for the policy network. We used 3\% of the training horizon to warm up to the maximum learning rate and then cosine decayed to $1 \times 10^{-5}$ for both networks. We used the same neural network architecture as those in~\cite{dai2022queueing}, see Appendix E of ~\cite{dai2022queueing} for more details.

For the $\pathwise$ policy gradient, we used the same hyperparameters across all experiments. We use an inverse temperature of $\beta = 10$ for the $\softmax$ relaxation of the event selection. We update the policy after every episode with the Adam optimizer, using constant step-size of $5\times 10^{-4}$, momentum parameters $(0.8, 0.9)$, and gradient clipping of $1$. For the policy neural network, we used a multilayer perceptron with 3 hidden layers, each hidden layer consisting of 128 hidden units. We use the work-conserving $\softmax$ for the final output.
% \subsection{Direct smoothing across inverse temperatures $\beta$}
% \label{sec:app-inv-temp}

% Figure~\ref{fig:direct_inv_temp} displays the average cosine similiarty of gradient estimators (with $B = 1$ trajectory) under direct smoothing for different inverse temperature parameters. The gradient estimators either suffer from high bias or high variance and are unable to achieve a high cosine similarity with the true gradient, which is estimated from the $\mathsf{REINFORCE}$ estimator. Our approach, which preserves the dynamics of the sample paths is able to achieve almost perfect cosine similarity with the true gradient under an inverse temperature of $\beta = 1.0$. This illustrates the importance of preserving the dynamics when generating trajectories.

% \begin{figure}[ht]
% \centering
%     \includegraphics[height = 2.3in]{plot/direct_smoothing_cos.png}
    
%     \caption{Comparison of average cosine similarity of gradient estimators using direct smoothing with $\beta  \in \{1, 10, 100, 1000 \}$ for the criss-cross network under a randomized backpressure policy for $N = 1000$ steps. The gradient estimators either suffer from high bias or high variance and are unable to achieve a high cosine similarity with the true gradient.}
%     \label{fig:direct_inv_temp}
%\end{figure}
\section{Proofs}
\label{sec:appendix_proofs}

\subsection{Proof of Theorem~\ref{thm:mm1-smoothing}}
\label{section:mm1-smoothing-proof}
We focus on the case where $x_k\geq1$. Then,
\begin{align*}
x_{k+1} & =x_{k}+1\{\tau_k^A<w_k/\mu\}-1\{\tau_k^A>w_k/\mu\}\\
 & =x_{k}+\frac{e^{-\beta \tau_k^A}}{e^{-\beta \tau_k^A}+e^{-\beta w_k/\mu}}-\frac{e^{-\beta w_k/\mu}}{e^{-\beta \tau_k^A}+e^{-\beta w_k/\mu}}\\
 & =x_{k}+\frac{e^{-\beta \tau_k^A}-e^{-\beta w_k/\mu}}{e^{-\beta \tau_k^A}+e^{-\beta w_k/\mu}}.
\end{align*}
Since the inter-arrival times and workloads are exponentially distributed, by the memoryless property, we have $\tau_k^A\sim\mathsf{Exp}\left(\lambda\right)$ and
$w_k\sim \mathsf{Exp}\left(1\right)$.

The true gradient is
\[
\frac{d}{d\mu}\mathbb{E}[x_{k+1}-x_{k}]=\frac{d}{d\mu}\frac{\lambda-\mu}{\lambda+\mu}=-\frac{2\lambda}{(\lambda+\mu)^{2}}.
\]
Under our $\mathsf{softmin}_{\beta}$ approximation for the event-selection, 
%the one-step transition is
%\[
%\mathbb{E}[x_{t+1}-x_{t}]=\frac{\lambda-\mu}{\lambda+\mu}
%\]
%the $\pathwise$ gradient can be expressed as
we have
\begin{align*}
 & \mathbb{E}\left[\frac{d}{d\mu}\frac{e^{-\beta \tau_k^A}-e^{-\beta w_k/\mu}}{e^{-\beta \tau_k^A}+e^{-\beta w_k/\mu}}\right]\\
 %& =\mathbb{E}\left[\frac{d}{d\mu}\frac{e^{-\beta \tau_k^A}-e^{-\beta w_k/\mu}}{e^{-\beta \tau_k^A}+e^{-\beta w_k/\mu}}\right]\\
 & =\mathbb{E}\left[-2\beta\frac{e^{-\beta(\tau_k^A+w_k/\mu)}}{(e^{-\beta \tau_k^A}+e^{-\beta w_k/\mu})^{2}}\frac{w_k}{\mu^{2}}\right]\\
 %& =-2\frac{\beta}{\mu}\mathbb{E}\left[S_{i}\left(\frac{e^{\beta(T_{i}+S_{i})}}{(e^{\beta T_{i}}+e^{\beta S_{i}})^{2}}1\{T_{i}<S_{i}\}+\frac{e^{\beta(T_{i}+S_{i})}}{(e^{\beta T_{i}}+e^{\beta S_{i}})^{2}}1\{T_{i}>S_{i}\}\right)\right]\\
 %& =-2\frac{\beta}{\mu}\mathbb{E}\left[S_{i}\left(\frac{e^{\beta(T_{i}-S_{i})}}{(e^{\beta(T_{i}-S_{i})}+1)^{2}}1\{T_{i}<S_{i}\}+\frac{e^{\beta(S_{i}-T_{i})}}{(e^{\beta(S_{i}-T_{i})}+1)^{2}}1\{T_{i}>S_{i}\}\right)\right]\\
 & =-2\frac{\beta}{\mu}\mathbb{E}\left[\tau_k^S\left(\frac{e^{\beta(\tau_k^A-\tau_k^S)}}{(e^{\beta(\tau_k^A-\tau_k^S)}+1)^{2}}1\{\tau_k^A<\tau_k^S\}+\frac{e^{\beta(\tau_k^S-\tau_k^A)}}{(e^{\beta(\tau_k^S-\tau_k^A)}+1)^{2}}1\{\tau_k^A>\tau_k^S\}\right)\right],
\end{align*}
for $\tau_k^S=w_k/\mu$.
%
% as
% \[
% \frac{e^{\beta(T_{i}+S_{i})}}{(e^{\beta T_{i}}+e^{\beta S_{i}})^{2}}=\frac{e^{\beta(T_{i}+S_{i})}}{e^{2\beta S_{i}}(e^{\beta(T_{i}-S_{i})}+1)^{2}}=\frac{e^{\beta(T_{i}-S_{i})}}{(e^{\beta(T_{i}-S_{i})}+1)^{2}}
% \]
%
Next, note that
\begin{align*}
 & \mathbb{E}\left[\tau_k^S\frac{e^{\beta(\tau_k^A-\tau_k^S)}}{(e^{\beta(\tau_k^A-\tau_k^S)}+1)^{2}}1\{\tau_k^A<\tau_k^S\}\right]\\
 & =\mathbb{E}\left[\left.\mathbb{E}\left[\tau_k^S\frac{e^{\beta(\tau_k^A-\tau_k^S)}}{(e^{\beta(\tau_k^A-\tau_k^S)}+1)^{2}}1\{\tau_k^A<\tau_k^S\}\right]\right |\tau_k^A=t\right]\\
 & =\mathbb{E}\left[\left.\mathbb{E}\left[\tau_k^S\frac{e^{\beta(t-\tau_k^S)}}{(e^{\beta(t-\tau_k^S)}+1)^{2}}\right|\tau_k^A=t,\tau_k^S>t\right]\mathbb{P}(\tau_k^S>t|\tau_k^A=t)\right]\\
 %& =\mathbb{E}\left[\mathbb{E}\left[(t+S')\frac{e^{-\beta S'}}{(e^{-\beta S'}+1)^{2}}\right]\mathbb{P}(\tau_k^S>t|\tau_k^A=t)\right]\\
 & =\mathbb{E}\left[\mathbb{E}\left[(t+S')\frac{e^{-\beta S'}}{(e^{-\beta S'}+1)^{2}}\right]\mathbb{P}(\tau_k^S>t|\tau_k^A=t)\right] \mbox{ for $S'\sim \mathsf{Exp}(\mu)$}\\
 & =\mathbb{E}\left[(tA(\beta,\mu)+B(\beta,\mu))\mathbb{P}(\tau_k^S>t|\tau_k^A=t)\right]\\
 & =\mathbb{E}\left[(\tau_k^A A(\beta,\mu)+B(\beta,\mu))e^{-\mu \tau_k^A}\right]\\
 & =\frac{\lambda}{(\lambda+\mu)^{2}}A(\beta,\mu)+\frac{\lambda}{(\lambda+\mu)}B(\beta,\mu)
\end{align*}
where
\begin{align*}
A(\beta,\mu) & =\frac{\mu\left(\beta-\mu H\left(\frac{\mu}{2\beta}\right)+\mu H\left(\frac{\mu}{2\beta}-\frac{1}{2}\right)\right)}{2\beta^{2}}\\
 & =\frac{\mu}{2\beta}+\frac{\mu^{2}}{2\beta^{2}}\left(\underbrace{H\left(\frac{\mu}{2\beta}-\frac{1}{2}\right)-H\left(\frac{\mu}{2\beta}\right)}_{\tilde{H}(\beta,\mu)}\right)\\
B(\beta,\mu) & =\frac{\mu}{4\beta^{3}}\left(2\beta H\left(\frac{\mu}{2\beta}\right)-2\beta H\left(\frac{\mu}{2\beta}-\frac{1}{2}\right)-\mu\psi^{(1)}\left(\frac{\beta+\mu}{2\beta}\right)+\mu\psi^{(1)}\left(\frac{2\beta+\mu}{2\beta}\right)\right)\\
 & =-\frac{\mu}{2\beta^{2}}\tilde{H}(\beta,\mu)+\frac{\mu^{2}}{4\beta^{3}}\left(\underbrace{\psi^{(1)}\left(\frac{2\beta+\mu}{2\beta}\right)-\psi^{(1)}\left(\frac{\beta+\mu}{2\beta}\right)}_{\tilde{\psi}^{(1)}(\beta,\mu)}\right)
\end{align*}
Moreover, note that
\begin{align*}
H\left(\frac{\mu}{2\beta}\right) & =\psi^{(0)}\left(\frac{\mu}{2\beta}+1\right)+\gamma\\
H\left(\frac{\mu}{2\beta}-\frac{1}{2}\right) & =\psi^{(0)}\left(\frac{\mu}{2\beta}+\frac{1}{2}\right)+\gamma
\end{align*}
Note that
\begin{align*}
H\left(\frac{\mu}{2\beta}\right)-H\left(\frac{\mu}{2\beta}-\frac{1}{2}\right) & =\log\left(\frac{\mu}{2\beta}+1\right)+\frac{1}{\frac{\mu}{2\beta}+1}\\
 & -\log\left(\frac{\mu}{2\beta}+\frac{1}{2}\right)+\frac{1}{\frac{\mu}{2\beta}+1}.
\end{align*}

Similarly,
\begin{align*}
 & \mathbb{E}\left[\tau_k^S\frac{e^{\beta(\tau_k^S-\tau_k^A)}}{(e^{\beta(\tau_k^S-\tau_k^A)}+1)^{2}}1\{\tau_k^A>\tau_k^S\}\right]\\
 & =\mathbb{E}\left[\mathbb{E}\left[\left.s\frac{e^{-\beta(\tau_k^A-s)}}{(e^{-\beta(\tau_k^A-s)}+1)^{2}}1\{\tau_k^A>s\}\right|\tau_k^S=s\right]\right]\\
 & =\mathbb{E}\left[\mathbb{E}\left[\left.s\frac{e^{-\beta(\tau_k^A-s)}}{(e^{-\beta(\tau_k^A-s)}+1)^{2}}\right|\tau_k^A>s,\tau_k^S=s\right]\mathbb{P}(\tau_k^A>s|\tau_k^S=s)\right]\\
 & =\mathbb{E}\left[\mathbb{E}\left[s\frac{e^{-\beta T'}}{(e^{-\beta T'}+1)^{2}}\right]\mathbb{P}(\tau_k^A>s|\tau_k^S=s)\right] \mbox{ for $T'\sim \mathsf{Exp}(\lambda)$}\\
 & =\mathbb{E}\left[\tau_k^S A(\beta,\lambda)e^{-\lambda \tau_k^S}\right]\\
 & =\frac{\mu}{(\lambda+\mu)^{2}} A(\beta,\lambda).
\end{align*}
Then,
\begin{align*}
% & -2\frac{\beta}{\mu}\mathbb{E}\left[\tau_k^S\left(e^{\beta(\tau_k^A-\tau_k^S)}1\{\tau_k^A<\tau_k^S\}+e^{\beta(\tau_k^S-\tau_k^A)}1\{\tau_k^A>\tau_k^S\}\right)\right]\\
 &-2\frac{\beta}{\mu}\mathbb{E}\left[\tau_k^S\left(\frac{e^{\beta(\tau_k^A-\tau_k^S)}}{(e^{\beta(\tau_k^A-\tau_k^S)}+1)^{2}}1\{\tau_k^A<\tau_k^S\}+\frac{e^{\beta(\tau_k^S-\tau_k^A)}}{(e^{\beta(\tau_k^S-\tau_k^A)}+1)^{2}}1\{\tau_k^A>\tau_k^S\}\right)\right]\\
 & =-2\frac{\beta}{\mu}\left(\frac{\lambda}{(\lambda+\mu)^{2}}A(\beta,\mu)+\frac{\lambda}{(\lambda+\mu)}B(\beta,\mu)+\frac{\mu}{(\lambda+\mu)^{2}}A(\beta,\lambda)\right)\\
 & =-2\frac{\beta}{\mu}\left(\frac{\lambda A(\beta,\mu)+\mu A(\beta,\lambda)}{(\lambda+\mu)^{2}}+\frac{\lambda}{(\lambda+\mu)}B(\beta,\mu)\right)\\
 & =\frac{-2\lambda}{(\lambda+\mu)^{2}}-\frac{2\beta}{\mu}\left(\frac{\lambda\mu^{2}\tilde{H}(\beta,\mu)+\mu\lambda^{2}\tilde{H}(\beta,\lambda)}{2\beta^{2}(\lambda+\mu)^{2}}-\frac{\lambda\mu}{2\beta^{2}(\lambda+\mu)}\tilde{H}(\beta,\mu)+\frac{\lambda}{(\lambda+\mu)}\frac{\mu^{2}}{4\beta^{3}}\tilde{\psi}^{(1)}(\beta,\mu)\right).
\end{align*}
Note that as $\beta\to\infty$,
\begin{align*}
\lim_{\beta\to\infty}\tilde{H}(\beta,\mu) & =\gamma+\psi^{(0)}\left(\frac{1}{2}\right),\\
\lim_{\beta\to\infty}\tilde{\psi}^{(1)}(\beta,\mu) & =-\frac{\pi^{2}}{3}.
\end{align*}
This means that the leading order term is $O(1/\beta^{2})$. In particular,
\begin{align*}
-\frac{2\beta}{\mu}\left(\frac{\lambda\mu^{2}\tilde{H}(\beta,\mu)+\mu\lambda^{2}\tilde{H}(\beta,\lambda)}{2\beta^{2}(\lambda+\mu)^{2}}-\frac{\lambda\mu\tilde{H}(\beta,\mu)}{2\beta^{2}(\lambda+\mu)}\right) & \sim\frac{\pi^{2}\lambda^{2}(\mu-\lambda)}{6\beta^{2}(\lambda+\mu)^{2}}
\end{align*}
Finally, we have the second-order term
\[
-\frac{2\beta}{\mu}\frac{\lambda}{(\lambda+\mu)}\frac{\mu^{2}}{4\beta^{3}}\tilde{\psi}^{(1)}(\beta,\mu)\sim\frac{\pi^{2}\lambda\mu}{6\beta^{2}(\lambda+\mu)}.
\]
Thus, we have the following characterization of the bias:
\[
\mathbb{E}\left[\frac{d}{d\mu}\frac{e^{-\beta \tau_k^A}-e^{-\beta w_k/\mu}}{e^{-\beta \tau_k^A}+e^{-\beta w_k/\mu}}\right]-\left(\frac{-2\lambda}{(\lambda+\mu)^{2}}\right)\sim\frac{1}{\beta^{2}}\frac{\pi^{2}\lambda(\mu^{2}-\lambda^{2}+2\mu\lambda)}{6(\lambda+\mu)^{2}}+o\left(\frac{1}{\beta^{2}}\right)
\]

For variance, we have
\begin{align*}
 & \mathbb{E}\left[\left(-2\beta\frac{e^{\beta(\tau_k^A+w_k/\mu)}}{(e^{\beta \tau_k^A}+e^{\beta w_k/\mu})^{2}}\frac{w_k}{\mu^{2}}\right)^{2}\right]\\
 & =\mathbb{E}\left[\frac{4\beta^{2}}{\mu^{2}}\frac{e^{2\beta(\tau_k^A+\tau_k^S)}}{\left(e^{\beta \tau_k^A}+e^{\beta \tau_k^S}\right)^{4}}\tau_k^{S,2}\right]\\
 & =\frac{4\beta^{2}}{\mu^{2}}\mathbb{E}\left[\tau_k^{S,2}\left(\frac{e^{2\beta(\tau_k^A+\tau_k^S)}}{\left(e^{\beta \tau_k^A}+e^{\beta \tau_k^S}\right)^{4}}1\{\tau_k^A<\tau_k^S\}+\frac{e^{2\beta(\tau_k^A+\tau_k^S)}}{\left(e^{\beta \tau_k^A}+e^{\beta \tau_k^S}\right)^{4}}1\{\tau_k^A>\tau_k^S\}\right)\right]\\
 & =\frac{4\beta^{2}}{\mu^{2}}\mathbb{E}\left[\tau_k^{S,2}\left(\frac{e^{2\beta(\tau_k^A-\tau_k^S)}}{\left(e^{\beta(\tau_k^A-\tau_k^S)}+1\right)^{4}}1\{\tau_k^A<\tau_k^S\}+\frac{e^{2\beta(\tau_k^S-\tau_k^A)}}{\left(e^{\beta(\tau_k^S-\tau_k^A)}+1\right)^{4}}1\{\tau_k^A>\tau_k^S\}\right)\right].
\end{align*}
Note that
\begin{align*}
 & \mathbb{E}\left[\tau_k^{S,2}\frac{e^{2\beta(\tau_k^A-\tau_k^S)}}{\left(e^{\beta(\tau_k^A-\tau_k^S)}+1\right)^{4}}1\{\tau_k^A<\tau_k^S\}\right]\\
 & =\mathbb{E}\left[\mathbb{E}\left[\left.\tau_k^{S,2}\frac{e^{2\beta(\tau_k^A-\tau_k^S)}}{\left(e^{\beta(\tau_k^A-\tau_k^S)}+1\right)^{4}}1\{t<\tau_k^S\}\right]\right|\tau_k^A=t\right]\\
 %& =\mathbb{E}\left[\mathbb{E}\left[S_{i}^{2}\frac{e^{2\beta(t-S_{i})}}{(e^{\beta(t-S_{i})}+1)^{2}}|T_{i}=t,S_{i}>t\right]\mathbb{P}(S_{i}>t|T_{i}=t)\right]\\
 & =\mathbb{E}\left[\mathbb{E}\left[(t+S')^{2}\frac{e^{-2\beta S'}}{(e^{-\beta S'}+1)^{4}}\right]\mathbb{P}(\tau_k^S>t|\tau_k^A=t)\right] \mbox{ for $S'\sim \mathsf{Exp}(\mu)$}\\
 & =\mathbb{E}\left[\mathbb{E}\left[(t^{2}+2S'+S'^{2})\frac{e^{-2\beta S'}}{(e^{-\beta S'}+1)^{4}}\right]\mathbb{P}(S_{i}>t|T_{i}=t)\right]\\
 & =\mathbb{E}\left[\left(t^{2}\tilde{A}(\beta,\mu)+t\tilde{B}(\beta,\mu)+\tilde{C}(\beta,\mu)\right)\mathbb{P}(S_{i}>t|T_{i}=t)\right]\\
 & =\mathbb{E}\left[\left(T_{i}^{2}\tilde{A}(\beta,\mu)+T_{i}\tilde{B}(\beta,\mu)+\tilde{C}(\beta,\mu)\right)e^{-\mu T_{i}}\right]\\
 & =\frac{2\lambda}{(\lambda+\mu)^{3}}\tilde{A}(\beta,\mu)+\frac{\lambda}{(\lambda+\mu)^{2}}\tilde{B}(\beta,\mu)+\frac{\lambda}{(\lambda+\mu)}\tilde{C}(\beta,\mu).
\end{align*}
Similarly,
\begin{align*}
 & \mathbb{E}\left[\tau_k^{S,2}\frac{e^{2\beta(\tau_k^S-\tau_k^A)}}{\left(e^{\beta(\tau_k^S-\tau_k^A)}+1\right)^{4}}1\{\tau_k^A>\tau_k^S\}\right]\\
 & =\mathbb{E}\left[\mathbb{E}\left[\left.s^{2}\frac{e^{2\beta(s-\tau_k^A)}}{\left(e^{\beta(s-\tau_k^A)}+1\right)^{4}}1\{\tau_k^A>s\}\right|\tau_k^S=s\right]\right]\\
 & =\mathbb{E}\left[\mathbb{E}\left[s^{2}\frac{e^{-2\beta(\tau_k^A-s)}}{(e^{-\beta(\tau_k^A-s)}+1)^{4}}|\tau_k^A>s,\tau_k^S=s\right]\mathbb{P}(\tau_k^A>s|\tau_k^S=s)\right]\\
 & =\mathbb{E}\left[\mathbb{E}\left[s^{2}\frac{e^{-2\beta T'}}{(e^{-\beta T'}+1)^{4}}\right]\mathbb{P}(\tau_k^A>s|\tau_k^S=s)\right] \mbox{ for $T'\sim \mathsf{Exp}(\lambda)$}\\
 & =\mathbb{E}\left[\tilde{A}(\beta,\lambda)\tau_k^{S,2}e^{-\lambda \tau_k^S}\right]\\
 & =\frac{2\mu}{(\lambda+\mu)^{2}}\tilde{A}(\beta,\lambda).
\end{align*}
Putting the above two parts together, we have
\begin{align*}
 %& \frac{4\beta^{2}}{\mu^{2}}\mathbb{E}\left[S_{i}^{2}\left(\frac{e^{\beta(T_{i}-S_{i})2}}{(e^{\beta(T_{i}-S_{i})}+1)^{4}}1\{T_{i}<S_{i}\}+\frac{e^{\beta(S_{i}-T_{i})2}}{(e^{\beta(S_{i}-T_{i})}+1)^{4}}1\{T_{i}>S_{i}\}\right)\right]\\
 &\frac{4\beta^{2}}{\mu^{2}}\mathbb{E}\left[\tau_k^{S,2}\left(\frac{e^{2\beta(\tau_k^A-\tau_k^S)}}{\left(e^{\beta(\tau_k^A-\tau_k^S)}+1\right)^{4}}1\{\tau_k^A<\tau_k^S\}+\frac{e^{2\beta(\tau_k^S-\tau_k^A)}}{\left(e^{\beta(\tau_k^S-\tau_k^A)}+1\right)^{4}}1\{\tau_k^A>\tau_k^S\}\right)\right]\\
 & =\frac{4\beta^{2}}{\mu^{2}}\left(\frac{2\lambda}{(\lambda+\mu)^{3}}\tilde{A}(\beta,\mu)+\frac{\lambda}{(\lambda+\mu)^{2}}\tilde{B}(\beta,\mu)+\frac{\lambda}{(\lambda+\mu)}\tilde{C}(\beta,\mu)+\frac{2\mu}{(\lambda+\mu)^{2}}\tilde{A}(\beta,\lambda)\right)\\
 & \sim\frac{4\beta\lambda}{3\mu(\lambda+\mu)^{2}} \mbox{ as $\beta\to\infty$.}
\end{align*}

\subsection{Proof of Theorem~\ref{thm:mm1_variance}}
\label{sec:proof_mm1_variance}

By assumption, $h \leq c(\mu - \lambda)$ for some $c < 1$. First, we develop bound for $\var_{\infty} (\hat{\nabla}^{\mathsf{R}} J_{N}(\theta; \xi_{1:N}))$.
%= \Theta\left(N^{-1} \theta^{-2} (1-\rho)^{-4} \right)$. 
We can compute the variance by conditioning on the value of $Y$:
\begin{align*}
    \var_{\infty}(\hat{Q}_{N}(\mu-hY)\cdot(\log Y-\frac{1}{\theta}))&=\mathbb{E}\left[\var_{\infty}\left(\hat{Q}_{N}(\mu-hy)(\log y-\frac{1}
{\theta})|Y=y\right)\right]\\
&+\var\left(\mathbb{E}\left[\hat{Q}_{N}(\mu-hy)(\log y-\frac{1}{\theta})|Y=y)\right]\right).
\end{align*}
For the first term, note that the asymptotic variance in the CLT for the ergodic estimator $\hat{Q}_{N}(\mu)$ is $\var_{\infty}(\hat Q(\mu))=\frac{2\rho(1+\rho)}{(1-\rho)^{4}}$. Then, we have $\var_{\infty}(\hat{Q}_{N}(\mu)) = \frac{2\rho(1+\rho)}{N(1-\rho)^{4}}$. Since $h \leq \mu$, $\var_{\infty}(\hat{Q}_{N}(\mu))= \frac{2\rho(1+\rho)}{N(1-\rho)^{4}}$.
Then,
\begin{align*}
&\E \left[ \var_{\infty}\left(\hat{Q}_{N}(\mu-hy)\cdot(\log y+\frac{1}{\theta})|Y = y\right) \right] \\
& =\E \left[  (\log y+\frac{1}{\theta})^{2} \var_{\infty}(\hat{Q}_{N}(\mu-hy)|Y = y) \right]\\
& \geq \E \left[  (\log Y+\frac{1}{\theta})^{2}
\frac{2\rho(1+\rho)}{N(1-\rho)^{4}} \right]\\
& = \frac{1}{\theta^{2}}
\frac{2\rho(1+\rho)}{N(1-\rho)^{4}}=\Theta\left( (1-\rho)^{-4} \right),
\end{align*}
where the last equality uses the fact that for $Y\sim\text{Beta}(\theta,1)$,
\[
\E[\log Y]=\psi(\theta) - \psi(\theta + 1)=-\frac{1}{\theta}
\]
and
\[
\E \left[  (\log Y+\frac{1}{\theta})^{2}\right]=\var(\log Y) = \psi_{1}(\theta) - \psi_{1}(\theta + 1) = \frac{1}{\theta^{2}}.
\]
For the second term, we plug in the true estimand as the expectation of $Q_{N}(\mu)$, i.e. $Q(\mu) = \frac{\lambda}{\mu - \lambda}$. Then,
\begin{align*}
&\var\left(\mathbb{E}\left[\hat{Q}_{N}(\mu-hy)(\log y+\frac{1}{\theta})|Y=y)\right]\right) \\
& = \var\left(\frac{\lambda}{\mu - hY - \lambda}
\left( \log Y + \frac{1}{\theta} \right)\right) \\
& = \E \left[ \left(\frac{\lambda}{\mu - hY - \lambda} \right)^{2}
\left( \log Y + \frac{1}{\theta} \right)^{2} \right]
- \E \left[ \left(\frac{\lambda}{\mu - hY - \lambda} \right)
\left( \log Y + \frac{1}{\theta} \right) \right]^{2}.
\end{align*}
We proceed to evaluate these expectations analytically.
\begin{align*}
& \E \left[ \left(\frac{\lambda}{\mu - hY - \lambda} \right)
\left( \log Y + \frac{1}{\theta} \right) \right] \\
&= \frac{\rho}{1-\rho}
\left(
\frac{\Gamma(\theta)}{\Gamma(1+\theta)}
F^{2}_{1}
\left(1,\theta,1+\theta,\frac{h}{\mu-\lambda} \right)
-
\theta\Phi \left(\frac{h}{\mu - \lambda},2,\theta \right)
\right) \\
&=O((1-\rho)^{-1}),
\end{align*}
where $F^{2}_{1}$ is the Hypergoemetric 2F1  function and $\Phi$ is the Lerch transcendental function.
We also have
\begin{align*}
& \E \left[ \left(\frac{\lambda}{\mu - hY - \lambda} \right)^{2}
\left( \log Y + \frac{1}{\theta} \right)^{2} \right] \\
& = \frac{\lambda^{2}}{\theta^{2}(\mu - \lambda)^{3}}
\left(
2(\mu - \lambda)
+ h\theta^{3}\Phi \left(\frac{h}{\mu - \lambda},2,\theta+1 \right)
+ h\theta^{3}(\theta - 1)\Phi \left(\frac{h}{\mu - \lambda},3,\theta+1 \right) \right. \\
& 
+ (\mu - \lambda) 
\left( \theta\frac{\mu - \lambda}{\mu - h - \lambda} - (\theta - 1) F_{1}^{2}(1,\theta,1+\theta,\frac{h}{\mu-\lambda})
\right) \\
&\left. +2(\mu -\lambda) F_{2}^{3}\left((2,\theta,\theta),(1+\theta,1+\theta),\frac{h}{\mu-\lambda}\right)
\right) \\
& = O((1-\rho)^{-2}).
\end{align*}
Taking the difference between the above two parts under the limit as $(1-\rho)\to 0$, we have
\[
\var\left(\frac{\lambda}{\mu - hY - \lambda}
\left( \log Y + \frac{1}{\theta} \right)\right) =O((1-\rho)^{-2}).
\]

Next, we develop a bound for $\var_{\infty} (\hat{\nabla} J_{N}(\theta; \xi_{1:N}))$.
\begin{align*}
    \var_{\infty}\left(
    h \cdot \hat{\nabla} Q_{N}(\mu - h Y) \left( \frac{1}{\theta} Y \log Y \right)
    \right)&=
    \E\left[\var_{\infty}\left(h \cdot \hat{\nabla} Q_{N}(\mu - h Y) \left( \frac{1}{\theta} Y \log Y \right)|Y=y\right)\right]\\
&+\var\left(\mathbb{E}\left[h \cdot \hat{\nabla} Q_{N}(\mu - h Y) \left( \frac{1}{\theta} Y \log Y \right)|Y = y\right]\right)
\end{align*}
For the first term, we can use the fact that $|Y\log Y| \leq 1/e$ almost surely since $Y\in[0,1]$. We also use recent results in~\cite{hu2023comparison}, which compute the asymptotic variance of the IPA estimator:
\[
\var_{\infty}(\hat{\nabla} Q_{N}(\mu)) = \frac{1 + 16\rho + 27 \rho^{2} + 2 \rho^{3} + 6\rho^{4}}{\mu^{2}N(1 + \rho)(1-\rho)^{5}} \leq 52\mu^{-2}N^{-1}(1-\rho)^{-5}
\]
Under $\mu - hy$, the congestion factor $1 - \frac{\lambda}{\mu - hy} = \frac{\mu - hy - \lambda}{\mu - hy} \geq \frac{\mu - cy(\mu - \lambda) - \lambda}{\mu} = (1-cy)(1-\rho)$ and $\mu-hy\geq \mu-h\geq (1-c)\mu$.
So we have the bound,
\begin{align*}
 &\var_{\infty}\left(h\hat{\nabla} Q_{N}(\mu - h Y) \left( \frac{1}{\theta} y \log y\right)|Y=y\right) \\
 & \leq  h^{2} \theta^{-2}(y \log y)^{2}\var_{\infty}(\hat{\nabla} Q_{N}(\mu - h y)) \\
 & \leq h^{2}\mu^{-2}\theta^{-2}e^{-2} 52N^{-1}(1-c)^{-7}(1-\rho)^{-5} \\
 & = O(N^{-1} h^{2} \mu^{-2} (1-\rho)^{-5}) \\
 &= O(N^{-1}  (1-\rho)^{-3})
\end{align*}
since $h = O(1-\rho)$.

For the second term, we plug in the true estimand as the mean of $\hat{\nabla} Q_{N}(\mu)$, i.e., $ \nabla Q(\mu) = -\frac{\rho}{\mu(1-\rho)^{2}}$,
\begin{align*}
&\E\left[h \cdot \hat{\nabla} Q_{N}(\mu - h Y) \left( \frac{1}{\theta} Y \log Y \right)|Y = y\right] 
= - h\frac{1}{\theta} (y \log y) \frac{\lambda }{(\mu - hy - \lambda)^{2}}.
\end{align*}
% Note that for any $y\in [0,1]$ this has lower bound $0$ and upper bound $h\theta^{-1} e^{-1} \frac{\rho}{(\mu - h)(1-c)^{2}(1-\rho)^{2}}$, which gives a variance upper bound of
We next evaluate the variance analytically,
\begin{align*}
&\var\left(
- h\frac{1}{\theta} (Y \log Y) \frac{\lambda }{(\mu - hY - \lambda)^{2}}
\right) \\
&= 
\E\left[ 
\left(- h\frac{1}{\theta} (Y \log Y) \frac{\lambda }{(\mu - hY - \lambda)^{2}}
\right)^{2}
\right] 
- \E\left[ 
- h\frac{1}{\theta} (Y \log Y) \frac{\lambda }{(\mu - hY - \lambda)^{2}}
\right]^{2}
\end{align*}
Since
\begin{align*}
&\E\left[ 
- h\frac{1}{\theta} (Y \log Y) \frac{\lambda }{(\mu - hY - \lambda)^{2}}
\right] \\
&= \frac{h}{(1+\theta)^{2}}
\frac{\lambda}{(\mu - \lambda)^{2}}
F^{3}_{2} 
\left((2,1+\theta,1+\theta), 
(2+\theta, 2+ \theta), \frac{h}{\mu -\lambda} \right) \\
& = O((1-\rho)^{-1})
\end{align*}
and
\begin{align*}
&\E\left[ 
\left(- h\frac{1}{\theta} (Y \log Y) \frac{\lambda }{(\mu - hY - \lambda)^{2}}
\right)^{2}
\right] \\
&=
2h\theta^{2}\Gamma(\theta)^{3}\Gamma(2+\theta)^{-3}\frac{\lambda^{2}}{(\mu - \lambda)^{2}} \times 
\left( F^{4}_{3} 
\left((3,1+\theta,1+\theta,1+\theta), 
(2+\theta, 2+ \theta,2+ \theta), \frac{h}{\mu -\lambda} \right) \right. \\
& \left. -F^{4}_{3} 
\left((4,1+\theta,1+\theta,1+\theta), 
(2+\theta, 2+ \theta,2+ \theta), \frac{h}{\mu -\lambda} \right) \right) \\
& = O((1-\rho)^{-2}),
\end{align*}
we have
\[
\var\left(
- h\frac{1}{\theta} (Y \log Y) \frac{\lambda }{(\mu - hY - \lambda)^{2}}
\right) = O( (1-\rho)^{-2}).
\]

\subsection{Proof of Corollary~\ref{cor:reinforce-baseline}}
\label{sec:proof_reinforce_baseline}

First, we can explicitly characterize the optimal baseline:
\begin{align*}
b^{*} &= \frac{\E[Q(\mu -hY) \nabla_{\theta} \log \pi_{\theta}(Y)^{2}]}{\E[\nabla_{\theta}\log \pi_{\theta}(Y)^{2}]} \\
 &= \frac{\E[Q(\mu -hY) \left( \log Y + 1/\theta \right)^{2}]}{\E[\left( \log Y + 1/\theta \right)^{2}]} \\
&= \frac{\lambda}{\mu - \lambda}\underbrace{\left[F^{2}_{1}\left(1,\theta,1+\theta,\frac{h}{\mu-\lambda}\right) -2\theta^{2}\Phi\left(\frac{h}{\mu-\lambda},2,\theta\right) + 2\frac{h}{\mu-\lambda}\theta^{3}\Phi\left(\frac{h}{\mu-\lambda},3,1+\theta\right)\right]}_{b(\theta)}  \\
& = O((1-\rho)^{-1}).
\end{align*}
Next, we plug this into the $\reinforce$ estimator. 
\begin{align*}
    \var_{\infty}((\hat{Q}_{N}(\mu-hY)-b^*)\cdot(\log Y-\frac{1}{\theta}))&=\mathbb{E}\left[\var_{\infty}\left((\hat{Q}_{N}(\mu-hy)-b^*)(\log y-\frac{1}
{\theta})|Y=y\right)\right]\\
&+\var\left(\mathbb{E}\left[(\hat{Q}_{N}(\mu-hy) - b^{*})(\log y-\frac{1}{\theta})|Y=y\right]\right).
\end{align*}
Note that since $b^{*}$ is a constant, the first term, i.e., the mean of the conditional variance given $Y$, has the same value as in Theorem \ref{thm:mm1_variance}. For the second term, note that since $h = c(\mu - \lambda)$,
% For the second, i.e., we can explicitly compute the variance of the conditional mean given $Y$ with Mathematica. However, this leads to an unwieldy expressio
\begin{align*}
& \var \left[ \left(\frac{\lambda}{\mu - hY - \lambda}  - b^{*}\right)
\left( \log Y + \frac{1}{\theta} \right) \right]  \\
&= \left(\frac{\lambda}{\mu - \lambda}\right)^{2} \var \left[ \left(\frac{1}{1- cY}  - b(\theta)\right)
\left( \log Y + \frac{1}{\theta} \right) \right].
\end{align*}
Since $\var \left[ \left(\frac{1}{1- cY}  - b(\theta)\right)
\left( \log Y + \frac{1}{\theta} \right) \right] > 0$ and doesn't depend on $\mu$ or $\lambda$, this confirms that the second term is $\Theta((1-\rho)^{-2})$

%\input{tv-bounds}
% \input{proof-limit-experiment}
% \input{proof-mdp}
% \input{proof-asymptotic}
% \input{implementation}

%%%%%%%%%%%%%%%%%%%%%%%%%%%%%%%%%%%%%%%%%%%%%%%%%%%%%%%%%%%%%%%%%%%%%%%%%%%%%%%
%%%%%%%%%%%%%%%%%%%%%%%%%%%%%%%%%%%%%%%%%%%%%%%%%%%%%%%%%%%%%%%%%%%%%%%%%%%%%%%
% APPENDIX
%%%%%%%%%%%%%%%%%%%%%%%%%%%%%%%%%%%%%%%%%%%%%%%%%%%%%%%%%%%%%%%%%%%%%%%%%%%%%%%
%%%%%%%%%%%%%%%%%%%%%%%%%%%%%%%%%%%%%%%%%%%%%%%%%%%%%%%%%%%%%%%%%%%%%%%%%%%%%%%

\end{document}